\documentclass[11pt]{article}

\usepackage[preprint]{acl}
\usepackage{times}
\usepackage{latexsym}

\usepackage[utf8]{inputenc}
\usepackage[T1]{fontenc}
\usepackage{hyperref}
\usepackage{url}
\usepackage{booktabs}
\usepackage{multirow}
\usepackage{amsmath,amssymb}
\usepackage{graphicx}
\usepackage{subcaption}
\usepackage{xcolor}
\usepackage{framed}
\usepackage{microtype}
\usepackage{nicefrac}
\usepackage{placeins}
\usepackage{tikz}
\usetikzlibrary{arrows.meta,positioning,fit}

\definecolor{forwardblue}{RGB}{44,103,173}
\definecolor{returnred}{RGB}{190,67,54}
\definecolor{storegray}{RGB}{245,246,248}
\definecolor{tablemethod}{RGB}{226,239,249}
\definecolor{resultgreen}{RGB}{20,135,75}
\definecolor{resultred}{RGB}{190,55,50}
\definecolor{qualplum}{RGB}{88,65,103}
\definecolor{qualmoss}{RGB}{93,103,62}
\definecolor{qualochre}{RGB}{151,101,35}
\definecolor{qualforest}{RGB}{48,91,72}
\definecolor{qualink}{RGB}{65,68,65}
\definecolor{casegray}{RGB}{103,103,97}
\definecolor{casesage}{RGB}{89,108,94}

\newcommand{\method}{\textsc{MemCryst}}
\newcommand{\preq}{S_{\mathrm{preq}}}
\newcommand{\term}{S_{\mathrm{term}}}
\newcommand{\gain}[1]{\textcolor{resultgreen}{\raisebox{.15ex}{$\uparrow #1$}}}
\newcommand{\drop}[1]{\textcolor{resultred}{\raisebox{.15ex}{$\downarrow #1$}}}
\newcommand{\scoregain}[2]{#1\hspace{.15em}\gain{#2}}
\newcommand{\scoredrop}[2]{#1\hspace{.15em}\drop{#2}}
\newcommand{\tablecell}[2]{\makebox[#1][c]{\strut #2}}
\newcommand{\tableleftcell}[2]{\makebox[#1][l]{\strut #2}}
\newcommand{\methodrowcell}[2]{%
  \begingroup
  \setlength{\fboxsep}{0pt}%
  \colorbox{tablemethod}{\makebox[#1][c]{\rule[-.65ex]{0pt}{2.9ex}#2}}%
  \endgroup}
\newcommand{\methodrowleftcell}[2]{%
  \begingroup
  \setlength{\fboxsep}{0pt}%
  \colorbox{tablemethod}{\makebox[#1][l]{\rule[-.65ex]{0pt}{2.9ex}\hspace{.4em}#2}}%
  \endgroup}

\newcommand{\qualcard}[4]{%
  \begin{minipage}[t]{#1}%
  \vspace{0pt}% establish a true top-alignment point for unequal card heights
  \setlength{\fboxsep}{0pt}%
  \setlength{\fboxrule}{.45pt}%
  \fcolorbox{#2!35!white}{#2!6!white}{%
    \begin{minipage}{\dimexpr\linewidth-2\fboxrule\relax}%
      \colorbox{#2!18!white}{%
        \makebox[\linewidth][l]{%
          \hspace*{6pt}\textcolor{#2!90!black}{\bfseries\strut #3}}}%
      \par
      \hspace*{6pt}%
      \begin{minipage}{\dimexpr\linewidth-12pt\relax}%
        \vspace{4pt}\footnotesize\raggedright #4%
        \par\vspace{5pt}%
      \end{minipage}%
      \hspace*{6pt}%
    \end{minipage}%
  }%
  \end{minipage}%
}
\newcommand{\cardmeta}[1]{\textcolor{qualink}{\scriptsize #1}}
\newcommand{\cardlabel}[1]{\textcolor{qualink}{\scriptsize\bfseries\MakeUppercase{#1}}}
\newcommand{\recordid}[1]{\textcolor{qualink}{\mbox{#1}}}
\newcommand{\casecard}[5]{%
  \qualcard{#1}{#2}{#3}{%
    \begin{minipage}[t][#4][t]{\linewidth}%
      \vspace{0pt}\raggedright #5%
    \end{minipage}}%
}
\newcommand{\caserow}[2]{%
  \begin{minipage}[t][12mm][t]{\linewidth}%
    \vspace{0pt}\raggedright\cardlabel{#1}\par\vspace{3pt}%
    #2\par
  \end{minipage}\par
}
\newcommand{\casecode}[1]{{\fontsize{9}{10}\selectfont\texttt{#1}}}
\newenvironment{casefigure}{%
  \par\smallskip\noindent
  \begin{minipage}[t]{\textwidth}
  \captionsetup{type=figure}
  \centering\small
  \setlength{\abovedisplayskip}{4pt}%
  \setlength{\belowdisplayskip}{4pt}%
  \setlength{\abovedisplayshortskip}{3pt}%
  \setlength{\belowdisplayshortskip}{3pt}%
}{%
  \end{minipage}\par\smallskip
}

\newcounter{algorithm}
\newcounter{algline}[algorithm]
\newcommand{\algline}[2][0pt]{%
  \stepcounter{algline}%
  \noindent
  \hangindent=2.7em\relax
  \advance\hangindent by #1\relax
  \hangafter=1
  \makebox[2.2em][r]{\arabic{algline}:}\hspace{.5em}%
  \hspace{#1}#2\par
}
\newenvironment{paperalgorithm}[2][]{%
  \par\medskip
  \refstepcounter{algorithm}%
  \begingroup
  \small
  \raggedright
  \setlength{\parindent}{0pt}%
  \setlength{\parskip}{0pt}%
  \hrule height .8pt
  \kern2pt
  \noindent\textbf{Algorithm~\thealgorithm:} #2\par
  \kern2pt
  \hrule height .4pt
  \kern2pt
}{%
  \kern1pt
  \hrule height .8pt
  \endgroup
  \par\medskip
}

\title{Lifelong Context Adaptation through Bidirectional Memory--Skill Learning}
\author{Tianyi Xu \qquad Huazheng Wang\thanks{Corresponding author.} \\
  {\normalfont Oregon State University} \\
  {\normalfont\ttfamily \{xut2, huazheng.wang\}@oregonstate.edu}}

\begin{document}
\maketitle

\begin{abstract}
Lifelong LLM agents must learn from an expanding history of interactions while making each decision within a bounded context window.
Memory makes past experience available, but the agent may still need to infer how to solve each new task from the retrieved records.
Skills provide reusable procedures, yet may omit the task-specific facts needed to apply them.
Because retrieved memory and skills shape the agent's behavior, they also influence the experience available for future learning.
We introduce \method{}, a framework for lifelong context adaptation through bidirectional learning of memory and skills with a frozen language model.
Recurring episodes stored in memory support skill creation and revision, and the interactions guided by those skills supply new memory for subsequent learning.
Facts retrieved from memory and learned skills jointly form the agent's context, while task outcomes adjust the future retrieval influence of the records actually served without rewriting their content.
Across ten settings spanning embodied control, long-document reading, conversation, mathematics, and code, \method{} leads seven reported comparisons.
On ALFWorld, joint serving achieves $63.17\%$ success, compared with $59.83\%$ for memory-only and $50.50\%$ for skill-only serving.
Code is available at \url{https://github.com/tianyi0216/Memcryst.git}.
\end{abstract}

% BEGIN INLINED FROM sections/01_introduction.tex
\section{Introduction}

LLM agents deployed over long periods encounter tasks whose solutions depend on earlier experience \citep{expel,memoryr1}.
An agent may need to recall an entity's properties, reuse a successful procedure, or avoid a failure that occurred many interactions earlier.
When the language model remains frozen, this experience can affect decisions only through the context supplied to it.
The interaction history grows throughout deployment, while the context available for any one decision remains bounded.
Lifelong adaptation therefore requires the agent to extract reusable decision structure from this history, preserve evidence for the current task, and adjust retrieval as further outcomes become available.
We study how an agent can learn this external state online so that past experience improves subsequent decisions.

Previous work has used external memory to store past interactions, facts, or compressed notes for retrieval on later tasks \citep{amem,lightmem,simplemem}.
These records provide evidence from earlier experience, but the agent may still need to work out a procedure each time a similar task recurs.
Other approaches learn reusable procedures, or skills, from experience to guide later tasks \citep{voyager,skillweaver,trace2skill}, but these procedures may omit the task-specific facts and constraints needed to apply them.
Recent systems combine memory and skills: MUSE-AutoSkill organizes experience around evolving skills, while MSCE uses stored evidence and feedback to guide skill development \citep{museautoskill,msce2026}.
Maintaining both, however, does not by itself ensure that the retrieved facts and skills are useful for the current task.
A fact may be correct but irrelevant, and a skill may not fit the current situation.
Failure can therefore inform future retrieval without establishing that the retrieved content should be rewritten.
Meanwhile, skill-guided interactions supply new memory for subsequent skill revision.
The challenge is to learn from this experience while using task outcomes to adjust which facts and skills enter later contexts.

We introduce \method{} to adapt context through bidirectional memory--skill learning.
Even with frozen model parameters, retrieved context changes the agent's actions and thus the experiences available for subsequent learning.
\method{} stores experience in memory as episodic records and typed facts, and consolidates recurring episodes into conditional skills.
At deployment, facts provide task-relevant evidence alongside the procedures supplied by skills.
Skill-guided interactions then enter memory and support later skill revision, completing the reverse path from skills to memory through behavior.
In parallel, each task outcome updates the retrieval influence of the facts and skills actually served.
Content construction draws on accumulated experience, while outcome return adjusts future use without regenerating record text.
Both processes adapt the agent's context while leaving the language model frozen.

Across ten settings spanning embodied control, long-document reading, conversation, mathematics, and code, \method{} leads seven reported comparisons.
On ALFWorld, it reaches $63.17\%$ success, exceeding memory-only by $3.33$ percentage points and skill-only serving by $12.67$ points.
A separate $2\!\times\!2$ intervention yields an observed interaction of $+5.33$ points between fact and skill return.
One shared, domain-scoped state also improves the mean score from $28.30\%$ to $44.50\%$ over 1,018 queries interleaved across five domains.
Frozen-state controls show higher success with relevant than random records, while skill-based context uses fewer injected tokens than episodic context.
These results support the following contributions:
\begin{itemize}
    \item \textbf{Bidirectional memory--skill learning.} We couple episode-based skill creation and revision with skill-guided behavior that supplies new memory for subsequent learning.
    \item \textbf{Content and use adaptation.} We separate evidence-based content updates from outcome-based updates to the retrieval influence of served facts and skills, coordinating both within a bounded context.
    \item \textbf{Empirical evidence for the coupled design.} We evaluate the complete system across task settings and interleaved deployment, and test joint serving, skill revision, and the interaction between the two return channels.
\end{itemize}
% END INLINED FROM sections/01_introduction.tex
% BEGIN INLINED FROM sections/02_related_work.tex
\section{Related work}

\paragraph{Memory construction and retrieval.}
External memory preserves experience in forms ranging from distilled guidance to source-linked evidence.
ExpeL, SAGE, and Dynamic Cheatsheet retain experience as insights, reflections, or strategies \citep{expel,sage,dynamiccheatsheet}.
A-Mem organizes linked notes, while TriMem combines source-linked dialogue with atomic facts and profiles \citep{amem,trimem}.
LightMem and SimpleMem compress evidence through sleep-time and within-session consolidation, respectively \citep{lightmem,simplemem}.
Repeated textual consolidation can, however, discard useful details \citep{faultymem2026}.
\method{} retains episode evidence for skill learning alongside independently retrievable typed facts.

\paragraph{Skill construction and revision.}
Voyager and SkillWeaver develop executable skills through interaction and refine them through execution feedback or testing \citep{voyager,skillweaver}, while EvoPool evolves executable annotator pools using validation feedback \citep{xu2026evopoolevolutionaryprogrammaticannotation}.
Agent Workflow Memory abstracts workflows from trajectories, while Trace2Skill combines trace-level lessons into portable skill files \citep{awm,trace2skill}.
MemP combines trajectories with abstract scripts and supports revision after failure, and AutoSkill maintains versioned skills from user interactions \citep{memp,autoskill2026}.
In \method{}, revision draws on both new episodes and the supporting episodes of an existing skill.
It preserves the skill's identity and accumulated use statistics, while the revised procedure can be served alongside independently retrieved facts.

\paragraph{Context processing and consolidation.}
For long-input processing, MemAgent repeatedly updates bounded textual memory, while Multi-Head Recurrent Memory updates one memory head at a time \citep{memagent,mhm2026}.
The Sleep paradigm consolidates knowledge into model parameters through distillation and replay \citep{sleepllm2026}.
\method{} maintains experience across deployment events in external stores, using a frozen model for both task execution and skill consolidation.

\begin{figure*}[!t]
\centering
\includegraphics[width=\linewidth]{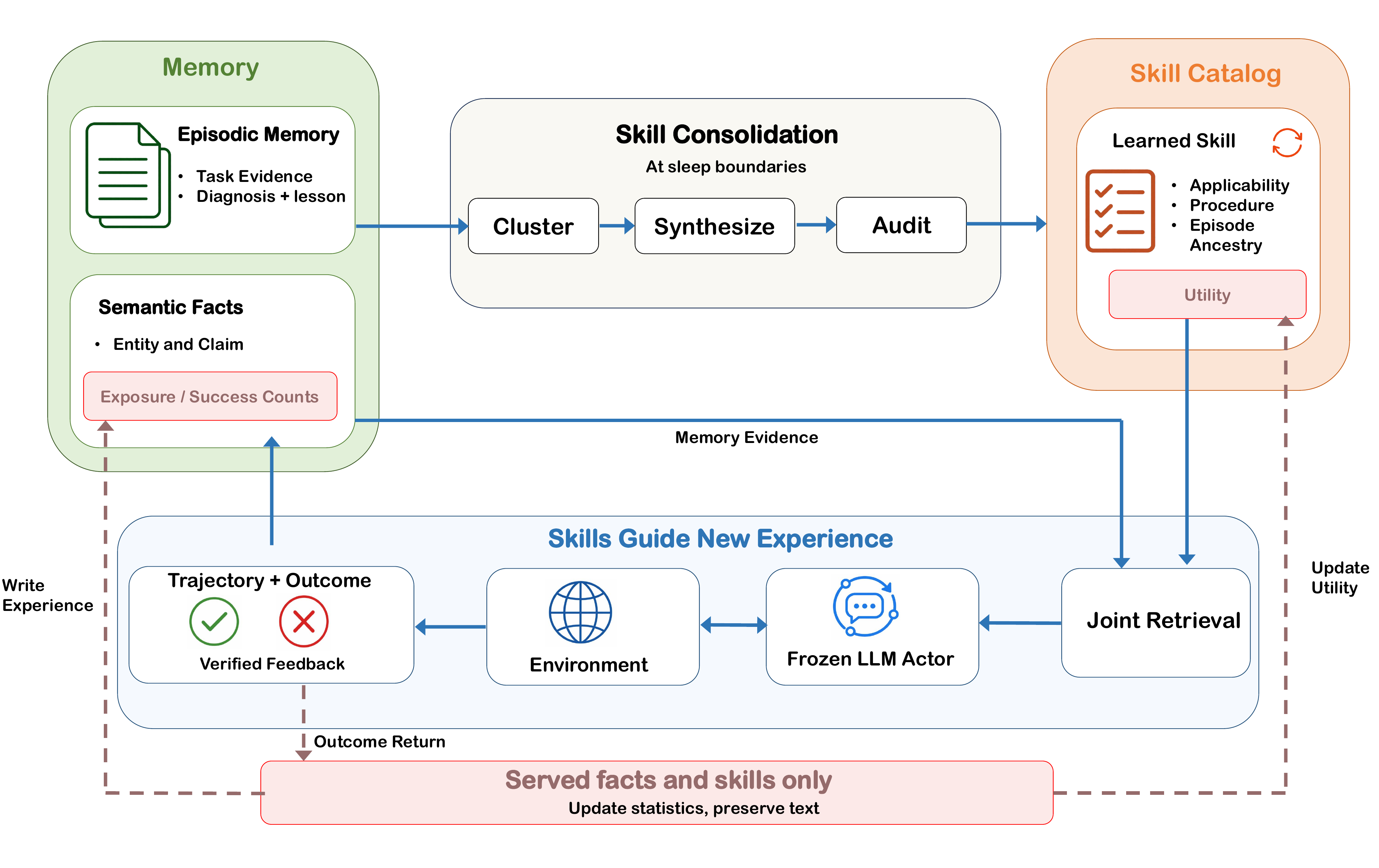}
\vspace{-22pt}
\caption{\textbf{Bidirectional memory--skill learning through context and behavior.}
Recurring episodes support skill creation and revision.
Jointly served memory and skills guide the frozen actor, whose completed interactions supply new memory for later skill consolidation.
Task outcomes separately update the retrieval influence of exposed facts and skills without rewriting their text.
Solid arrows show content and behavior flow, and dashed arrows show statistical updates.
Return applies before new records are written.}
\vspace{-15pt}
\label{fig:loop}
\end{figure*}

\paragraph{Joint learning and outcome feedback.}
Memory-R1 learns memory management and use through reinforcement learning, whereas Live-Evo adjusts experience weights from reward differences between executions with and without memory \citep{memoryr1,liveevo}.
MUSE-AutoSkill maintains skill-specific usage notes and revises skills from execution feedback \citep{museautoskill}.
MSCE uses reflection-weighted terminal values of traces to guide retrieval and skill evolution, with skill reliability also governing retrieval \citep{msce2026}.
EvoChamber extends coevolution to memories, team collaboration, and population structure \citep{evochamber2026}; UCoE co-evolves verifiers with agent harness state \citep{xu2026verifierharnesscoevolutiontrainingfreeselfimprovement}, and EvalResearchBench studies autonomous evaluator construction \citep{zhang2026evalresearchbenchaiagentsbuild}.
In \method{}, skill-guided interactions supply episodes for subsequent skill revision.
Unlike Live-Evo's paired comparison, return uses the observed task outcome and applies it only to records actually served.
It updates fact exposure and success counts and skill utilities without rewriting their content.
Episode-based consolidation remains responsible for skill creation and revision.
% END INLINED FROM sections/02_related_work.tex
% BEGIN INLINED FROM sections/03_method.tex

\section{Method}
\label{sec:method}

\method{} learns what experience to reuse and how to use it through the cycle in Figure~\ref{fig:loop}.
We first define adaptation through external context, then describe how memory supports skills, how skill-guided behavior supplies new memory, and how outcomes change later retrieval.

\subsection{Problem setup}

We consider an agent processing an ordered stream of deployment events $q_1,\ldots,q_T$.
Events either supply external evidence or require an action or answer.
The agent uses a frozen language model with policy $\pi_{\theta_0}$ and maintains external state $\Xi_t$ containing only information from earlier events.
For an event requiring an action or answer, it constructs context $I_t=I(q_t;\Xi_t)$ within the available token budget.
This context defines its effective policy:
\begin{equation}
 \pi_t^{\rm eff}(\tau\mid q)
 =\pi_{\theta_0}\!\left(\tau\mid q,I(q;\Xi_t)\right).
 \label{eq:effective-policy}
\end{equation}
The actor produces trajectory $\tau_t$ and receives reward $r_t\in[0,1]$.
After observing the outcome, the agent may update its external state for subsequent events.
Evidence-ingestion events may also update the state but are unscored.

The objective is to adapt the external state and the context it supplies to maximize expected prequential reward $\mathbb E[\preq]$.
Let $\mathcal I_T$ index the transitions scored in evaluation.
We consider streams starting from empty external state:
\begin{equation}
 \begin{aligned}
 \preq&=\frac1{|\mathcal I_T|}\sum_{t\in\mathcal I_T} r_t,\\
 \theta_t&=\theta_0,\qquad \Xi_1=\varnothing .
 \end{aligned}
 \label{eq:preq}
\end{equation}
Every scored transition counts, including the initial cold start and later online errors.

\subsection{Context construction and online updates}

To supply both task-specific evidence and reusable procedures within the context budget, \method{} maintains episodic memory, typed facts, and skills.
The external state separates stored content from the statistics controlling its use:
\begin{equation}
 \begin{aligned}
 \Theta_t&=\mathcal E_t\uplus\mathcal F_t\uplus\mathcal S_t,\\
 \Xi_t&=(\Theta_t,\rho_t).
 \end{aligned}
 \label{eq:state}
\end{equation}
Memory comprises episodes in $\mathcal E_t$ and typed facts in $\mathcal F_t$.
Episodes retain task evidence, outcome diagnoses, and lessons for later skill learning.
Facts retain concrete claims about entities, together with their sources and recurrence support.
The skill catalog $\mathcal S_t$ stores applicability conditions and parameterized procedures, linked to the episodes supporting them.
Skills enter context as textual guidance, not executable functions.
The actor uses the task, observations, and retrieved facts to apply their parameterized steps.
The metadata $\rho_t$ contains episode weights $w_e$, fact exposure and success counts $(n_f,h_f)$, and skill utilities $U_s$.

Each adapter chooses which record types to serve.
Retrieval $R$ filters by scope and applicability, ranks eligible records by relevance and retrieval statistics, and formats selected records under the adapter's context limits.
The retained records $\widehat{\mathcal U}_t$ form the context $I_t$ supplied to the frozen actor:
\begin{equation}
 \begin{aligned}
 \widehat{\mathcal U}_t&=R_{\rho_t}(q_t,\Theta_t),\\
 \tau_t&\sim\pi_{\theta_0}(\,\cdot\mid q_t,I_t).
 \end{aligned}
 \label{eq:servecycle}
\end{equation}
Exposure is defined by $x_{tv}=\mathbf 1[v\in\widehat{\mathcal U}_t]$, so only records present in the actor's context count as exposed.
Episodes remain available for later skill learning even when omitted from the actor's context.

After execution, task evaluation yields $o_t=(r_t,y_t)$, where $r_t$ is the reward and $y_t\in\{0,1\}$ is the evaluator-defined success signal used for state updates.
The external state is then updated in two stages:
\begin{equation}
 \begin{aligned}
 \Xi_t^{\rm ret}&=B(\Xi_t,o_t,\widehat{\mathcal U}_t),\\
 \Xi_{t+1}&=G(\Xi_t^{\rm ret},\tau_t,o_t).
 \end{aligned}
 \label{eq:cycle}
\end{equation}
The outcome-return operator $B$ updates the retrieval statistics of exposed facts and skills without rewriting their content.
The content update operator $G$ combines memory writing after each event with skill consolidation at scheduled sleep boundaries.
Return precedes new writes, so records created from the current outcome cannot receive return from that outcome.
Appendices~\ref{app:full-loop} and~\ref{app:retrieval} specify the event protocol, eligibility rules, and context assembly.

\subsection{From episodic evidence to reusable skills}

Recurring episodes provide evidence for a procedure that can be reused without replaying each source interaction in the actor's context.
At a sleep boundary, $G$ begins skill consolidation by mining related episodes from the episode store after the current memory write:
\begin{equation}
 (C_1,\ldots,C_M)
 =\operatorname{Mine}(\mathcal E_t^{\rm wr},\rho_t^{\rm wr}).
 \label{eq:crystallize}
\end{equation}
The miner ranks recurrent clusters using outcome quality, size, and cluster stability.
It retains failure evidence as well as successful episodes, allowing a proposed procedure to address recurring mistakes.

For each cluster, a frozen LM synthesizes a conditional procedure from a small subset of high-weight episodes.
Instance-specific values become parameter slots, while applicability conditions specify when the procedure should be used.
A separate audit in a fresh context checks the proposal's parametricity, coverage, and scope against the selected episodes.
Each admitted skill retains links to all episodes in its source cluster for later revision.

After a proposal passes the initial audit, $G$ checks whether its complete capability key matches an existing skill.
This key specifies the verb family, target type, affordance, and output.
Unmatched proposals enter the catalog unless blocked by the adapter's duplicate-slug guard.
For a match, $G$ combines the new cluster with the existing skill's supporting episodes, synthesizes a revision, and replaces the skill only after a second audit accepts it.
An accepted revision preserves the skill identity and accumulated utility while extending its links to supporting episodes.
Later experience can therefore refine an existing procedure rather than merely expand the catalog.
Appendix~\ref{app:crystallize} gives the ordered update, matching fields, and audit details.

\subsection{From skill-guided behavior to new memory}

Skills shape the agent's subsequent interactions, which supply new evidence for memory and later skill revision.
Together with retrieved facts and current observations, they condition $\tau_t$ in Eq.~\ref{eq:servecycle}.
For example, a cleaning skill can guide the agent to clean an object before placing it at the destination.
The resulting episode preserves evidence of how the procedure was applied, together with an outcome diagnosis and lessons for later revision.
Appendix~\ref{app:qualitative-examples} illustrates this path with saved cleaning episodes and skill snapshots.

After outcome return, $G$ reflects on the completed interaction and extracts permitted typed facts.
For a valid episode write,
\begin{align}
 \mathcal E_t^{\rm wr}
 &=\mathcal E_t\cup\{\mathcal W_E(\tau_t,o_t)\},\nonumber\\
 \mathcal F_t^{\rm wr}
 &=\operatorname{Merge}\!\left(\mathcal F_t,\mathcal X_F(\tau_t,o_t)\right).
 \label{eq:forwardwrite}
\end{align}
Here $\mathcal W_E$ writes the episode's evidence, diagnosis, and lesson, while $\mathcal X_F$ extracts facts and may return none.
Matching facts add recurrence support and provenance to the existing record for the same entity, scope, and claim, preserving its text and exposure statistics.
Unmatched facts create new records.
The new episode is available for subsequent skill mining, including a sleep pass at the current boundary.
Episodes thus support continued skill learning, while facts remain independently retrievable for later tasks.
Skills influence new memory through behavior without directly editing fact text.

\subsection{Adapting future context through outcomes}

A record's content and its usefulness during deployment answer different questions.
An unsuccessful task can motivate an adjustment to future retrieval influence without providing evidence to revise a factual claim or identify a faulty procedure step.
\method{} therefore uses outcome return to update influence separately from content updates:
\begin{equation}
 \Xi_t^{\rm ret}
 =B(\Xi_t,o_t,\widehat{\mathcal U}_t)
 =(\Theta_t,\rho_t^{\rm ret}).
 \label{eq:return-separation}
\end{equation}
The content and membership of $\Theta_t$ remain unchanged during this stage.

For facts, $n_f$ counts exposed tasks and $h_f$ counts successes on those tasks.
The fact outcome score $C_f=(h_f+1)/(n_f+2)$ is a smoothed success rate over exposed tasks.
For each skill, utility $U_s$ summarizes recent outcomes under its persistent identity.
Given update rate $\alpha_S\in(0,1]$, return applies
\begin{align}
 &U_s\leftarrow U_s+\alpha_Sx_{ts}(y_t-U_s),
 \label{eq:skillreturn}\\
 &\begin{aligned}
 n_f&\leftarrow n_f+x_{tf},\\
 h_f&\leftarrow h_f+x_{tf}y_t.
 \end{aligned}
 \label{eq:factreturn}
\end{align}
Only exposed records change.
The step size for $C_f$ decreases as exposures accumulate, whereas the constant-step skill update continues to emphasize recent use.
Both scores stay in $[0,1]$, with derivations in Appendix~\ref{app:return-properties}.
These are success statistics conditioned on exposure: $C_f$ is not the probability that a claim is true, and assigning the same outcome to all served records does not estimate individual causal contributions.

During later retrieval, $C_f$ and $U_s$ weight query relevance when ranking eligible facts and skills, respectively.
The content update operator carries these statistics forward when merging facts or revising skills, while new record identities receive their own initialization.
Consequently, the next context reflects both new content learned from the interaction and updated influence for previously served content.
Appendix~\ref{app:records} specifies update ownership and metadata initialization.
% END INLINED FROM sections/03_method.tex
% BEGIN INLINED FROM sections/04_experiments.tex
% BEGIN INLINED FROM sections/05_experiments.tex
\section{Experimental setup}
\label{sec:experiments}

\noindent\textbf{Model.}
We use frozen Qwen3-8B as our main backbone \citep{qwen3}.
We also evaluate Qwen3-1.7B and GPT-4o-mini \citep{gpt4omini} on ALFWorld to assess performance across backbones.

\noindent\textbf{Benchmarks and baselines.}
We evaluate ten settings across five domains.
ALFWorld covers embodied control \citep{alfworld}, while RULER and BABILong cover long-document reading \citep{ruler,babilong}.
We evaluate conversation memory on LoCoMo, LongMemEval, and name-substituted DialSim \citep{locomo,longmemeval,dialsim}.
MATH-hard tests mathematical reasoning \citep{mathdataset}, while MBPP, CodeContests, and LCB+ cover code generation \citep{mbpp,codecontests,livecodebench,livecodebenchpro}.
Our main baselines are ReAct without persistent state \citep{react}, experience-based ExpeL and MemP \citep{expel,memp}, and the memory systems SimpleMem and LightMem \citep{simplemem,lightmem}.

% BEGIN INLINED FROM tables/performance_streams.tex
\begin{table*}[t]
\centering
\small
\setlength{\tabcolsep}{0pt}
\renewcommand{\arraystretch}{1.04}

\textcolor{forwardblue}{\itshape Panel a: Embodied control and reading long documents}\par\vspace{1pt}
\begin{tabular}{@{}lllll@{}}
\toprule
\tableleftcell{.20\linewidth}{\textbf{Method}}
& \tablecell{.20\linewidth}{\textbf{ALFWorld}}
& \tablecell{.20\linewidth}{\textbf{RULER}}
& \tablecell{.20\linewidth}{\textbf{BABILong}}
& \tablecell{.20\linewidth}{\textbf{Overall}} \\
\midrule
\tableleftcell{.20\linewidth}{ReAct}
& \tablecell{.20\linewidth}{$51.50$}
& \tablecell{.20\linewidth}{$41.30$}
& \tablecell{.20\linewidth}{$34.50$}
& \tablecell{.20\linewidth}{$42.43$} \\
\tableleftcell{.20\linewidth}{ExpeL}
& \tablecell{.20\linewidth}{\scoredrop{$44.80$}{6.70}}
& \tablecell{.20\linewidth}{\scoregain{$\mathbf{71.27}$}{29.97}}
& \tablecell{.20\linewidth}{\scoregain{$44.50$}{10.00}}
& \tablecell{.20\linewidth}{\scoregain{$53.52$}{11.09}} \\
\tableleftcell{.20\linewidth}{MemP}
& \tablecell{.20\linewidth}{\scoredrop{$48.80$}{2.70}}
& \tablecell{.20\linewidth}{\scoregain{$62.55$}{21.25}}
& \tablecell{.20\linewidth}{\scoregain{$43.00$}{8.50}}
& \tablecell{.20\linewidth}{\scoregain{$51.45$}{9.02}} \\
\tableleftcell{.20\linewidth}{SimpleMem}
& \tablecell{.20\linewidth}{\scoredrop{$44.30$}{7.20}}
& \tablecell{.20\linewidth}{\scoredrop{$35.10$}{6.20}}
& \tablecell{.20\linewidth}{\scoredrop{$33.30$}{1.20}}
& \tablecell{.20\linewidth}{\scoredrop{$37.57$}{4.87}} \\
\tableleftcell{.20\linewidth}{LightMem}
& \tablecell{.20\linewidth}{\scoregain{$52.30$}{0.80}}
& \tablecell{.20\linewidth}{\scoredrop{$12.20$}{29.10}}
& \tablecell{.20\linewidth}{\scoredrop{$14.50$}{20.00}}
& \tablecell{.20\linewidth}{\scoredrop{$26.33$}{16.10}} \\
\methodrowleftcell{.20\linewidth}{\textbf{\method{}}}
& \methodrowcell{.20\linewidth}{\scoregain{$\mathbf{63.17}$}{11.67}}
& \methodrowcell{.20\linewidth}{\scoregain{$64.70$}{23.40}}
& \methodrowcell{.20\linewidth}{\scoregain{$\mathbf{59.00}$}{24.50}}
& \methodrowcell{.20\linewidth}{\scoregain{$\mathbf{62.29}$}{19.86}} \\
\bottomrule
\end{tabular}

\vspace{4pt}
\textcolor{forwardblue}{\itshape Panel b: Conversation and memory}\par\vspace{1pt}
\begin{tabular}{@{}lllll@{}}
\toprule
\tableleftcell{.20\linewidth}{\textbf{Method}}
& \tablecell{.20\linewidth}{\textbf{LoCoMo}}
& \tablecell{.20\linewidth}{\textbf{LongMemEval}}
& \tablecell{.20\linewidth}{\textbf{DialSim-subst.}}
& \tablecell{.20\linewidth}{\textbf{Overall}} \\
\midrule
\tableleftcell{.20\linewidth}{ReAct}
& \tablecell{.20\linewidth}{$0.73$}
& \tablecell{.20\linewidth}{$4.20$}
& \tablecell{.20\linewidth}{$7.54$}
& \tablecell{.20\linewidth}{$4.16$} \\
\tableleftcell{.20\linewidth}{ExpeL}
& \tablecell{.20\linewidth}{\scoredrop{$0.63$}{0.10}}
& \tablecell{.20\linewidth}{\scoregain{$6.85$}{2.65}}
& \tablecell{.20\linewidth}{\scoregain{$11.83$}{4.29}}
& \tablecell{.20\linewidth}{\scoregain{$6.44$}{2.28}} \\
\tableleftcell{.20\linewidth}{MemP}
& \tablecell{.20\linewidth}{\scoregain{$9.64$}{8.91}}
& \tablecell{.20\linewidth}{\scoregain{$5.29$}{1.09}}
& \tablecell{.20\linewidth}{\scoregain{$31.83$}{24.29}}
& \tablecell{.20\linewidth}{\scoregain{$15.59$}{11.43}} \\
\tableleftcell{.20\linewidth}{SimpleMem}
& \tablecell{.20\linewidth}{\scoregain{$26.85$}{26.12}}
& \tablecell{.20\linewidth}{\scoregain{$20.11$}{15.91}}
& \tablecell{.20\linewidth}{\scoregain{$47.85$}{40.31}}
& \tablecell{.20\linewidth}{\scoregain{$31.60$}{27.45}} \\
\tableleftcell{.20\linewidth}{LightMem}
& \tablecell{.20\linewidth}{\scoredrop{$0.67$}{0.06}}
& \tablecell{.20\linewidth}{\scoredrop{$1.55$}{2.65}}
& \tablecell{.20\linewidth}{$7.54$}
& \tablecell{.20\linewidth}{\scoredrop{$3.25$}{0.90}} \\
\methodrowleftcell{.20\linewidth}{\textbf{\method{}}}
& \methodrowcell{.20\linewidth}{\scoregain{$\mathbf{31.75}$}{31.02}}
& \methodrowcell{.20\linewidth}{\scoregain{$\mathbf{22.20}$}{18.00}}
& \methodrowcell{.20\linewidth}{\scoregain{$\mathbf{74.24}$}{66.70}}
& \methodrowcell{.20\linewidth}{\scoregain{$\mathbf{42.73}$}{38.57}} \\
\bottomrule
\end{tabular}

\vspace{4pt}
\textcolor{forwardblue}{\itshape Panel c: Mathematical reasoning and code generation}\par\vspace{1pt}
\begin{tabular}{@{}llllll@{}}
\toprule
\tableleftcell{.17\linewidth}{\textbf{Method}}
& \tablecell{.166\linewidth}{\textbf{MATH-hard}}
& \tablecell{.166\linewidth}{\textbf{MBPP}}
& \tablecell{.166\linewidth}{\textbf{CodeContests}}
& \tablecell{.166\linewidth}{\textbf{LCB+}}
& \tablecell{.166\linewidth}{\textbf{Overall}} \\
\midrule
\tableleftcell{.17\linewidth}{ReAct}
& \tablecell{.166\linewidth}{$57.50$}
& \tablecell{.166\linewidth}{$81.30$}
& \tablecell{.166\linewidth}{$26.23$}
& \tablecell{.166\linewidth}{$35.05$}
& \tablecell{.166\linewidth}{$50.02$} \\
\tableleftcell{.17\linewidth}{ExpeL}
& \tablecell{.166\linewidth}{\scoregain{$59.16$}{1.66}}
& \tablecell{.166\linewidth}{\scoregain{$\mathbf{82.60}$}{1.30}}
& \tablecell{.166\linewidth}{\scoregain{$\mathbf{31.30}$}{5.07}}
& \tablecell{.166\linewidth}{\scoregain{$36.04$}{0.99}}
& \tablecell{.166\linewidth}{\scoregain{$\mathbf{52.28}$}{2.26}} \\
\tableleftcell{.17\linewidth}{MemP}
& \tablecell{.166\linewidth}{\scoregain{$59.16$}{1.66}}
& \tablecell{.166\linewidth}{\scoregain{$81.70$}{0.40}}
& \tablecell{.166\linewidth}{\scoregain{$28.70$}{2.47}}
& \tablecell{.166\linewidth}{\scoredrop{$33.14$}{1.91}}
& \tablecell{.166\linewidth}{\scoregain{$50.68$}{0.66}} \\
\tableleftcell{.17\linewidth}{SimpleMem}
& \tablecell{.166\linewidth}{\scoregain{$58.02$}{0.52}}
& \tablecell{.166\linewidth}{\scoredrop{$79.80$}{1.50}}
& \tablecell{.166\linewidth}{\scoregain{$28.00$}{1.77}}
& \tablecell{.166\linewidth}{\scoregain{$36.51$}{1.46}}
& \tablecell{.166\linewidth}{\scoregain{$50.58$}{0.56}} \\
\tableleftcell{.17\linewidth}{LightMem}
& \tablecell{.166\linewidth}{\scoredrop{$56.87$}{0.63}}
& \tablecell{.166\linewidth}{\scoredrop{$80.70$}{0.60}}
& \tablecell{.166\linewidth}{\scoregain{$27.40$}{1.17}}
& \tablecell{.166\linewidth}{\scoregain{$35.61$}{0.56}}
& \tablecell{.166\linewidth}{\scoregain{$50.15$}{0.13}} \\
\methodrowleftcell{.17\linewidth}{\textbf{\method{}}}
& \methodrowcell{.166\linewidth}{\scoregain{$\mathbf{60.90}$}{3.40}}
& \methodrowcell{.166\linewidth}{\scoredrop{$80.42$}{0.88}}
& \methodrowcell{.166\linewidth}{\scoregain{$28.67$}{2.44}}
& \methodrowcell{.166\linewidth}{\scoregain{$\mathbf{37.08}$}{2.03}}
& \methodrowcell{.166\linewidth}{\scoregain{$51.77$}{1.75}} \\
\bottomrule
\end{tabular}

\vspace{4pt}
\textcolor{forwardblue}{\itshape Panel d: Overall across categories}\par\vspace{1pt}
\begin{minipage}{.62\linewidth}
\setlength{\tabcolsep}{0pt}
\begin{tabular}{@{}ll@{}}
\toprule
\tableleftcell{.50\linewidth}{\textbf{Method}}
& \tablecell{.50\linewidth}{\textbf{Overall}} \\
\midrule
\tableleftcell{.50\linewidth}{ReAct}
& \tablecell{.50\linewidth}{$32.20$} \\
\tableleftcell{.50\linewidth}{ExpeL}
& \tablecell{.50\linewidth}{\scoregain{$37.41$}{5.21}} \\
\tableleftcell{.50\linewidth}{MemP}
& \tablecell{.50\linewidth}{\scoregain{$39.24$}{7.04}} \\
\tableleftcell{.50\linewidth}{SimpleMem}
& \tablecell{.50\linewidth}{\scoregain{$39.92$}{7.71}} \\
\tableleftcell{.50\linewidth}{LightMem}
& \tablecell{.50\linewidth}{\scoredrop{$26.58$}{5.63}} \\
\methodrowleftcell{.50\linewidth}{\textbf{\method{}}}
& \methodrowcell{.50\linewidth}{\scoregain{$\mathbf{52.26}$}{20.06}} \\
\bottomrule
\end{tabular}
\end{minipage}
\vspace{-8pt}
\caption{\textbf{Prequential performance across ten task streams.}
Scores are percentages, with changes from ReAct in pp.
Overall equally weights benchmarks within panels and the three panel means in panel d.}
\vspace{-10pt}
\label{tab:performance}
\end{table*}

% END INLINED FROM sections/05_experiments.tex
% END INLINED FROM sections/04_experiments.tex
% BEGIN INLINED FROM sections/05_results.tex
% Main empirical results and mechanism analyses.
% BEGIN INLINED FROM sections/06_results.tex
% Queue the leading result tables together before the setup ends.
\begin{table*}[!t]
\centering
\small
\setlength{\tabcolsep}{0pt}
\renewcommand{\arraystretch}{1.06}
\begin{tabular}{@{}lllllll@{}}
\toprule
\tableleftcell{.18\linewidth}{\textbf{System}}
& \tablecell{.1366\linewidth}{\textbf{RULER}}
& \tablecell{.1366\linewidth}{\textbf{BABILong}}
& \tablecell{.1366\linewidth}{\textbf{MATH-hard}}
& \tablecell{.1366\linewidth}{\textbf{MBPP}}
& \tablecell{.1366\linewidth}{\textbf{LoCoMo}}
& \tablecell{.1366\linewidth}{\textbf{Overall}} \\
\midrule
\tableleftcell{.18\linewidth}{ReAct}
& \tablecell{.1366\linewidth}{$36.90$}
& \tablecell{.1366\linewidth}{$35.00$}
& \tablecell{.1366\linewidth}{$61.60$}
& \tablecell{.1366\linewidth}{$76.00$}
& \tablecell{.1366\linewidth}{$0.50$}
& \tablecell{.1366\linewidth}{$28.30$} \\
\tableleftcell{.18\linewidth}{Specialist}
& \tablecell{.1366\linewidth}{\scoregain{$64.70$}{27.80}}
& \tablecell{.1366\linewidth}{\scoregain{$50.00$}{15.00}}
& \tablecell{.1366\linewidth}{\scoredrop{$60.90$}{0.70}}
& \tablecell{.1366\linewidth}{\scoregain{$77.40$}{1.40}}
& \tablecell{.1366\linewidth}{\scoregain{$31.80$}{31.30}}
& \tablecell{.1366\linewidth}{\scoregain{$\mathbf{48.50}$}{20.20}} \\
\methodrowleftcell{.18\linewidth}{\textbf{\method{}}}
& \methodrowcell{.1366\linewidth}{\scoregain{$55.90$}{19.00}}
& \methodrowcell{.1366\linewidth}{\scoregain{$52.00$}{17.00}}
& \methodrowcell{.1366\linewidth}{\scoredrop{$58.30$}{3.30}}
& \methodrowcell{.1366\linewidth}{\scoregain{$80.00$}{4.00}}
& \methodrowcell{.1366\linewidth}{\scoregain{$25.30$}{24.80}}
& \methodrowcell{.1366\linewidth}{\scoregain{$44.50$}{16.20}} \\
\bottomrule
\end{tabular}
\vspace{-10pt}
\caption{\textbf{Performance on a stream that interleaves five domains.}
Scores are percentages.
Specialist uses separate domain runs.
\method{} uses one shared external state over 1,088 events.
Overall averages 1,018 scored queries.}
\vspace{-10pt}
\label{tab:capstone}
\end{table*}

% END INLINED FROM tables/performance_streams.tex

\noindent\textbf{Evaluation.}
We report prequential mean reward, scoring each task before its interaction updates the external state and excluding standalone evidence-ingestion events.
Benchmark metrics include success rate, recall, token F1, exact match, and unit-test reward.
All scores are percentages, and absolute differences are in percentage points, denoted by pp.
See Appendix~\ref{app:experimental-details} for detailed settings and aggregation rules.

\section{Results}
\label{sec:results}

\begin{table*}[t]
\centering
\small
\setlength{\tabcolsep}{0pt}
\renewcommand{\arraystretch}{1.04}
\textcolor{forwardblue}{\itshape ALFWorld prequential success rate}\par\vspace{1pt}
\begin{tabular}{@{}llll@{}}
\toprule
\tableleftcell{.34\linewidth}{\textbf{Backbone}}
& \tablecell{.20\linewidth}{\textbf{ReAct}}
& \tablecell{.26\linewidth}{\textbf{\method{}}}
& \tablecell{.20\linewidth}{\textbf{$\Delta$ pp}} \\
\midrule
\tableleftcell{.34\linewidth}{Qwen3-1.7B}
& \tablecell{.20\linewidth}{$13.50$}
& \methodrowcell{.26\linewidth}{$27.67$}
& \tablecell{.20\linewidth}{\gain{14.17}} \\
\tableleftcell{.34\linewidth}{Qwen3-8B}
& \tablecell{.20\linewidth}{$51.50$}
& \methodrowcell{.26\linewidth}{$63.17$}
& \tablecell{.20\linewidth}{\gain{11.67}} \\
\tableleftcell{.34\linewidth}{GPT-4o-mini}
& \tablecell{.20\linewidth}{$28.33$}
& \methodrowcell{.26\linewidth}{$40.00$}
& \tablecell{.20\linewidth}{\gain{11.67}} \\
\bottomrule
\end{tabular}
\vspace{-8pt}
\caption{\textbf{ALFWorld performance across backbones.}
Rates are percentages. Each task allows at most 30 actions.}
\vspace{-10pt}
\label{tab:backbone-diagnostic}
\end{table*}

\begin{table*}[t]
\centering
\small
\begin{minipage}[t]{.53\linewidth}
\vspace{0pt}
\centering
\textcolor{forwardblue}{\itshape Panel a: Memory, skills, and revision}\par\vspace{1pt}
\setlength{\tabcolsep}{0pt}
\begin{tabular}{@{}lll@{}}
\toprule
\tableleftcell{.56\linewidth}{\textbf{Configuration}}
& \tablecell{.20\linewidth}{\textbf{SR\%}}
& \tablecell{.24\linewidth}{\textbf{$\Delta$ vs. Full}} \\
\midrule
\tableleftcell{.56\linewidth}{ReAct}
& \tablecell{.20\linewidth}{$56.17$}
& \tablecell{.24\linewidth}{$-7.00$} \\
\tableleftcell{.56\linewidth}{Memory-only}
& \tablecell{.20\linewidth}{$59.83$}
& \tablecell{.24\linewidth}{$-3.33$} \\
\tableleftcell{.56\linewidth}{Skill-only serving}
& \tablecell{.20\linewidth}{$50.50$}
& \tablecell{.24\linewidth}{$-12.67$} \\
\tableleftcell{.56\linewidth}{No skill revision}
& \tablecell{.20\linewidth}{$62.00$}
& \tablecell{.24\linewidth}{$-1.17$} \\
\methodrowleftcell{.56\linewidth}{\textbf{Full \method{}}}
& \methodrowcell{.20\linewidth}{$\mathbf{63.17}$}
& \methodrowcell{.24\linewidth}{$0.00$} \\
\bottomrule
\end{tabular}
\end{minipage}\hfill
\begin{minipage}[t]{.43\linewidth}
\vspace{0pt}
\centering
\textcolor{forwardblue}{\itshape Panel b: Outcome-return channels}\par\vspace{1pt}
\setlength{\tabcolsep}{0pt}
\begin{tabular}{@{}lll@{}}
\toprule
& \multicolumn{2}{c}{\textbf{Skill return}} \\
\cmidrule(l){2-3}
\tableleftcell{.42\linewidth}{\textbf{Fact return}}
& \tablecell{.29\linewidth}{\textbf{Off}}
& \tablecell{.29\linewidth}{\textbf{On}} \\
\midrule
\tableleftcell{.42\linewidth}{Off}
& \tablecell{.29\linewidth}{$59.67$}
& \tablecell{.29\linewidth}{$57.17$} \\
\tableleftcell{.42\linewidth}{On}
& \tablecell{.29\linewidth}{$60.33$}
& \methodrowcell{.29\linewidth}{$\mathbf{63.17}$} \\
\midrule
\tableleftcell{.42\linewidth}{Interaction, pp}
& \multicolumn{2}{c}{\gain{\mathbf{5.33}}} \\
\bottomrule
\end{tabular}
\end{minipage}
\vspace{-5pt}
\caption{\textbf{ALFWorld component ablations over 600 tasks.}
All configurations use the same revisit guard and navigation hint.
Panel a removes skills, memory retrieval, or skill revision and reports differences from Full in percentage points.
Skill-only serving retains internal memory for skill learning.
Panel b switches fact and skill return while retaining construction and retrieval of memory and skills.}
\vspace{-10pt}
\label{tab:ablations}
\end{table*}

% Queue the lifecycle float early to keep it with the results tables.
\begin{figure*}[!t]
\centering
\begin{minipage}{.98\linewidth}
\begin{subfigure}[t]{.49\linewidth}
\centering
\includegraphics[width=\linewidth]{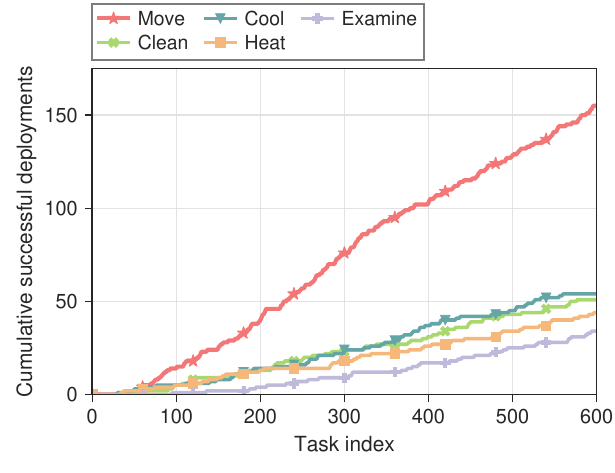}
\caption{Persistent skill reuse.}
\label{fig:coevolution-reuse}
\end{subfigure}\hfill
\begin{subfigure}[t]{.49\linewidth}
\centering
\includegraphics[width=\linewidth]{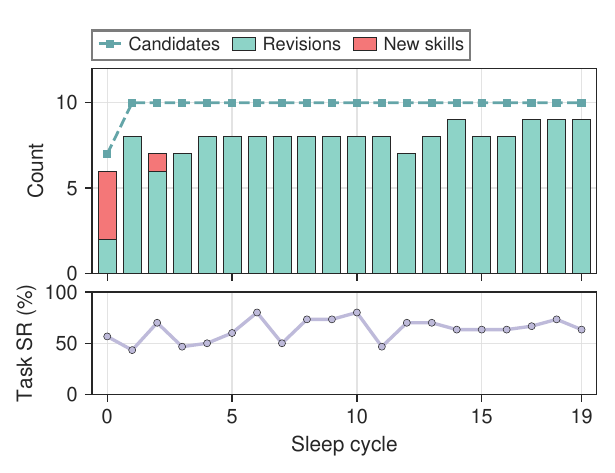}
\caption{Skill consolidation.}
\label{fig:coevolution-consolidation}
\end{subfigure}
\end{minipage}
\caption{\textbf{Evolution and reuse of the learned state.}
The ALFWorld run with both return channels enabled processes $T=600$ tasks and achieves $\preq=63.17\%$.
Successful deployments accumulate throughout the stream.
After cycle 2, all accepted updates revise existing skills instead of creating new ones.}
\vspace{-18pt}
\label{fig:coevolution}
\end{figure*}

\noindent\textbf{\method{} leads seven of ten settings.}
Table~\ref{tab:performance} reports an Overall of $52.26\%$ for \method{}, compared with $39.92\%$ for SimpleMem, the strongest baseline under this aggregate.
\method{} leads ALFWorld and BABILong and ranks second to ExpeL on RULER.
In ALFWorld, skills specify action sequences, while facts supply the object and environment information needed to apply them.
The joint-serving ablation below tests this complementarity.
The largest gains over ReAct occur in conversation memory, where \method{} also exceeds SimpleMem on all three settings.
On name-substituted DialSim, it reaches $74.24\%$ against SimpleMem's $47.85\%$.
Here, retrieval supplies names, dates, and events that a general procedure cannot recover from the question alone.
The larger gain after name substitution in Appendix~\ref{app:dialsim-intervention} further supports this role for deployment-specific evidence.
Benefits are smaller in mathematics and code, where ExpeL leads MBPP and CodeContests and \method{} falls below ReAct on MBPP.
Appendix~\ref{app:efficiency} shows a modest MBPP gain when episodes are retained for skill learning but withheld from the actor, supporting selective context construction.
See Appendices~\ref{app:baselines} and~\ref{app:msce} for additional baseline comparisons.

% BEGIN INLINED FROM tables/capstone.tex
% END INLINED FROM tables/capstone.tex

\noindent\textbf{A shared state improves overall performance under interleaving.}
On the interleaved stream in Table~\ref{tab:capstone}, \method{} uses one shared, domain-scoped state to raise mean performance from $28.30\%$ to $44.50\%$ across 1,018 queries in five domains.
It retains improvements on both document slices and about $80\%$ of the isolated LoCoMo gain.
The shared system scores $4.00$ points below the specialist mean of $48.50\%$, while exceeding the separate specialists on BABILong and MBPP.
These results show that one evolving state can supply useful context as deployment alternates between domains.
See Appendix~\ref{app:evaluation-config} for the stream composition and specialist settings.

\phantomsection\label{sec:backbone-robustness}
\noindent\textbf{Gains extend across model sizes and families.}
Table~\ref{tab:backbone-diagnostic} shows higher ALFWorld success with all three frozen backbones.
Relative to ReAct, \method{} gains $14.17$ percentage points with Qwen3-1.7B, $11.67$ with Qwen3-8B, and $11.67$ with GPT-4o-mini.
Each run builds its own external state from an empty start, adapting the context supplied to a backbone whose parameters remain frozen.
See Appendix~\ref{app:order-robustness} for additional Qwen3-8B results across shuffled task orders.

% BEGIN INLINED FROM tables/ablations.tex
% END INLINED FROM tables/ablations.tex

\begin{figure*}[!t]
\centering
\includegraphics[width=.98\linewidth]{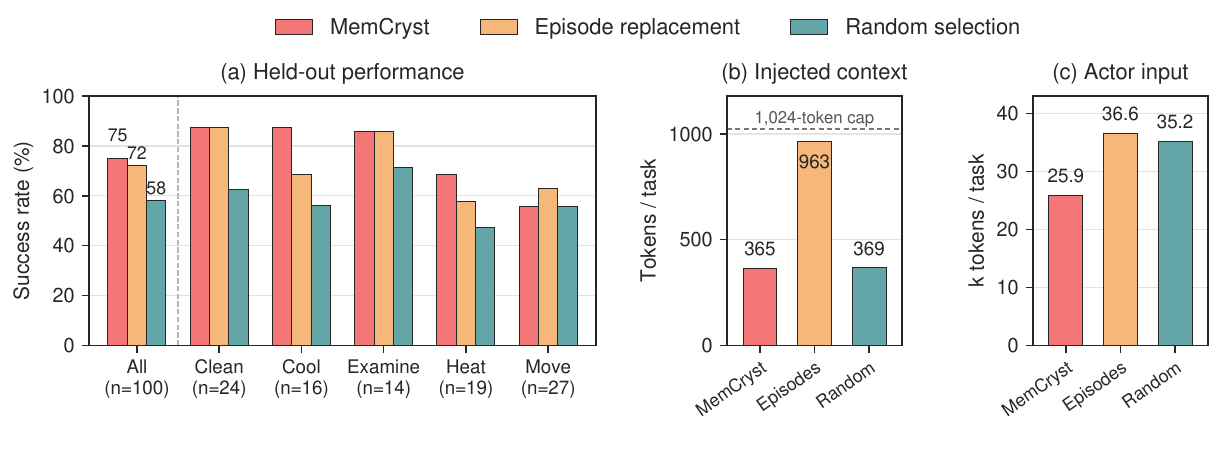}
\vspace{-20pt}
\caption{\textbf{Frozen-state context controls on 100 held-out ALFWorld tasks.}
All conditions use the same learned state and a 1,024-token injected-context limit.
Episode replacement keeps \method{}'s facts and replaces skills with retrieved episodic records.
Random selection matches \method{}'s record types and approximate context length.
Panel a shows overall and task-family success, with task counts below the axis.
Panels b and c show mean injected context and actor input per task, respectively.
Actor input includes retries.}
\vspace{-15pt}
\label{fig:frozen-context}
\end{figure*}

\phantomsection\label{sec:ablations}
\noindent\textbf{Ablation studies.}
On the same ALFWorld stream in Table~\ref{tab:ablations}a, joint serving achieves $63.17\%$ success, compared with $59.83\%$ for memory-only and $50.50\%$ for skill-only serving.
Memory-only retains memory writing, fact retrieval, and fact return but disables skills.
Skill-only serving retains internal memory for skill learning while withholding memory from the actor's context.
Each configuration learns from its own trajectories.
Allowing skill revision adds $1.17$ points over the $62.00\%$ score obtained with admitted skill text frozen.
The no-revision control still creates and serves skills and updates their utilities, so this comparison tests the value of continued procedure revision.

Outcome return further improves joint serving in Table~\ref{tab:ablations}b.
Enabling both channels raises success by $3.50$ points over disabling both, with the same content-construction and retrieval rules.
Memory writing and skill consolidation continue with both channels off.
The gain therefore supports adapting future use in addition to learning record content.
For $a,b\in\{0,1\}$, let $B_{a,b}$ enable skill return when $a=1$ and fact return when $b=1$.
The interaction between the two return channels is
\begin{equation}
\begin{aligned}
J_{ab}&=\mathbb E[\preq\mid B_{a,b}],\\
\gamma&=J_{11}-J_{10}-J_{01}+J_{00}.
\end{aligned}
\label{eq:interaction}
\end{equation}
The observed interaction is $+5.33$ points: skill return changes from a $-2.50$-point effect without fact return to a $+2.83$-point effect with it.
This dependence supports jointly adapting the retrieval influence of facts and skills when constructing future context.
See Appendix~\ref{app:factorial} for factorial details and Appendix~\ref{app:efficiency} for actor-call efficiency and MBPP context results.

\phantomsection\label{sec:state-dynamics}
\noindent\textbf{Skill revision continues after catalog growth stops.}
Figure~\ref{fig:coevolution} records 154 accepted revisions over the 600-task stream, with all five final skills present by cycle 2.
The skills are served 496 times, including 338 deployments on successful tasks.
For example, the earlier lamp-examination skill in Figure~\ref{fig:skill-revision-example} accepts lamp visibility or object acquisition as its trigger.
The later version requires both lamp visibility and possession of the object.
This tighter precondition illustrates how episodes from skill-guided behavior can refine the procedure supplied in later contexts under the same skill identity.
See Appendices~\ref{app:state-dynamics} and~\ref{app:qualitative-examples} for lifecycle analyses and further saved examples.

\phantomsection\label{sec:frozen-context}
\noindent\textbf{Relevant selection improves success with a frozen state.}
To separate context use from online learning, we freeze one learned state and compare three serving conditions on the same 100 held-out ALFWorld tasks.
Figure~\ref{fig:frozen-context} shows $75\%$ success for \method{}, versus $58\%$ with approximately length-matched random records.
Random selection also matches record types, so this contrast tests which existing facts and skills enter context.
Keeping the facts fixed and replacing skills with retrieved episodes yields $72\%$.
The skill-based context uses about $62\%$ fewer injected tokens and $29\%$ fewer actor input tokens than episode replacement.
This second contrast supports representing recurring experience as compact procedures, without establishing an accuracy advantage over episodic context.
Appendix~\ref{app:frozen-context} explains the protocol.

% END INLINED FROM sections/06_results.tex
% END INLINED FROM sections/05_results.tex
% BEGIN INLINED FROM sections/06_conclusion.tex
% BEGIN INLINED FROM sections/08_limitations.tex
\section{Conclusion}

\method{} adapts a frozen LLM by learning the evidence and procedures supplied in its bounded context.
Recurring episodes support skill creation and revision, while skills and retrieved facts guide behavior that supplies new memory for subsequent learning.
Outcome return separately adjusts which facts and skills are favored in later retrieval.
Across the evaluated settings, benefits are strongest in conversation and mixed in code.
ALFWorld ablations support joint serving and show interaction between return channels.
Frozen-state controls show higher success with relevant record selection and fewer injected tokens with skills than with episodic context.
Together, these results support bidirectional memory--skill learning as a way to turn accumulated experience into useful context without updating model parameters.

\section*{Limitations}
Outcome return assigns the same task outcome to every served fact and skill, so it does not identify individual causal contributions or estimate factual truth.
It relies on an observed success signal, and robustness to erroneous feedback remains untested.
The state and co-serving analyses are descriptive.
The LM-based skill audit screens proposals against selected episodes but does not guarantee correctness.
Our interleaved evaluation uses domain-scoped retrieval and does not establish transfer of learned skills to unseen domains.

\section*{Ethical Considerations}
This study evaluates agents on existing benchmarks and does not recruit human participants or release a new foundation model or scraped dataset.
Persistent records and provenance can help users inspect adaptation and diagnose repeated errors, but they can also retain sensitive or false information.
Repeated retrieval may amplify biases in either stored content or outcome feedback.
Because the return mechanism rewards task success, it should not be interpreted as validating a claim's accuracy or an action's appropriateness.

Deployments with user data should separate stores by user and domain, restrict access, and provide retention and deletion controls.
Users should be able to inspect and contest stored information, with human review before consequential actions.
These deployment safeguards require separate validation, and our benchmark results do not establish real-world safety.
% END INLINED FROM sections/08_limitations.tex
% END INLINED FROM sections/06_conclusion.tex

\bibliography{references}

\clearpage
\appendix
% BEGIN INLINED FROM appendix.tex
\section{Formal and algorithmic details}
\label{app:formal}

This appendix specifies the records and updates that turn stored experience into context for later decisions.
It follows the online loop from retrieval through outcome return and memory writing to skill consolidation, then derives the properties of the outcome statistics used in retrieval.

\subsection{Notation}
\label{app:notation}

\begin{table*}[t]
\centering
\small
\setlength{\tabcolsep}{3pt}
\begin{tabular}{@{}p{.14\linewidth}p{.32\linewidth}p{.14\linewidth}p{.32\linewidth}@{}}
\toprule
Symbol & Meaning & Symbol & Meaning \\
\midrule
$t,T$ & event index and stream horizon
& $q_t,m_t,\iota_t,\mathcal I_T$ & event, mode, scoring flag, and scored set \\
$\pi_{\theta_0},\tau_t$ & frozen policy and produced trajectory
& $o_t=(r_t,y_t)$ & reward and evaluator-defined success \\
$\mathcal E_t,\mathcal F_t,\mathcal S_t$ & episode store, fact store, and skill catalog
& $\Theta_t$ & disjoint union of the three content stores \\
$\rho_t$ & mutable retrieval and outcome statistics
& $\Xi_t$ & behaviorally active external state \\
$\mathcal A_s$ & supporting episode IDs
& $e,f,s$ & one episode, fact, and skill record \\
\shortstack[l]{$\delta,\ell,\varphi,z,$\\$\chi,\psi,\varepsilon$} & diagnosis, lesson, claim, type, applicability, procedure, and scope
& $P_f,a_e$ & fact provenance and episode lifecycle state \\
$R,G,B$ & retrieval, content update, and outcome return operators
& $g_S,g_F$ & skill and fact return switches \\
$\widehat{\mathcal U}_t,x_{tv}$ & served records and exposure indicator
& $I_t,K_X,L_t$ & actor context and channel/token budgets \\
$c_f,n_f,h_f,C_f$ & fact support, exposures, successes, and outcome score
& $U_s,\alpha_S$ & skill utility and return step size \\
$w_e$ & episode retrieval weight
& $\phi,\eta$ & embedding and merge threshold \\
$\mathcal C_t^{\rm cand},A_C$ & candidate clusters and audit score
& $\preq,\term$ & prequential and terminal metrics \\
$J_{ab},\gamma$ & factorial cell mean and interaction
& $\mathcal L,\mathcal H_t^-$ & sleep boundaries and information available before a decision \\
\bottomrule
\end{tabular}
\caption{Notation used throughout the paper. The index $X\in\{E,F,S\}$ identifies one of the three stores.}
\label{tab:notation}
\end{table*}

\subsection{Record schemas and update ownership}
\label{app:records}

The episode and fact stores hold episodic and semantic memory, respectively, while the skill catalog holds reusable procedures.
The record fields that govern behavior are
\begin{equation}
\begin{aligned}
e&=\langle q_e,\delta_e,\ell_e,\varepsilon_e,a_e\rangle,\\
f&=\langle \operatorname{ent}_f,\varphi_f,z_f,c_f,P_f,\varepsilon_f\rangle,\\
s&=\langle \operatorname{name}_s,\operatorname{desc}_s,\chi_s,\psi_s,\\
 &\qquad\mathcal A_s,\varepsilon_s\rangle.
\end{aligned}
\label{eq:record-payloads}
\end{equation}
For an episode, $q_e$ is task evidence, $\delta_e$ is the outcome diagnosis, and $\ell_e$ is the reusable lesson.
The lifecycle field $a_e\in\{\mathrm{active},\mathrm{archived}\}$ controls ordinary wake retrieval.
For a fact, $\operatorname{ent}_f$, $\varphi_f$, and $z_f$ are its entity, claim, and type.
The field $c_f$ records recurrence support and $P_f$ records provenance from source events.
For a skill, $\chi_s$ contains its applicability conditions and capability key, $\psi_s$ is its procedure, and $\mathcal A_s$ contains the IDs of its accepted supporting episodes.
Each record carries a domain or stream scope stamp $\varepsilon_v$.
The stored records also include identifiers, timestamps, and cached embeddings, which do not appear as evidence in the policy input.
The runner saves diagnostic skill usage counts, fact base rate accumulators, and links between facts and skills served together.
These diagnostic fields lie outside $\Xi_t$ for the transition defined here.
Of the statistics updated by outcome return, retrieval $R$ uses only the fact counts $n_f$ and $h_f$ and the skill utility $U_s$.

The metadata in $\rho_t$ is keyed by record identity:
\begin{equation}
\begin{aligned}
\rho_t&=
\bigl(\{w_e\}_{e\in\mathcal E_t},
      \{n_f,h_f\}_{f\in\mathcal F_t},\\
&\qquad\{U_s\}_{s\in\mathcal S_t}\bigr),\\
C_f&=\frac{h_f+1}{n_f+2}.
\end{aligned}
\label{eq:influence}
\end{equation}
The fact counts determine the outcome score $C_f$.
Retrieval $R$ reads the state without changing it.
Outcome return $B$ changes $(n_f,h_f)$ and $U_s$ for exposed records while leaving payloads, store membership, and episode weights fixed.
Memory writing and skill consolidation belong to the content update operator $G$.
This operator may add or merge payloads, change support and provenance, admit or revise skills, and archive the source episodes of accepted proposals.
It carries the metadata produced by $B$ into the next state so that retained records keep their use history.

The empty state condition in Eq.~\ref{eq:preq} expands to
$\mathcal E_1=\mathcal F_1=\mathcal S_1=\varnothing$ and $\rho_1=\varnothing$.
An episode from an input event starts with $w_e=1$, while a transition episode uses Eq.~\ref{eq:episodeevolve}.
A new fact starts with $c_f=1$, $P_f=\{t\}$, and $n_f=h_f=0$, giving $C_f=.5$.
A new skill stores the IDs of every episode in its proposal's full source cluster in $\mathcal A_s$ and initializes $U_s$ as specified in Section~\ref{app:crystallize}.
An accepted revision preserves the skill identifier and its accumulated utility, combines the old and new ancestry, and replaces only the audited skill content.

\subsection{Full online algorithm and causal chronology}
\label{app:full-loop}

Learning changes subsequent contexts rather than the context that produced the current outcome.
To make this chronology explicit, define
\[
\mathcal H_t^- = \sigma(q_t,\{q_i,\tau_i,o_i,\upsilon_i\}_{i<t})
\]
as the information available just before event $t$ is processed.
The variable $\upsilon_i$ contains the randomness used for extraction, clustering, synthesis, and audit at event $i$.
For input events, $\tau_i=o_i=\varnothing$.
Both $\Xi_t$ and $\widehat{\mathcal U}_t$ are measurable with respect to $\mathcal H_t^-$.
The adapter sets $\iota_t=0$ for input events, so
$\mathcal I_T=\{t:m_t=\mathrm{transition},\ \iota_t=1\}$.
We write $G_{\rm write}$ for the memory writing stage of $G$, which runs after each event and before any scheduled skill consolidation.

\begin{figure*}[t]
\begin{paperalgorithm}{Full \method{} online loop}
\label{alg:full-loop}
\algline{\textbf{Input:} an ordered stream $q_{1:T}$, a declared adapter, a frozen policy $\pi_{\theta_0}$, return switches $g_S$ and $g_F$, and sleep boundaries $\mathcal L$. The stream must contain at least one scored transition.}
\algline{\textbf{Initialize:} $\mathcal E_1=\mathcal F_1=\mathcal S_1=\varnothing$, $\rho_1=\varnothing$, score sum $R_{\rm sum}\leftarrow0$, and scored count $N\leftarrow0$.}
\algline{\textbf{for} $t=1,\ldots,T$ \textbf{do}}
\algline[1.35em]{Parse mode $m_t$, scoring flag $\iota_t$, scope $\varepsilon_t$, and payload from the adapter.}
\algline[1.35em]{\textbf{if} $m_t=\mathrm{input}$ \textbf{then}}
\algline[2.7em]{Require $\iota_t=0$. Obtain $\Xi_t^{\rm wr}$ from $\Xi_t$ by appending the declared episode with $w_e=1$ and sequentially merging the declared facts. Carry metadata for retained identities, initialize new facts with $n_f=h_f=0$, and do not call $\pi_{\theta_0}$ or $B$.}
\algline[1.35em]{\textbf{else} \emph{process a transition event}}
\algline[2.7em]{Retrieve and format $\widehat{\mathcal U}_t$ into $I_t$ under the adapter's context limits. Count as exposed only records represented in $I_t$.}
\algline[2.7em]{Sample $\tau_t\sim\pi_{\theta_0}(\cdot\mid q_t,I_t)$ and observe $o_t=(r_t,y_t)$. If $\iota_t=1$, set $R_{\rm sum}\leftarrow R_{\rm sum}+r_t$ and $N\leftarrow N+1$.}
\algline[2.7em]{$\Xi_t^{\rm ret}\leftarrow B_{g_S,g_F}(\Xi_t,o_t,\widehat{\mathcal U}_t)$. Update only served facts and skills without changing record text or membership.}
\algline[2.7em]{$\Xi_t^{\rm wr}\leftarrow G_{\rm write}(\Xi_t^{\rm ret},\tau_t,o_t)$. Append the episode if writing succeeds and merge valid permitted facts. New records receive no return from $o_t$.}
\algline[1.35em]{\textbf{end if}}
\algline[1.35em]{\textbf{if} $t\in\mathcal L$ \textbf{then}}
\algline[2.7em]{Reactivate archived episodes, compute ranked candidate clusters $C_1,\ldots,C_M$ from the resulting pool at sleep time using Eqs.~\ref{eq:clusters} and \ref{eq:cluster-rank}, and set the working catalog $\mathcal S\leftarrow\mathcal S_t$.}
\algline[2.7em]{\textbf{for} $j=1,\ldots,M$ \textbf{do}}
\algline[4.05em]{Synthesize skill proposal $\widetilde s_{C_j}$ from up to five episodes in $C_j$ with the highest weights. Audit the proposal in a fresh context and set $u\leftarrow0$.}
\algline[4.05em]{\textbf{if} the first audit accepts \textbf{then} set $s^\star\leftarrow\operatorname{Match}(\widetilde s_{C_j},\mathcal S)$.}
\algline[5.4em]{\textbf{if} $s^\star=\varnothing$ \textbf{then} consult the declared duplicate-slug guard. If it permits insertion, set $\mathcal S\leftarrow\mathcal S\cup\{\operatorname{Insert}(\widetilde s_{C_j},C_j)\}$. Set $u\leftarrow1$ whether insertion occurs or the duplicate-slug guard intercepts it.}
\algline[5.4em]{\textbf{else} retrieve the episodes indexed by $\mathcal A_{s^\star}$, pool those records with $C_j$, synthesize a revision $\overline s_{C_j}$, and audit it in a fresh context. If accepted, replace $s^\star$ by $\operatorname{Replace}(s^\star,\overline s_{C_j},C_j)$ and set $u\leftarrow1$.}
\algline[5.4em]{\textbf{end if}}
\algline[4.05em]{\textbf{end if}}
\algline[4.05em]{\textbf{if} $u=1$ \textbf{then} archive the episodes in the current cluster $C_j$ from ordinary wake retrieval.}
\algline[2.7em]{\textbf{end for}}
\algline[2.7em]{Set $\Xi_{t+1}$ to the state after writing with the updated working catalog and lifecycle flags.}
\algline[1.35em]{\textbf{else} set $\Xi_{t+1}\leftarrow\Xi_t^{\rm wr}$.}
\algline[1.35em]{\textbf{end if}}
\algline{\textbf{end for}}
\algline{Set $\preq\leftarrow R_{\rm sum}/N$ and \textbf{return} $(\Xi_{T+1},\preq)$. For $\term$, hold $\Xi_{T+1}$ fixed, run retrieval, the policy, and the task evaluator, and do not call $B$, $G$, or sleep updates.}
\end{paperalgorithm}
\end{figure*}

Algorithm~\ref{alg:full-loop} applies outcome return before memory writing and consolidates skills at scheduled sleep boundaries.
The runner appends the episode, applies return to previously served facts, writes new facts, and then applies return to previously served skills.
This order produces the same state as $B$ followed by $G_{\rm write}$: outcome return does not update episodes, fact return precedes new fact insertion, and skill return does not change memory content.
Skill consolidation follows all four substeps at a sleep boundary.
Because the action context is fixed before $o_t$ arrives, these updates influence only later decisions.

Memory writing preserves the interaction for later retrieval and skill learning.
For transition events, the main configuration extracts facts only when $y_t=1$, though declared adapters may also retain facts from failed trajectories.
For a valid episode returned by $\mathcal W_E$, the memory write in Eq.~\ref{eq:forwardwrite} is
\begin{align}
&\begin{aligned}
e_t&=\mathcal W_E(\tau_t,o_t),\\
\mathcal E_t^{\rm wr}&=\mathcal E_t\cup\{e_t\},\\
w_{e_t}&=1+.3(2y_t-1),
\end{aligned} \label{eq:episodeevolve}\\
&\begin{aligned}
Z_t&=\mathcal X_F(\tau_t,o_t),\\
\mathcal F_t^{\rm wr}&=\operatorname{Merge}_{.86}(\mathcal F_t,Z_t).
\end{aligned}
\label{eq:factevolve}
\end{align}
If the episode call or parse fails, the episode store remains $\mathcal E_t^{\rm wr}=\mathcal E_t$, but fact extraction and merging still follow Eq.~\ref{eq:factevolve}.
The stores $\mathcal E_t^{\rm wr}$ and $\mathcal F_t^{\rm wr}$ therefore describe the state after writing and before sleep.
The corresponding metadata $\rho_t^{\rm wr}$ extends $\rho_t^{\rm ret}$ to include the newly written records.
It assigns the weight above to each successfully written episode and initializes new facts with zero exposures and successes.

Fact merging increases recurrence support without treating recurrence as an exposed success.
Write $Z_t=(\widetilde f_1,\ldots,\widetilde f_{|Z_t|})$ in extraction order and set a working store $\mathcal F\leftarrow\mathcal F_t$.
$\operatorname{Merge}_\eta$ processes extracted facts sequentially, allowing later facts from the same event to match records inserted or updated earlier in that event.
For any fact $f$, let $\operatorname{ent}(f)$ and $\varphi(f)$ denote its entity and claim.
For each extracted fact $\widetilde f$, it scans the working store in stable order and takes the first record with the same entity and scope whose similarity meets the threshold:
\begin{equation}
\begin{aligned}
f^\star&=\operatorname{first}\!\bigl\{f\in\mathcal F:\\
&\quad\operatorname{ent}(f)=\operatorname{ent}(\widetilde f),\\
&\quad\varepsilon_f=\varepsilon_{\widetilde f},\\
&\quad\cos\!\bigl(\phi(\operatorname{ent}(f)\!:\!\varphi(f)),\\
&\qquad\phi(\operatorname{ent}(\widetilde f)\!:\!\varphi(\widetilde f))\bigr)\ge\eta\bigr\}.
\end{aligned}
\label{eq:factmatch}
\end{equation}
When $f^\star$ exists, $\operatorname{Merge}_\eta$ applies
\begin{equation}
(c_{f^\star},P_{f^\star})
\leftarrow(c_{f^\star}+1,\ P_{f^\star}\cup\{t\}).
\label{eq:factmerge}
\end{equation}
Otherwise, it appends $\widetilde f$ to $\mathcal F$ with $c_{\widetilde f}=1$, $n_{\widetilde f}=h_{\widetilde f}=0$, and provenance $P_{\widetilde f}=\{t\}$.
After all extracted facts have been processed, $\mathcal F_t^{\rm wr}=\mathcal F$.

\subsection{Skill mining, synthesis, and admission}
\label{app:crystallize}

Skill consolidation turns recurring episodes into reusable procedures at scheduled sleep boundaries.
Because it follows the current memory write, its evidence may include the newly completed episode.
Before each sleep pass, the miner reactivates archived episodes and includes them in the pool.
It clusters their embeddings and reads the weights $w_e$ stored in $\Xi_t^{\rm wr}$ to score the candidate groups.
HDBSCAN uses Euclidean distance, \texttt{min\_cluster\_size}$=3$, \texttt{min\_samples}$=1$, and excess of mass selection \citep{hdbscan}.
Let $\operatorname{NonNoise}$ drop the HDBSCAN noise group.
For each non-noise group $H$, $\operatorname{FamilySplit}_3(H)$ returns $\{H\}$ when $H$ contains one family or when no family has three members.
Otherwise, it returns each family subgroup with at least three members and drops the rest.
Each returned subgroup inherits the HDBSCAN persistence of its parent $H$.
The $\operatorname{Mine}$ operator in Eq.~\ref{eq:crystallize} uses
\begin{equation}
\begin{aligned}
Q(C)&=\frac1{|C|}\sum_{e\in C}\operatorname{clip}(w_e/1.5,0,1),\\
\mathcal R_t
&=\operatorname{NonNoise}\!\bigl(\\
&\qquad\operatorname{HDBSCAN}(\phi(\mathcal E_t^{\rm wr}))\bigr),\\
\mathcal R_t^{\rm qual}&=\left\{H\in\mathcal R_t:Q(H)\ge.4\right\},\\
\mathcal D_t&=\bigcup_{H\in\mathcal R_t^{\rm qual}}
\operatorname{FamilySplit}_3(H),\\
\mathcal C_t^{\rm cand}
&=\operatorname{Top}^{d}_{10}\!\bigl(\\
&\quad\left\{C\in\mathcal D_t:|C|\ge3,\ Q(C)\ge.4\right\}\bigr).
\end{aligned}
\label{eq:clusters}
\end{equation}
Let $p(C)$ be the HDBSCAN persistence of a cluster.
The ranking used by $\operatorname{Top}^{d}_{10}$ in Eq.~\ref{eq:clusters} is
\begin{equation}
d(C)=Q(C)|C|\max\{.05,p(C)\}.
\label{eq:cluster-rank}
\end{equation}
At most ten ranked candidate clusters are processed per sleep boundary.
Under the main episode initialization, even a failure episode has $w_e/1.5>.4$, so this quality threshold does not exclude failure-only clusters.
The quality score instead helps rank candidate clusters and sets the initial utility of an admitted skill.

A separate frozen-LM call synthesizes a structured skill proposal from up to five members of $C$ with the largest $w_e$.
The parser requires a name and description, and it rejects a proposal whose procedure contains no parameter slot.
The stored ancestry $\mathcal A_s$ contains the IDs of all episodes in the accepted proposal's source cluster, including those omitted from the synthesis prompt.
The audit calls the same frozen LM in a fresh context containing the complete skill proposal, cluster statistics, and the selected episodes.
Its prompt requests one verdict for each of the 13 checks in Table~\ref{tab:audit-checks}, and the admission rule counts the returned \texttt{PASS} verdicts.
Let $\widehat m_C$ be the number of valid dictionary objects in the returned checks, with $a_{Ck}=1$ exactly when the parser reads \texttt{PASS} for returned object $k$.
The implemented score is
\begin{equation}
A_C=\sum_{k=1}^{\widehat m_C}a_{Ck}.
\label{eq:audit-score}
\end{equation}
The parser skips entries that are not objects and rejects unparseable JSON.
Unrecognized verdicts count as failures, and a skill proposal passes the first screen exactly when $A_C\ge10$.
Candidate clusters are processed in the fixed ranking order.
A matching capability still requires revision synthesis and a second audit before the catalog changes.
For a new capability, the implementation initializes
$\mathcal A_{\widetilde s_C}=\{\operatorname{id}(e):e\in C\}$ and
$U_{\widetilde s_C}=\operatorname{clip}(Q(C),.4,.95)$.
The audit does not treat affordances fixed by the domain as instance literals.

\begin{table*}[t]
\centering
\small
\setlength{\tabcolsep}{3pt}
\begin{tabular}{@{}cp{.21\linewidth}p{.69\linewidth}@{}}
\toprule
& Check & Passing condition \\
\midrule
1 & Instance literals & No names, identifiers, or varying numeric values from a single episode are baked into the procedure. \\
2 & Borrowed framing & The skill is a reusable procedure rather than a paraphrase of one episode. \\
3 & Runtime probes & Relevant preconditions are checked when the action space supplies a probe, or are stated explicitly otherwise. \\
4 & Parametricity & Values that vary across uses appear as parameter slots. \\
5 & Necessary setup & The procedure contains no setup step that cannot contribute to the task. \\
6 & Workflow integrity & No declared tool call or required stage is silently bypassed. \\
7 & Traceability & Correctness claims are grounded in observable state. \\
8 & Stable shape & The procedure does not assume incompatible concrete shapes across uses. \\
9 & Cluster coverage & The procedure applies to a majority of the source cluster. \\
10 & Failure coverage & It addresses at least half of the nonempty recorded failures or improvements, or passes vacuously when none exist. \\
11 & Scope control & It does not claim task shapes unsupported by the cluster. \\
12 & Model invariance & It abstracts style tied to one model family when multiple model families appear and passes when only one family appears. \\
13 & Capability consistency & Verb, target, affordance, and output fields are complete and agree with the procedure. \\
\bottomrule
\end{tabular}
\caption{\textbf{Skill audit checks.} Each skill proposal is evaluated in a fresh context.}
\label{tab:audit-checks}
\end{table*}

Catalog updates accumulate evidence for a capability across sleep passes.
Set the working catalog $\mathcal S\leftarrow\mathcal S_t$ and process the ranked clusters in order, so each cluster sees earlier accepted updates from the same pass.
For candidate cluster $C$, synthesis produces skill proposal $\widetilde s_C$.
After the first audit accepts it, matching searches the working catalog:
\begin{equation}
\begin{aligned}
\widetilde s_C&=\operatorname{Synth}(C),\\
s_C^\star&=\operatorname{Match}(\widetilde s_C,\mathcal S)\\
&\in\mathcal S\cup\{\varnothing\}.
\end{aligned}
\label{eq:skill-proposal}
\end{equation}
The complete capability key is
$(\text{verb family},\allowbreak\text{target type},\allowbreak\text{affordance},\allowbreak\text{output})$.
Matching returns $\varnothing$ when the catalog has no skill with that complete key.
For a match, $G$ pools $C$ with the episodes indexed by the incumbent's ancestry, selects up to five records with the highest weights, and synthesizes revision $\overline s_C$ for a second audit.
The catalog transition is
\begin{equation}
\begin{aligned}
&\mathcal S\leftarrow\\
&\quad
\begin{cases}
\begin{aligned}
&\mathcal S\cup\{\operatorname{Insert}(\widetilde s_C,C)\},\\
&\quad\text{accepted new skill},
\end{aligned}\\[3pt]
\begin{aligned}
&\bigl(\mathcal S\setminus\{s_C^\star\}\bigr)\cup\\
&\quad\{\operatorname{Replace}(s_C^\star,\overline s_C,C)\},\\
&\quad\text{accepted revision},
\end{aligned}\\[3pt]
\begin{aligned}
&\mathcal S,\\
&\quad\text{otherwise}.
\end{aligned}
\end{cases}
\end{aligned}
\label{eq:skill-transition}
\end{equation}
Insertion creates a fresh identity whose ancestry contains every episode in $C$.
Replacement regenerates the incumbent's text and capability fields while preserving its identifier, creation time, and utility after outcome return.
It also combines the old and new ancestry, so evidence for an existing capability supports revision under the same skill identity.
An accepted insertion or revision archives the current cluster from ordinary wake retrieval.
Some adapters also apply a duplicate-slug guard that archives the cluster even when it intercepts an insertion and leaves the catalog unchanged.
These episodes remain stored for provenance and are reactivated at the next sleep boundary.

After the ordered pass, $\mathcal S_{t+1}$ is the working catalog and $\mathcal E_{t+1}$ includes the resulting lifecycle flags.
Skill consolidation leaves the fact store at $\mathcal F_{t+1}=\mathcal F_t^{\rm wr}$.
The completed state is
\begin{equation}
\begin{aligned}
\Theta_{t+1}
&=\mathcal E_{t+1}\uplus\mathcal F_{t+1}\uplus\mathcal S_{t+1},\\
\Xi_{t+1}&=(\Theta_{t+1},\rho_{t+1}).
\end{aligned}
\label{eq:next-state}
\end{equation}
For retained records, $\rho_{t+1}$ carries forward $\rho_t^{\rm wr}$, including the updated $(n_f,h_f)$ for merged facts and $U_s$ for revised skills.
New skill identities receive their cluster-based initial utility.
Away from a sleep boundary, the catalog remains $\mathcal S_t$ and $\Xi_{t+1}=\Xi_t^{\rm wr}$.

\subsection{Exact retrieval scores and exposure}
\label{app:retrieval}

Retrieval determines which parts of the evolving state can affect the next decision.
The retriever first checks scope and applicability, then ranks eligible records by relevance and retrieval influence.
An episode is eligible for ordinary wake retrieval only when it is active and scope compatible.
Archived episodes remain stored for provenance and later mining.
A fact must satisfy the adapter's scope, type, and minimum support filters.
For skills, it extracts the four-field capability key defined in Section~\ref{app:crystallize} and first searches for an exact match.
If none exists, the main capability route relaxes the affordance requirement and then both the verb and affordance requirements.
This route does not fall back to global cosine retrieval.
All routing choices and thresholds are fixed before a stream begins, and $K_X=0$ disables channel $X$.

Let $\mathcal O$ be the channel order declared by the adapter and let $\varepsilon(q_t)$ be the query scope.
$\operatorname{Eligible}_X(q_t,\Theta_t)$ denotes the records from channel $X$ that pass the preceding constraints.
$\operatorname{TopK}_{K_X}$ returns the $K_X$ records with the highest scores.
The serving rule is
\begin{align}
&\begin{aligned}
\mathcal V_t&=\operatorname{Concat}_{X\in\mathcal O}
\operatorname{TopK}_{K_X}
\bigl(\\
&\qquad\operatorname{Eligible}_X(q_t,\Theta_t),\\
&\qquad R_X(\cdot,q_t)\bigr),\\
\widehat{\mathcal U}_t
&=\operatorname{Pack}_{L_t,\varepsilon(q_t)}(\mathcal V_t),\\
I_t
&=\operatorname{Fmt}_{\varepsilon(q_t)}(\widehat{\mathcal U}_t).
\end{aligned}
\label{eq:serve}
\end{align}
Here $\operatorname{Pack}$ denotes context assembly under the adapter's limits, using the same rendering rule as $I_t$.
The online actor path uses retrieval quotas and field-length limits, while the frozen-state controls enforce an explicit token budget $L_t$.

In the retrieval scores below, $q_e$, $\delta_e$, and $\ell_e$ denote an episode's task evidence, outcome diagnosis, and lesson.
For a fact $f$, $\operatorname{ent}(f)$ and $\varphi(f)$ denote its entity and claim.
The fields $\mathrm{name}_s$ and $\mathrm{desc}_s$ form the skill retrieval text.
The operator $\oplus$ denotes text concatenation.
The channel scores used in Eq.~\ref{eq:serve} are
\begin{align}
&\begin{aligned}
R_S(s,q)&=\max(.05,U_s)\,\\
&\quad
 \cos\!\left(\phi(q),\phi(\mathrm{name}_s\!\oplus\!\mathrm{desc}_s)\right),
\end{aligned}
 \label{eq:skillscore}\\
&\begin{aligned}
C_f&=\frac{h_f+1}{n_f+2},\\
R_F(f,q)&=\cos\!\bigl(\phi(q),\\
&\qquad\phi(\operatorname{ent}(f)\!:\!\varphi(f))\bigr)\\
&\quad
 \left(1+.05\min(c_f,6)\right)(.5+C_f),
\end{aligned}
 \label{eq:factscore}\\
&\begin{aligned}
R_E(e,q)&=w_e\cos\!\bigl(\phi(q),\\
&\qquad\phi(q_e\!\oplus\!\delta_e\!\oplus\!\ell_e)\bigr).
\end{aligned}
 \label{eq:epscore}
\end{align}
These equations reproduce the implemented signed cosine products.
Their multiplicative influence factors increase a score when its cosine relevance is nonnegative.
When cosine relevance is negative, increasing the factor makes the signed score more negative.
Before exact or relaxed lookup, the skill router maps the verb, target, affordance, and output to their predefined canonical values.
The fact filter checks type, support, and the optional domain stamp.
For dialogue, define
\[
\begin{aligned}
b_{\rm lex}(v,q)&=.12\min\bigl(\\
&\quad|\operatorname{Tok}(q)\cap\operatorname{Tok}(v)|,4\bigr)
\end{aligned}
\]
using lowercased whitespace tokens.
The episode's dense retrieval text is
$v_e^{\rm den}=q_e\!\oplus\!\delta_e\!\oplus\!\ell_e$.
The lexical retrieval texts concatenate fields with spaces:
$v_f^{\rm lex}=\operatorname{ent}(f)\!\oplus\!\varphi(f)$ and
$v_e^{\rm lex}=q_e\!\oplus\!\ell_e$.
The dialogue retriever adds this lexical anchor to the dense score:
\begin{align}
&\begin{aligned}
R_F^{\rm dlg}(f,q)&=R_F(f,q)\\
&\quad+b_{\rm lex}(v_f^{\rm lex},q),\\
R_E^{\rm dlg}(e,q)
&=w_e\!\bigl(\cos(\phi(q),\phi(v_e^{\rm den}))\\
&\qquad+b_{\rm lex}(v_e^{\rm lex},q)\bigr).
\end{aligned}
\label{eq:lexical-bonus}
\end{align}
The lexical bonus helps retrieve names and dates alongside dense semantic matches.
The context formatter renders an episode as $(q_e,\delta_e,\ell_e)$.
Facts contribute their entity, claim, and support, while skills contribute their name, description, applicability, and procedure.
\method{}'s actor formatter limits skill applicability text to 300 characters and procedure text to 600 characters before assembly, so the actor may receive only a prefix of a stored procedure.
The metadata $C_f$ and $U_s$ affect ranking but are not presented as textual evidence.
Context assembly preserves the adapter's channel order in $\mathcal V_t$.
Thus $\widehat{\mathcal U}_t$ contains exactly the records represented in $I_t$, while any omitted candidate has $x_{tv}=0$ and receives no return.

\subsection{Outcome return properties}
\label{app:return-properties}

The fact and skill statistics differ in how they weight past exposures.
To derive their behavior and bounds, we first write the return updates for an existing record with both binary switches explicit:
\begin{align}
&\begin{aligned}
U_s^{\rm ret}
&=U_s+g_S\alpha_Sx_{ts}(y_t-U_s),\\
n_f^{\rm ret}
&=n_f+g_Fx_{tf},\\
h_f^{\rm ret}
&=h_f+g_Fx_{tf}y_t,
\end{aligned}
\label{eq:switched-return}
\end{align}
where $(g_S,g_F)\in\{0,1\}^2$.

The reported implementation uses $\alpha_S=.2$.
Let $y_{s,1},\ldots,y_{s,m}$ be the outcomes from the first $m\ge1$ tasks that expose skill $s$.
For $g_S=1$ and $0<\alpha_S<1$, unrolling Eq.~\ref{eq:skillreturn} gives
\begin{equation}
\begin{aligned}
U_s^{(m)}&=(1-\alpha_S)^mU_s^{(0)}\\
&\quad+\alpha_S\sum_{i=1}^{m}(1-\alpha_S)^{m-i}y_{s,i}.
\end{aligned}
\label{eq:skill-closed-form}
\end{equation}
For $0<\alpha_S<1$, the displayed coefficients are nonnegative and sum to one, so $U_s^{(m)}\in[0,1]$ whenever $U_s^{(0)},y_{s,i}\in[0,1]$.
The endpoint $\alpha_S=1$ follows directly from applying the recurrence once.
When $g_S=0$, $B$ leaves the utility of every existing skill unchanged.
The constant step size gives more weight to recent exposures while retaining part of the cluster-based initialization.
Because revisions preserve identity and utility, this history remains associated with the skill as its text changes.

For $g_F=1$, an exchangeable Bernoulli working model treats outcomes on tasks that expose fact $f$ as conditionally independent with a common success probability $p_f$.
A $\operatorname{Beta}(1,1)$ prior on $p_f$ smooths $h_f$ successes from $n_f$ exposures:
\begin{equation}
\begin{aligned}
p_f\mid h_f,n_f&\sim
\operatorname{Beta}(h_f+1,n_f-h_f+1),\\
\mathbb E[p_f\mid h_f,n_f]&=\frac{h_f+1}{n_f+2}=C_f.
\end{aligned}
\label{eq:fact-posterior}
\end{equation}
Under this working model, $C_f$ estimates smoothed success conditional on exposure, not factual truth.
The interpretation explains the smoothing rule, although adaptive retrieval and changing skills can alter the exposed-task distribution during deployment.
When $x_{tf}=1$, one more exposed outcome $y_t$ gives the recursion
\begin{equation}
C_f^{\rm ret}-C_f=\frac{y_t-C_f}{n_f+3}.
\label{eq:fact-recursion}
\end{equation}
When $g_F=0$, both counts and the derived score remain unchanged.
The fact's step size decreases with accumulated exposures, whereas the skill's fixed step size continues to emphasize recent use.
Equations~\ref{eq:skill-closed-form} and~\ref{eq:fact-recursion} bound both scores in $[0,1]$ without requiring the Bernoulli working model.
Because all exposed records share the task outcome, neither statistic identifies an individual record's causal contribution.
Retrieval uses these summaries to adjust which stored facts and procedures enter later contexts.

\subsection{Interpretation of the outcome-return factorial}
\label{app:factorial}

The four cell means in Eq.~\ref{eq:interaction} determine the saturated expansion
\begin{equation}
\begin{aligned}
J_{ab}&=\mu+\alpha a+\beta b+\gamma ab,\\
\mu&=J_{00},\\
\alpha&=J_{10}-J_{00},\\
\beta&=J_{01}-J_{00},\\
\gamma&=J_{11}-J_{10}-J_{01}+J_{00}.
\end{aligned}
\label{eq:factorial-expansion}
\end{equation}
Under additivity, $\gamma=0$.
The observed success counts give the interaction in percentage points:
\begin{equation}
\begin{aligned}
100\widehat\gamma
&=100\,\frac{379-343-362+358}{600}\\
&\approx 5.33\;\text{pp}.
\end{aligned}
\label{eq:interaction-result}
\end{equation}
Relative to both channels enabled, disabling fact return lowers success by $6.00$ points, disabling skill return by $2.83$, and disabling both by $3.50$.

The interaction measures how the effect of skill return changes when fact return is enabled, or equivalently how the effect of fact return changes when skill return is enabled.
All four cells use the same content update algorithm $G$ and retrieval rules, differing only in the return switches.
Because each run develops its own trajectories and state, the contrast includes the interventions' downstream effects on context and learning.
The joint-serving ablation separately tests the value of supplying memory and skills together.

\section{Additional results and analyses}
\label{app:additional-results}

These analyses examine how deployment conditions affect performance, how learned records are reused, and how much actor effort they require.

\noindent\textbf{ALFWorld navigation adaptation.}
On the same 600-task ALFWorld stream with frozen Qwen3-8B and a 30-action limit, \method{} achieves $63.17\%$ success with the revisit guard and $57.00\%$ without it.
The navigation hint and all other settings remain fixed.
Appendix~\ref{app:evaluation-config} describes the guard's task-local navigation rule.

\subsection{ALFWorld performance across shuffled task orders}
\label{app:order-robustness}

\noindent\textbf{Performance across orders.}
Across the four shuffled task orders in Table~\ref{tab:order-robustness}, \method{} achieves success rates from $57.50\%$ to $63.83\%$, with a mean of $61.08\%$ and a sample standard deviation of $2.96$ percentage points.
One run exceeds the original fixed-order score of $63.17\%$, one matches it, and two score lower.
The lowest shuffled-order score remains above the fixed-order references of $52.70\%$ for Reflexion in Table~\ref{tab:alfworld-additional} and $56.17\%$ for guarded ReAct.

\noindent\textbf{Settings.}
We reconstruct the original 600-game pool from ALFWorld's train ordering and apply four global permutations with NumPy \citep{numpy} \texttt{RandomState} seeds 101, 202, 303, and 404.
Each run starts with an empty episode store, fact store, and skill catalog and scores every task before writeback.
Skill consolidation follows each block of 30 tasks for 20 cycles.
The Full configuration uses frozen Qwen3-8B, at most 30 actions per task, and retrieval of up to four facts and two skills.
Both outcome-return channels, skill creation and revision, the revisit guard, and the navigation hint remain enabled.
Episodic records support skill learning but do not enter the actor's context in this configuration.
The recorded game files match each prescribed permutation exactly, with 600 distinct games and no missing tasks or environment exceptions.
The four shuffled runs share one code snapshot and configuration.
We report the original fixed-order result separately.

\begin{table*}[t]
\centering
\small
\setlength{\tabcolsep}{0pt}
\renewcommand{\arraystretch}{1.08}
% Fixed-width cells keep all eight columns inside the text block.
\newcommand{\orderrow}[8]{%
  \tableleftcell{.22\linewidth}{#1}
  & \tablecell{.14\linewidth}{#2}
  & \tablecell{.10\linewidth}{#3}
  & \tablecell{.108\linewidth}{#4}
  & \tablecell{.108\linewidth}{#5}
  & \tablecell{.108\linewidth}{#6}
  & \tablecell{.108\linewidth}{#7}
  & \tablecell{.108\linewidth}{#8} \\}
\begin{tabular}{@{}llllllll@{}}
\toprule
\orderrow{\textbf{Order}}{\textbf{Successes}}{\textbf{SR\%}}{\textbf{Clean}}{\textbf{Cool}}{\textbf{Examine}}{\textbf{Heat}}{\textbf{Move}}
\midrule
\orderrow{Seed 101}{359/600}{$59.83$}{$60.2$}{$54.3$}{$66.1$}{$62.2$}{$59.6$}
\orderrow{Seed 202}{383/600}{$63.83$}{$62.2$}{$60.9$}{$86.4$}{$62.2$}{$61.0$}
\orderrow{Seed 303}{379/600}{$63.17$}{$61.2$}{$62.0$}{$83.1$}{$62.2$}{$60.3$}
\orderrow{Seed 404}{345/600}{$57.50$}{$57.1$}{$56.5$}{$64.4$}{$52.7$}{$57.8$}
\midrule
\orderrow{Mean}{}{$61.08$}{$60.2$}{$58.4$}{$75.0$}{$59.8$}{$59.7$}
\orderrow{Sample SD}{}{$2.96$}{$2.2$}{$3.6$}{$11.4$}{$4.7$}{$1.4$}
\midrule
\orderrow{Original fixed order}{379/600}{$63.17$}{$53.1$}{$62.0$}{$81.4$}{$63.5$}{$63.2$}
\bottomrule
\end{tabular}
\caption{\textbf{ALFWorld performance across four shuffled-order runs.}
Scores are percentages, and standard deviations are in percentage points computed over the four new runs only.
Each run contains 98 clean, 92 cool, 59 examine, 74 heat, and 277 move tasks.}
\label{tab:order-robustness}
\end{table*}

\noindent\textbf{Variation across runs.}
Variation is largest for examine, with a standard deviation of $11.4$ points compared with $1.4$--$4.7$ points for the other families.
Clean success exceeds the fixed-order result in every shuffled run, while move remains comparatively stable.
The run-level Student-$t$ 95\% confidence interval for the mean is $[56.37,65.80]\%$, based on four runs and three degrees of freedom.

\subsection{Terminal evaluation}
\label{app:terminal}

\noindent\textbf{Protocol.}
Prequential evaluation measures performance during learning, while terminal evaluation measures performance with the accumulated state after learning stops.
For every benchmark in this evaluation, we freeze the final state and score tasks held out from the learning stream:
\begin{equation}
\term=\mathbb E_{q\sim\mathcal Q_{\rm heldout}}
\left[r(q,\Xi_{T+1})\right].
\label{eq:terminal-metric}
\end{equation}
Table~\ref{tab:terminal} reports held-out performance with the final state alongside the prequential results from Table~\ref{tab:performance}.
The terminal scores are higher on ALFWorld, RULER, and MBPP, but lower on BABILong and LoCoMo.

\begin{table*}[t]
\centering
\small
\setlength{\tabcolsep}{0pt}
\renewcommand{\arraystretch}{1.06}
\begin{tabular}{@{}lllll@{}}
\toprule
\tableleftcell{.19\linewidth}{\textbf{Benchmark}}
& \tableleftcell{.23\linewidth}{\textbf{Metric}}
& \tablecell{.21\linewidth}{\textbf{Terminal score}}
& \tablecell{.23\linewidth}{\textbf{Prequential score}}
& \tablecell{.14\linewidth}{\textbf{$\Delta$}} \\
\midrule
\tableleftcell{.19\linewidth}{ALFWorld}
& \tableleftcell{.23\linewidth}{Success rate}
& \tablecell{.21\linewidth}{$70.00$}
& \tablecell{.23\linewidth}{$63.17$}
& \tablecell{.14\linewidth}{\gain{6.83}} \\
\tableleftcell{.19\linewidth}{RULER}
& \tableleftcell{.23\linewidth}{Recall}
& \tablecell{.21\linewidth}{$65.70$}
& \tablecell{.23\linewidth}{$64.70$}
& \tablecell{.14\linewidth}{\gain{1.00}} \\
\tableleftcell{.19\linewidth}{BABILong}
& \tableleftcell{.23\linewidth}{Mean reward}
& \tablecell{.21\linewidth}{$57.00$}
& \tablecell{.23\linewidth}{$59.00$}
& \tablecell{.14\linewidth}{\drop{2.00}} \\
\tableleftcell{.19\linewidth}{LoCoMo}
& \tableleftcell{.23\linewidth}{Token F1}
& \tablecell{.21\linewidth}{$23.70$}
& \tablecell{.23\linewidth}{$31.75$}
& \tablecell{.14\linewidth}{\drop{8.05}} \\
\tableleftcell{.19\linewidth}{MBPP}
& \tableleftcell{.23\linewidth}{Unit test reward}
& \tablecell{.21\linewidth}{$81.59$}
& \tablecell{.23\linewidth}{$80.42$}
& \tablecell{.14\linewidth}{\gain{1.17}} \\
\bottomrule
\end{tabular}
\caption{\textbf{Terminal and prequential performance of \method{}.}
Scores are percentages, and $\Delta=100(\term-\preq)$ is in percentage points.}
\label{tab:terminal}
\end{table*}

\subsection{Frozen-state context controls}
\label{app:frozen-context}

\noindent\textbf{Serving conditions.}
We evaluate three context conditions on the same 100 held-out ALFWorld games using frozen Qwen3-8B and one learned external state.
The state content and retrieval statistics remain fixed throughout evaluation, with no memory writes, outcome return, or skill consolidation.
All conditions retain the revisit guard, navigation hint, and 30-action limit.
All conditions have a 1,024-token injected-context limit, measured with the backbone tokenizer and including block headings.
\method{} retrieves facts and skills, whereas episode replacement keeps the same facts and fills the remaining budget with retrieved episodes, including archived records.
The replacement serves an average of 10.8 episodes per task.
Random selection samples facts and skills from the same state while matching record types and approximate context length.
Its length is within $10\%$ of \method{} on every task, with a mean absolute relative deviation of $2.6\%$.
All 300 evaluations are valid, the game files match across conditions, and no injected context is truncated.

\noindent\textbf{Paired outcomes and actor cost.}
Compared with episode replacement, \method{} alone solves 12 tasks and the replacement alone solves nine.
The corresponding counts against random selection are 24 and seven.
Two-sided exact McNemar tests give $p=.664$ and $p=.0033$, respectively.
The episode comparison does not establish accuracy equivalence.
Although all conditions share the same limit, their mean injected contexts differ: 365 tokens for \method{}, 963 for episode replacement, and 369 for random selection.
Mean actor input, including retries, is 25.9k, 36.6k, and 35.2k tokens per task in the same order.
These costs cover serving the frozen state but not constructing it.
This experiment is separate from both the online prequential scores and the terminal evaluation in Table~\ref{tab:terminal}.

\subsection{Additional baselines and dialogue diagnostics}
\label{app:baselines}

\noindent\textbf{Additional ALFWorld systems.}
Table~\ref{tab:alfworld-additional} extends the comparison to six systems adapted to the ALFWorld interface: Reflexion and SAGE \citep{reflexion,sage}, Live-Evo and Dynamic Cheatsheet \citep{liveevo,dynamiccheatsheet}, and A-Mem and AWM \citep{amem,awm}.
The retrieval-only control retains memory writing and fact retrieval but disables skills and fact return.
It therefore differs from the memory-only configuration in Table~\ref{tab:ablations}, which keeps fact return enabled.

\begin{table*}[t]
\centering
\small
\setlength{\tabcolsep}{0pt}
\renewcommand{\arraystretch}{1.06}
\begin{tabular}{@{}lll@{}}
\toprule
\tableleftcell{.48\linewidth}{\textbf{Method}}
& \tablecell{.26\linewidth}{\textbf{SR\%}}
& \tablecell{.26\linewidth}{\textbf{$\Delta$ vs. ReAct}} \\
\midrule
\tableleftcell{.48\linewidth}{ReAct}
& \tablecell{.26\linewidth}{$51.5$}
& \tablecell{.26\linewidth}{n/a} \\
\tableleftcell{.48\linewidth}{Reflexion}
& \tablecell{.26\linewidth}{$52.7$}
& \tablecell{.26\linewidth}{\gain{1.2}} \\
\tableleftcell{.48\linewidth}{SAGE}
& \tablecell{.26\linewidth}{$49.5$}
& \tablecell{.26\linewidth}{\drop{2.0}} \\
\tableleftcell{.48\linewidth}{Live-Evo}
& \tablecell{.26\linewidth}{$49.0$}
& \tablecell{.26\linewidth}{\drop{2.5}} \\
\tableleftcell{.48\linewidth}{Dynamic Cheatsheet}
& \tablecell{.26\linewidth}{$47.8$}
& \tablecell{.26\linewidth}{\drop{3.7}} \\
\tableleftcell{.48\linewidth}{A-Mem}
& \tablecell{.26\linewidth}{$43.2$}
& \tablecell{.26\linewidth}{\drop{8.3}} \\
\tableleftcell{.48\linewidth}{AWM}
& \tablecell{.26\linewidth}{$41.5$}
& \tablecell{.26\linewidth}{\drop{10.0}} \\
\tableleftcell{.48\linewidth}{Retrieval only}
& \tablecell{.26\linewidth}{$\mathbf{60.67}$}
& \tablecell{.26\linewidth}{\gain{\mathbf{9.17}}} \\
\bottomrule
\end{tabular}
\caption{\textbf{Additional ALFWorld success rates under the benchmark interface.}}
\label{tab:alfworld-additional}
\end{table*}

\noindent\textbf{Collection and serving.}
ExpeL scores $44.8\%$ when its serving path injects the collected insights.
A collection-only run scores $52.8\%$, but never presents those insights to the policy and therefore does not implement lifelong ExpeL.
For Reflexion, we report the first attempt under the matched action budget, rather than success across any of three attempts.

\noindent\textbf{Reader and dialogue settings.}
The document-reading results compare complete context-construction systems under an 8,192-token answer-context budget.
Each item supplies a document and question together.
The external readers rebuild their stores from each document, whereas \method{} retains one persistent external state across the fixed sequence.

On the common DialSim prefix, SimpleMem answers 457 of 955 scored questions correctly, giving the $47.85\%$ entry in Table~\ref{tab:performance}.
This metric excludes the 200 session ingestion operations.

\subsection{Comparison with MSCE}
\label{app:msce}

\noindent\textbf{Comparison settings.}
We compare \method{} with MSCE, which also learns memory and skills from experience \citep{msce2026}, on ALFWorld and name-substituted DialSim.
All systems use frozen Qwen3-8B, and persistent runs start from empty state.
Table~\ref{tab:msce} reports the highest-scoring MSCE configuration tested on DialSim.
\method{} achieves higher success on both benchmarks.

\begin{table*}[t]
\centering
\small
\renewcommand{\arraystretch}{1.08}
\textcolor{forwardblue}{\itshape Panel a: ALFWorld, with the revisit guard and navigation hint enabled}\par\vspace{1pt}
\begin{tabular*}{\linewidth}{@{\extracolsep{\fill}}lrrrr@{}}
\toprule
\textbf{System} & \textbf{Successes / $N$} & \textbf{SR\%} & \textbf{30 actions reached} & \textbf{Invalid actions} \\
\midrule
ReAct & 337/600 & $56.17$ & 268 & 905 \\
MSCE & 296/600 & $49.33$ & 308 & 1,082 \\
\textbf{\method{}} & 379/600 & $\mathbf{63.17}$ & 226 & 661 \\
\bottomrule
\end{tabular*}

\vspace{6pt}
\textcolor{forwardblue}{\itshape Panel b: Name-substituted DialSim}\par\vspace{1pt}
\begin{tabular*}{\linewidth}{@{\extracolsep{\fill}}lrrrrr@{}}
\toprule
\textbf{System} & \textbf{Successes / $N$} & \textbf{SR\%} & \textbf{Timed} & \textbf{Untimed} & \textbf{Unanswerable} \\
\midrule
ReAct & 72/955 & $7.54$ & $0.0$ & $3.4$ & $\mathbf{48.8}$ \\
MSCE & 157/957 & $16.41$ & $3.2$ & $32.9$ & $33.9$ \\
\textbf{\method{}} & 709/955 & $\mathbf{74.24}$ & $\mathbf{97.0}$ & $\mathbf{44.4}$ & $48.0$ \\
\bottomrule
\end{tabular*}
\caption{\textbf{Comparison with MSCE on ALFWorld and DialSim.}
Scores are percentages.
The ``30 actions reached'' count includes tasks solved on the final allowed action.
Panel b uses the highest-scoring tested MSCE configuration, with 957 questions and 535 timed questions.
ReAct and \method{} answer 955 questions, including 533 timed questions.
All three runs include 295 untimed and 127 unanswerable questions.}
\label{tab:msce}
\end{table*}

\noindent\textbf{ALFWorld: learning reusable procedures.}
\method{} achieves $63.17\%$ success, compared with $49.33\%$ for MSCE and $56.17\%$ for ReAct under the same navigation settings.
MSCE retrieves experience on 503 of 600 tasks but produces more invalid actions and reaches the action limit more often than either reference system.
Its skill candidates include procedures tied to particular objects and destinations, which may offer little guidance when either changes.
\method{} instead learns parameterized procedures from recurring episodes and retrieves task-specific facts separately, so the same procedure can apply to different objects and destinations.

\noindent\textbf{DialSim: retrieving relevant facts.}
\method{} reaches $74.24\%$ success, compared with $16.41\%$ for MSCE and $7.54\%$ for ReAct.
The largest gap is on timed questions, where accuracy is $97.0\%$ for \method{} and $3.2\%$ for MSCE.
In the selected MSCE run, only 347 of 995 recorded retrieval results contain source conversation passages, and the actor repeats the same incorrect answer on 153 questions.
MSCE selects experience by its assessed value, but these observations suggest that a useful past interaction may still lack the names or dates required by a new question.
\method{} retrieves typed facts with explicit name and date anchors to provide these details in the answer context.

\subsection{State dynamics and skill lifecycle}
\label{app:state-dynamics}

\noindent\textbf{Recorded deployments and joint exposure.}
The state audit follows skill deployments and the facts served alongside them in the 600-task ALFWorld run with both return channels enabled.
The task ledger records deployments and outcomes, while the terminal co-serving relation identifies facts and skills that have entered the context together.
Let $\operatorname{Co}(f)$ denote the deduplicated list of skill identities serialized beside terminal fact $f$.
Restricted to the five final persistent skill identities, the terminal relation is
\begin{equation}
\begin{aligned}
\mathcal G_T^{\rm co}
&=\bigl\{(f,s)\in\mathcal F_{T+1}\times\mathcal S_{T+1}:\\
&\qquad\operatorname{id}(s)\in\operatorname{Co}(f)\bigr\}.
\end{aligned}
\label{eq:terminal-coaudit}
\end{equation}
Of the 367 terminal facts, 223 have degree at least one, forming 453 distinct edges to the five final skill identities.
An edge indicates at least one joint exposure, possibly to an earlier version of the skill, but does not record its count, timestamp, or outcome.
These diagnostic links describe joint context rather than synthesis provenance and do not affect retrieval.

\begin{figure*}[t]
\centering
\begin{subfigure}[t]{.485\linewidth}
\centering
\includegraphics[width=\linewidth]{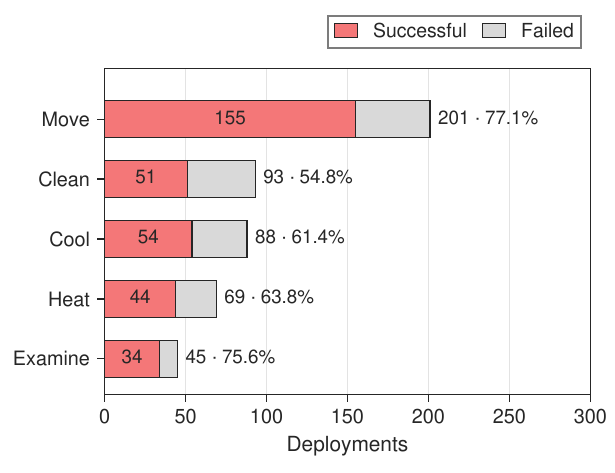}
\caption{Final task family coverage.}
\label{fig:coevolution-full-coverage}
\end{subfigure}\hfill
\begin{subfigure}[t]{.485\linewidth}
\centering
\includegraphics[width=\linewidth]{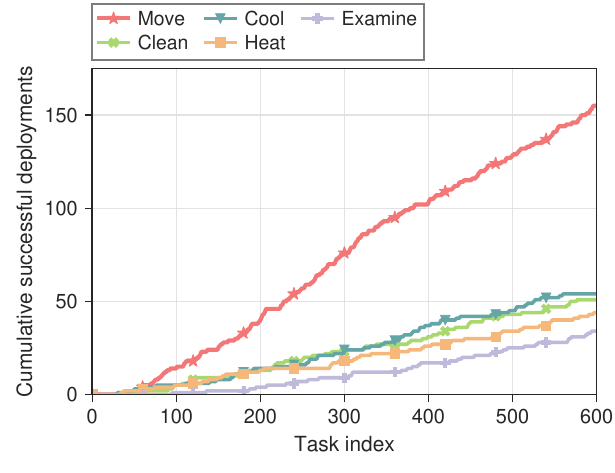}
\caption{Cumulative skill reuse.}
\label{fig:coevolution-full-reuse}
\end{subfigure}

\medskip
\begin{subfigure}[t]{.485\linewidth}
\centering
\includegraphics[width=\linewidth]{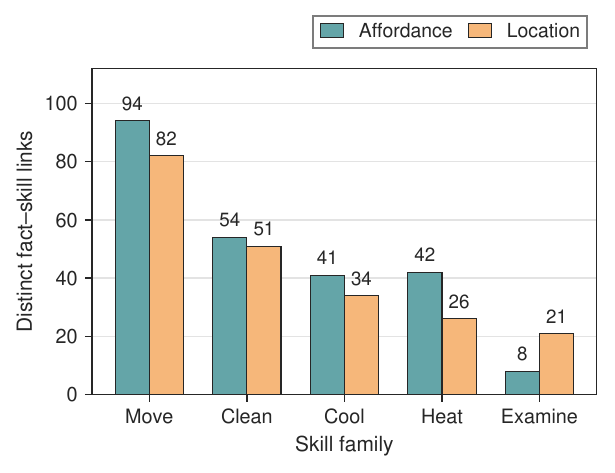}
\caption{Fact--skill co-serving links.}
\label{fig:coevolution-full-coserving}
\end{subfigure}\hfill
\begin{subfigure}[t]{.485\linewidth}
\centering
\includegraphics[width=\linewidth]{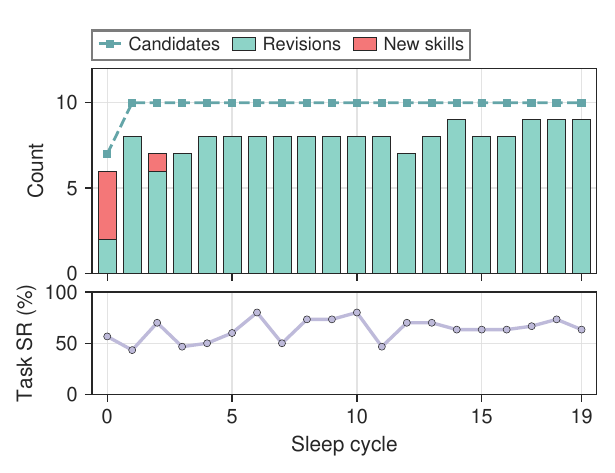}
\caption{Skill consolidation and selection.}
\label{fig:coevolution-full-consolidation}
\end{subfigure}
\caption{\textbf{ALFWorld state dynamics with both return channels enabled.}
Panel~\subref{fig:coevolution-full-coverage} reports deployments, successes, and success rates by family.
Panels~\subref{fig:coevolution-full-reuse} and~\subref{fig:coevolution-full-coserving} show cumulative successful deployments and terminal co-serving links by fact type.
Panel~\subref{fig:coevolution-full-consolidation} shows candidates as a dashed line, accepted new skills and revisions as stacked bars, and task success in each cycle.}
\label{fig:coevolution-full}
\end{figure*}

\noindent\textbf{Continued reuse and revision.}
The final catalog contains one skill for each ALFWorld task family, with 496 deployments and 338 successes across the stream.
Figure~\ref{fig:coevolution-full-coverage} shows how deployments and success rates differ across these skills, while Figure~\ref{fig:coevolution-full-coserving} groups their co-served facts by type.
The consolidation history records 159 accepted operations, of which five create skills and 154 revise existing skills.
Thus the catalog continues to change after it covers all task families, as also shown in Figure~\ref{fig:coevolution}.

\begin{figure*}[t]
\centering
\includegraphics[width=\linewidth]{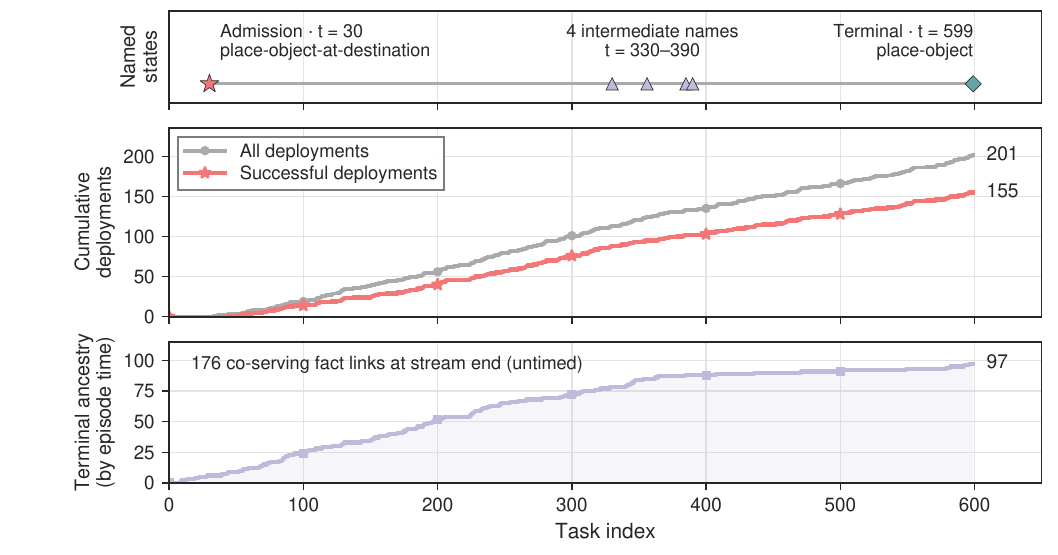}
\caption{\textbf{Lifecycle of the most frequently deployed final skill.}
From top to bottom, the traces show admission and recorded name changes, all 201 deployments with 155 successes, and the 97 episodes in the final synthesis ancestry.
The 176 distinct terminal fact co-serving links have no timestamps and may involve earlier skill texts.
Same-name revisions overwrite prior versions, so the records do not recover every revision or ancestry size over time.}
\label{fig:skill-lifecycle}
\end{figure*}

\noindent\textbf{One skill's lifecycle.}
The place-object skill is admitted near task 30 and remains active throughout the stream.
Its deployment history, changes in name and scope, and final episode ancestry show reuse under a persistent identity as the procedure evolves.
The snapshot comparison in Figure~\ref{fig:skill-revision-example} gives concrete examples of these textual changes.

\subsection{Prospective intervention on name-substituted DialSim}
\label{app:dialsim-intervention}

\noindent\textbf{Intervention and findings.}
The advantage of persistent state grows when the original character names are removed from DialSim.
We consistently renamed every character while preserving the dialogue structure and answers.
Before running the intervention, we specified that persistent state should improve the name-substituted condition by at least $8$ percentage points and that this gain should exceed the gain with the original names.
We also expected ReAct to perform better with the original names.
All three criteria were met.
\method{}'s gain over ReAct increased from $47.0$ points with the original names to $66.7$ points after renaming, while ReAct fell from $26.9\%$ to $7.54\%$.
The result supports the value of retrieving deployment-specific evidence when cues associated with familiar names are unavailable.

\subsection{Actor-call efficiency and an MBPP context case study}
\label{app:efficiency}

\noindent\textbf{Measurement.}
We measure actor effort from the completed runs.
For ALFWorld, logical actor calls include action steps and logged invalid-action retries on scored wake tasks, excluding held-out evaluation.
For MBPP, the saved trajectories report actor steps but not a complete call counter.
Both measures describe actor effort rather than total LLM cost because they exclude memory writes, synthesis, audit, and transport retries.
The retrieval quotas below are configured limits, not measured token counts.

\begin{figure*}[t]
\centering
\begin{subfigure}[t]{.485\linewidth}
\centering
\includegraphics[width=\linewidth]{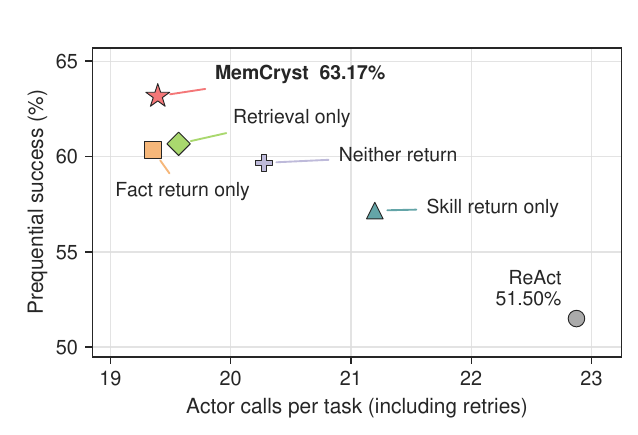}
\caption{Cost vs. performance.}
\label{fig:actor-efficiency-cost}
\end{subfigure}\hfill
\begin{subfigure}[t]{.485\linewidth}
\centering
\includegraphics[width=\linewidth]{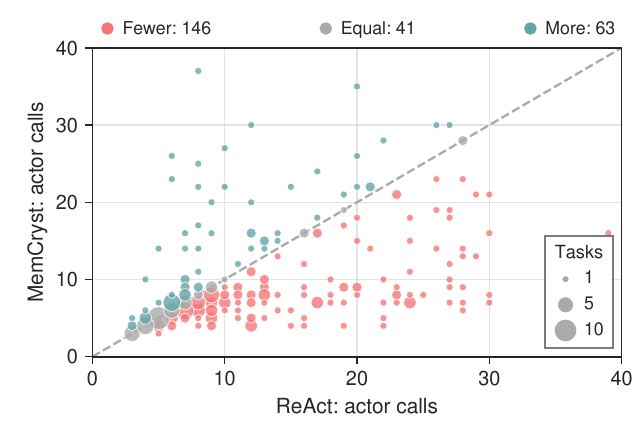}
\caption{Paired calls on shared successes.}
\label{fig:actor-efficiency-paired}
\end{subfigure}
\caption{\textbf{ALFWorld actor-call efficiency.}
All methods use Qwen3-8B and 600 tasks.
ReAct uses its original unguarded setting, while the other configurations retain the revisit guard.
Calls include action retries but exclude memory maintenance.
Panel~\subref{fig:actor-efficiency-paired} shows the 250 shared successes, with marker size indicating task count.}
\label{fig:actor-efficiency}
\end{figure*}

\noindent\textbf{ALFWorld improves both completion and actor-call efficiency.}
\method{} raises success from $51.50\%$ to $63.17\%$ while using $15.22\%$ fewer actor calls than ReAct.
Figure~\ref{fig:actor-efficiency} compares six configurations over 600 tasks using the original ReAct baseline.
The full system uses 11,636 calls, comprising 10,975 action steps and 661 invalid-action retries.
ReAct uses 13,725 calls, comprising 12,851 steps and 874 retries.
Fact return alone uses slightly fewer calls at 11,612 but solves 362 tasks rather than 379.
In these runs, enabling both returns therefore yields 17 more completions for 24 additional calls.

\noindent\textbf{Shared successful tasks.}
On the 250 tasks solved by both systems, mean actor calls fall from $12.90$ to $10.13$, a $21.49\%$ reduction.
\method{} uses fewer calls on 146 of these tasks, the same number on 41, and more on 63.
The reduction in overall actor effort is therefore also present among tasks that both systems solve.

\begin{figure*}[t]
\centering
\begin{subfigure}[t]{.485\linewidth}
\centering
\includegraphics[width=\linewidth]{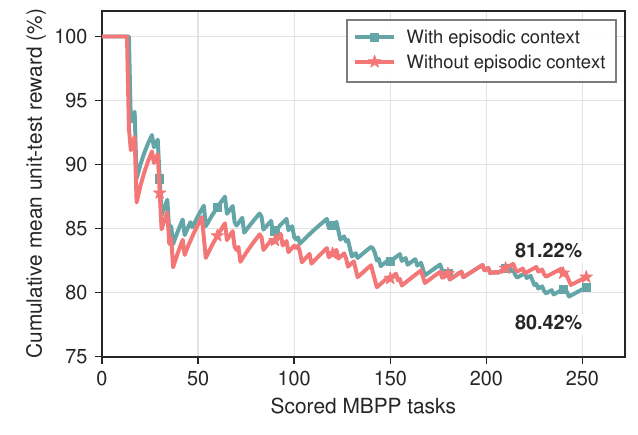}
\caption{Higher reward.}
\label{fig:context-efficiency-reward}
\end{subfigure}\hfill
\begin{subfigure}[t]{.485\linewidth}
\centering
\includegraphics[width=\linewidth]{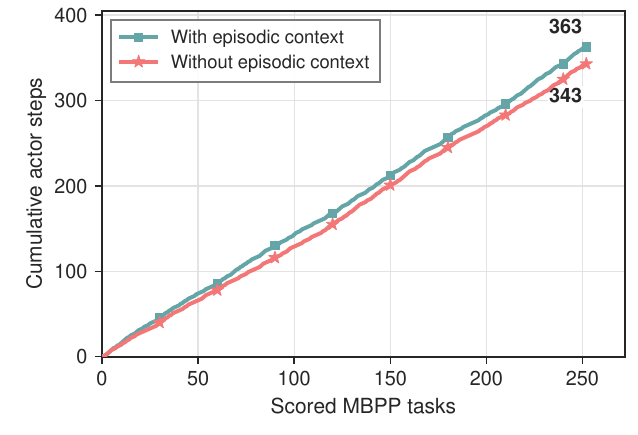}
\caption{Fewer actor steps.}
\label{fig:context-efficiency-steps}
\end{subfigure}
\caption{\textbf{MBPP with and without episodic context.}
Curves show cumulative mean reward and actor steps over the same 252 tasks with Qwen3-8B and retrieval limits of four facts and two skills.}
\label{fig:context-efficiency}
\end{figure*}

\noindent\textbf{MBPP context case study.}
Retaining episodes for learning does not require serving them on every task.
Figure~\ref{fig:context-efficiency} compares two configurations on the same 252 MBPP tasks, differing only in whether past episodes enter the actor's context.
Both configurations learn online from their own trajectories and retrieve up to four facts and two skills.
They use the same code execution and self-test interface, deduplicate skills, and require fact support of at least two.

\noindent\textbf{Reward and actor steps.}
Removing episodic context raises mean reward from $80.42\%$ to $81.22\%$, a gain of $0.79$ percentage points computed before rounding.
Actor steps fall from 363 to 343, a $5.51\%$ reduction.
This comparison supports keeping episodes available for learning while selecting more narrowly what enters the immediate action context.

\FloatBarrier
\section{Experimental details}
\label{app:experimental-details}

\subsection{Default framework settings}
\label{app:framework-hyperparameters}

\noindent\textbf{Shared settings.}
Table~\ref{tab:framework-hyperparameters} specifies the common rules for memory writing, skill learning, and outcome return.
Each benchmark adapter sets the state lifetime, retrieval depth, encoder, sleep schedule, and task budget for its interface.
The default text encoder is \texttt{all-MiniLM-L6-v2} \citep{allminilml6v2}.

\begin{table*}[t]
\centering
\small
\setlength{\tabcolsep}{4pt}
\renewcommand{\arraystretch}{1.05}
\begin{tabular}{@{}p{.12\linewidth}p{.22\linewidth}@{\hspace{5pt}\vrule\hspace{7pt}}p{.61\linewidth}@{}}
\toprule
Block & Hyperparameter & Default \\
\midrule
\multirow{2}{*}{LLM}
& Primary backbone & Frozen Qwen3-8B \\
& Role temperatures & actor $0$, writer $.2$, synthesizer $.3$, and auditor $0$ \\
\midrule
State & Initial external state & empty episode store, fact store, and skill catalog \\
\midrule
\multirow{3}{*}{Memory}
& Episode weight & $1.3$ after success and $.7$ after failure \\
& Prior for a new fact & $n_f=h_f=0$, hence $C_f=.5$ \\
& Fact merge & same entity and scope with cosine threshold $\eta=.86$ \\
\midrule
\multirow{3}{*}{Retrieval}
& Default text encoder & \texttt{all-MiniLM-L6-v2} \\
& Skill utility floor & $\max\{.05,U_s\}$ \\
& Fact score constants & support coefficient $.05$, support cap $6$, outcome-score offset $.5$ \\
\midrule
\multirow{2}{*}{Return}
& Skill update rate & EMA rate $\alpha_S=.2$ \\
& Eligible records & facts and skills served in the current action context \\
\midrule
\multirow{5}{*}{\shortstack[l]{Skill\\consolidation}}
& Clusterer & HDBSCAN, Euclidean distance, EOM selection \\
& HDBSCAN settings & minimum cluster size $3$, minimum samples $1$, and selection $\epsilon=0$ \\
& Cluster gate & $|C|\geq3$ and $Q(C)\geq.4$ \\
& Cluster order & $Q(C)|C|\max\{.05,p(C)\}$ \\
& Limits per pass & at most $10$ candidate clusters and up to $5$ representative episodes \\
\midrule
\multirow{3}{*}{Skill catalog}
& Audit gate & at least $10$ returned passes, with $13$ checks requested \\
& Initial skill utility & $U_s=\operatorname{clip}(Q(C),.4,.95)$ \\
& Capability match & revise an exact match in place and insert a new skill otherwise \\
\bottomrule
\end{tabular}
\caption{\textbf{Default \method{} framework settings.}}
\label{tab:framework-hyperparameters}
\end{table*}

\subsection{Benchmark evaluation configuration}
\label{app:evaluation-config}

\noindent\textbf{Comparison settings.}
Within each benchmark comparison, all methods share the evaluator, answer or action interface, and backbone.
State lifetime and serving behavior follow the evaluated system.
We use frozen Qwen3-8B for the ten-setting comparison, the interleaved stream, and the mechanism studies, including \method{}'s memory and skill updates.
Table~\ref{tab:benchmark-streams} gives the number of items scored for \method{}, with additional baselines in Appendix~\ref{app:baselines} and the MSCE comparison in Appendix~\ref{app:msce}.

\noindent\textbf{ALFWorld comparisons.}
The headline comparison pairs unguarded ReAct with guarded \method{} and therefore compares deployed systems.
The component and return ablations instead share the same 600 tasks, task order, and guard.
The no-revision configuration freezes admitted skill text but retains skill creation, serving, and utility updates.

\noindent\textbf{Reader and dataset variants.}
For RULER and BABILong, ReAct reads only the document tail, while the external readers rebuild their stores for each document as described in Section~\ref{app:baselines}.
\method{} retains one persistent external state across each fixed document sequence, so these comparisons evaluate complete reader systems rather than isolate memory lifetime.
LongMemEval uses a reduced haystack, and DialSim-subst. denotes the name-substituted version of DialSim.
LCB+ combines LiveCodeBench with a LiveCodeBench Pro extension \citep{livecodebench,livecodebenchpro}.

\noindent\textbf{Interleaved stream.}
The 1,088-event stream in Table~\ref{tab:capstone} includes 70 unscored LoCoMo session ingestions, followed or interleaved with 120 RULER, 100 BABILong, 151 MATH-hard, 150 MBPP, and 497 LoCoMo queries.
The frozen backbone uses one evolving external state throughout, with domain stamps restricting evidence and capability routing selecting skills.
Task adapters provide the action interfaces.
The specialist reference uses separate domain runs, with BABILong and MBPP scores from contemporaneous runs that preceded the later sweeps in Table~\ref{tab:performance}.

\noindent\textbf{Online protocol.}
Persistent runs start from empty state and score each item before writing its completed interaction.
Standalone ingestion events contribute evidence but are unscored.
Each run follows a fixed task order, with some baseline runs ending at different fixed prefixes.
Unless stated otherwise, each result comes from one run.
Appendix~\ref{app:full-loop} specifies the event chronology, and Appendix~\ref{app:terminal} reports supplementary evaluations with the final state frozen.
The native reward $r_t$ enters $\preq$, whereas the binary signal $y_t$ updates the influence of exposed records through Eqs.~\ref{eq:skillreturn} and \ref{eq:factreturn}.
Table~\ref{tab:evaluation-config} specifies the scoring metric and return signal for each benchmark.

\begin{table*}[t]
\centering
\small
\setlength{\tabcolsep}{3pt}
\renewcommand{\arraystretch}{1.05}
\begin{tabular}{@{}p{.14\linewidth}p{.14\linewidth}p{.365\linewidth}p{.15\linewidth}p{.15\linewidth}@{}}
\toprule
Benchmark & State scope & Evaluation mode & Native reward $r_t$ & Return signal $y_t$ \\
\midrule
\multicolumn{5}{c}{\itshape Embodied control and reading long documents} \\
\midrule
ALFWorld & persistent & legal text actions with a limit of 30 steps & binary success & $r_t$ \\
RULER & persistent for \method{} & one document and one question per item with an answer context of 8,192 tokens & mean recall reward & $\mathbf 1[r_t\geq.99]$ \\
BABILong & persistent for \method{} & one document and one question per item with an answer context of 8,192 tokens & mean reward & $\mathbf 1[r_t\geq.99]$ \\
\midrule
\multicolumn{5}{c}{\itshape Conversation and memory} \\
\midrule
LoCoMo & per dialogue & session ingestion followed by short answers & token F1 & $\mathbf 1[\mathrm{F1}\geq.5]$ \\
LongMemEval & per instance & reduced haystack ingestion followed by a short answer & token F1 & $\mathbf 1[\mathrm{F1}\geq.5]$ \\
DialSim-subst. & persistent & session ingestion followed by short answers & binary success & $r_t$ \\
\midrule
\multicolumn{5}{c}{\itshape Mathematical reasoning and code generation} \\
\midrule
MATH-hard & persistent & unconstrained reasoning and answer & exact answer reward & $\mathbf 1[r_t\geq.99]$ \\
MBPP & persistent & generate code, execute it, and test it & unit test reward & $\mathbf 1[r_t\geq.99]$ \\
CodeContests & persistent & generate code, execute it, and test it & unit test reward & $\mathbf 1[r_t\geq.99]$ \\
LCB+ & persistent & generate code, execute it, and test it & unit test reward & $\mathbf 1[r_t\geq.99]$ \\
\bottomrule
\end{tabular}
\caption{\textbf{Benchmark evaluation settings.}
Persistent streams begin empty and score each item before writeback.}
\label{tab:evaluation-config}
\end{table*}

\begin{table*}[!t]
\centering
\small
\renewcommand{\arraystretch}{1.11}
\begin{tabular*}{\linewidth}{@{\extracolsep{\fill}}llrl@{}}
\toprule
\textbf{Benchmark} & \textbf{Domain} & \textbf{$N$} & \textbf{Metric} \\
\midrule
ALFWorld & Embodied control & 600 & Success rate \\
RULER & Long document reading & 400 & Recall \\
BABILong & Long document reading & 200 & Mean reward \\
\midrule
LoCoMo & Conversation memory & 1,977 & Token F1 \\
LongMemEval & Conversation memory & 60 & Token F1 \\
DialSim-subst. & Conversation memory & 955 & Success rate \\
\midrule
MATH-hard & Mathematics & 261 & Exact match \\
MBPP & Code generation & 252 & Unit test reward \\
CodeContests & Code generation & 160 & Unit test reward \\
LCB+ & Code generation & 232 & Unit test reward \\
\bottomrule
\end{tabular*}
\caption{\textbf{Benchmarks, sample counts, and metrics.}
$N$ counts the items scored for \method{}.}
\label{tab:benchmark-streams}
\end{table*}

Some baseline runs score more items than the \method{} counts in Table~\ref{tab:benchmark-streams}.
The main DialSim comparison in Table~\ref{tab:performance} uses the common 955-question prefix, whereas Table~\ref{tab:msce} reports the MSCE run over 957 questions with its original denominator.

\noindent\textbf{Score aggregation.}
Benchmark scores are percentages, and absolute differences are in percentage points.
In Table~\ref{tab:performance}, Overall equally weights benchmarks within panels and the three panel means in panel d.
Trailing zeros standardize table formatting without increasing the precision of recorded summaries.
We compute averages and differences before final display rounding and report Overall only when all required scores are available.

\noindent\textbf{Adapter settings.}
For ALFWorld, \method{} uses a navigation hint and a task-local revisit guard.
If the agent holds no object and attempts to revisit a location, the guard substitutes the first valid navigation action to an unvisited location, when available.
This adaptation changes within-task action execution while leaving the memory and skill update rules unchanged.
The LoCoMo adapter uses BGE-M3 \citep{bgem3}, retrieves 12 facts, and may serve verbatim excerpts within an answer budget of four steps.
The code adapter retains episodic records and retrieves two deduplicated skills and four facts with support of at least two.
It uses the same execution and self-test interface as ReAct.

\subsection{Generated state and parser contracts}
\label{app:generation-contracts}

\noindent\textbf{Memory writes.}
The episode writer turns a completed trajectory and its outcome into one JSON record containing task evidence, diagnosis, and a reusable lesson.
If the call or parse fails, the writer skips the episode.
The fact extractor produces typed entity and claim records, while dialogue adapters parse one \texttt{entity | fact} record per line.
The parser drops malformed items and adds no facts from an unparseable response.
A completed task can therefore update the influence of previously served records even if writing new content fails.

\noindent\textbf{Skill writes.}
The synthesizer receives candidate-cluster statistics and up to five episodes with the highest weights.
It returns a \texttt{SKILL.md} record with the required frontmatter and a parameterized procedure, which the parser rejects if it is malformed or nonparametric.
The auditor then reads the complete skill proposal, cluster statistics, and the same episodes in a fresh context without the synthesis transcript.
The prompt requests the 13 checks in Table~\ref{tab:audit-checks}, and admission requires at least 10 returned \texttt{PASS} verdicts.
The parser also rejects unparseable audit JSON.
A rejected proposal does not enter the skill catalog, but the sleep pass continues with the remaining candidates.

\subsection{Compute resources}
\label{app:resources}

\noindent\textbf{Inference resources.}
We served Qwen3-8B and Qwen3-1.7B locally on NVIDIA L40 GPUs.
For the Qwen3 experiments, local inference supported task execution as well as memory writing, skill synthesis, and auditing.
The GPT-4o-mini experiments used the OpenAI API with API-key authentication.

\FloatBarrier
% Original source excerpts and matched-context case sheets.
% Original saved text and metadata; extracted by extract_qualitative_originals.py.
\twocolumn[{%
\begin{minipage}{\textwidth}
\section{Qualitative examples of joint memory and skill evolution}
\label{app:qualitative-examples}

These examples show how stored experience becomes procedural guidance and context for task execution.
Section~\ref{app:qualitative-originals} presents memory records and skill snapshots from the 600-task ALFWorld run with both return channels enabled.
The saved \texttt{improvement} fields contain the memory writer's lessons, whereas Section~\ref{app:qualitative-trajectories} presents the actual inputs and outputs of four task executions.
Source wording and any truncation in the logged context are preserved, with line wrapping for readability.
The cards display skill utilities and derived outcome scores as percentages.
Repeated prompt blocks appear once, followed by each call's changing input, complete response, and environment return.
Task indices start at zero.

The cleaning and heating records link experience with different objects to persistent skills.
Later skill snapshots give more direct placement guidance and require both lamp visibility and possession of the target object.
The cooling example combines object-location facts with a procedure, while the dialogue example preserves the evidence needed to answer a purchase question.
All four task executions include retrieved memory in the actor's input, and ALFWorld, MATH-hard, and MBPP also include skills.
These selected examples show the supplied evidence and recorded behavior without isolating the contribution of individual records.

\subsection{Original memory records and skill snapshots}
\label{app:qualitative-originals}

\begin{casefigure}
\casecard{.485\linewidth}{casegray}{Successful cleaning: original memory}{48mm}{%
\cardmeta{Task 31 \enspace / \enspace episode \recordid{5b2000b6}}\par\smallskip
\cardlabel{question}\par
\casecode{clean a <target> and place it in a <destination>}\par\smallskip
\cardlabel{improvement}\par
\casecode{go to countertop 2, take mug 1 from countertop 2, go to sinkbasin 1, clean mug 1 with sinkbasin 1, go to coffeemachine 1, move mug 1 to coffeemachine 1}\par\medskip
\cardlabel{success}\enspace\casecode{true}\quad
\cardlabel{steps}\enspace15\par
\cardlabel{Served skill}\enspace\recordid{e654d0ba}}%
\hfill
\casecard{.485\linewidth}{casesage}{Cleaning skill: terminal procedure}{48mm}{%
\cardmeta{Persistent skill \recordid{e654d0ba}}\par\smallskip
\cardlabel{Original procedure excerpt}\par
Clean Before Placing: Do not place \casecode{<target\_object>} at \casecode{<destination>} until you have run
\casecode{clean <target\_object> with sink}. Apply this when holding the correct object, before any placement.\par\medskip
\cardlabel{Ancestry excerpt}\enspace\recordid{5b2000b6}}%
\caption{\textbf{A cleaning lesson and a terminal skill excerpt.}
The episode belongs to the skill's terminal ancestry.}
\label{fig:running-example}
\end{casefigure}

\begin{casefigure}
\casecard{.485\linewidth}{casegray}{Bread: original stored lesson}{57mm}{%
\cardmeta{Task 2 \enspace / \enspace episode \recordid{2501fa18}}\par\smallskip
\cardlabel{question}\par
\casecode{heat a <target> and place it in a <destination>}\par\smallskip
\cardlabel{improvement}\par
\casecode{go to countertop 1, take bread 1 from countertop 1, go to microwave 1, heat bread 1 with microwave 1, go to countertop 1, move bread 1 to countertop 1}\par\medskip
\cardlabel{success}\enspace\casecode{true}\quad
\cardlabel{steps}\enspace22\par
\cardlabel{Served skills}\enspace\casecode{[]}\par\smallskip
\cardlabel{Terminal skill ancestry}\enspace\recordid{3b5e1faf}}%
\hfill
\casecard{.485\linewidth}{casesage}{Cup: original stored lesson}{57mm}{%
\cardmeta{Task 38 \enspace / \enspace episode \recordid{51c73cdf}}\par\smallskip
\cardlabel{improvement}\par
\casecode{go to cabinet 5, open cabinet 5, take cup 2 from cabinet 5, go to microwave 1, open microwave 1, heat cup 2 with microwave 1, go to sidetable 1, move cup 2 to sidetable 1}\par\medskip
\cardlabel{success}\enspace\casecode{true}\quad
\cardlabel{steps}\enspace14\par
\cardlabel{Served skill}\enspace\recordid{3b5e1faf}\par\smallskip
\cardlabel{Facts from task 38}\par
\casecode{cup: often stored in cabinets}\par
\casecode{sidetable: often used to place objects after heating}}%
\caption{\textbf{Heating lessons for different objects and destinations.}
Both episodes belong to the heating skill's terminal ancestry.}
\label{fig:heating-example}
\end{casefigure}
\end{minipage}
}]

\twocolumn[{%
\begin{minipage}{\textwidth}
\begin{casefigure}
\noindent\textbf{Object placement}\hfill
\cardmeta{Same persistent skill \recordid{d94f43bc}}\par\vspace{6pt}
\casecard{.485\linewidth}{casegray}{Earlier saved skill}{41mm}{%
\caserow{When to navigate}{%
Apply this when the target object is in hand and \textbf{the destination is in view}.}
\caserow{How to deliver}{%
Once at \casecode{<destination>}, put \casecode{<target\_object>} in/on \casecode{<destination>}.}
\caserow{What to avoid}{%
Failing to search systematically for the target object.}
\cardmeta{40 supporting episodes \enspace / \enspace 33 successes in 49 deployments}\par
\cardmeta{Utility $U=81.57\%$}}%
\hfill
\casecard{.485\linewidth}{casesage}{Terminal saved skill}{41mm}{%
\caserow{When to navigate}{%
Apply this \textbf{after confirming} the correct object is in hand.}
\caserow{How to deliver}{%
Once at \casecode{<destination>}, put \casecode{<target\_object>} in/on \casecode{<destination>} \textbf{with no detours}.}
\caserow{What to avoid}{%
\textbf{Assuming the object is at the expected location} without searching}
\cardmeta{97 supporting episodes \enspace / \enspace 155 successes in 201 deployments}\par
\cardmeta{Utility $U=68.66\%$}}%

\vspace{8pt}
\noindent\textbf{Examination with a lamp}\hfill
\cardmeta{Same persistent skill \recordid{b6188de6}}\par\vspace{6pt}
\casecard{.485\linewidth}{casegray}{Earlier saved skill}{20mm}{%
\cardlabel{When to use the lamp}\par\smallskip
Apply this as soon as the desklamp is in view \textbf{or} right after acquiring \casecode{<target\_object>}.\par\medskip
\cardmeta{8 supporting episodes \enspace / \enspace 2 successes in 7 deployments}}%
\hfill
\casecard{.485\linewidth}{casesage}{Terminal saved skill}{20mm}{%
\cardlabel{When to use the lamp}\par\smallskip
Apply this when the desklamp is in view \textbf{and the target object is in inventory}.\par\medskip
\cardmeta{23 supporting episodes \enspace / \enspace 34 successes in 45 deployments}}%
\caption{\textbf{Procedure revisions under persistent skill identities.}
The nonconsecutive snapshots retain the saved wording, with emphasis added.}
\label{fig:skill-revision-example}
\end{casefigure}

\begin{casefigure}
\casecard{.485\linewidth}{qualochre}{Cleaning skill: saved terminal statistics}{30mm}{%
\cardmeta{Skill \recordid{e654d0ba} \enspace / \enspace \texttt{clean-and-place}}\par\smallskip
\cardlabel{use\_count}\enspace 93\par
\cardlabel{failure\_count}\enspace 42\par
\cardlabel{ancestry entries}\enspace 55\par\smallskip
\cardlabel{utility as a percentage}\par
$U=66.81\%$}%
\hfill
\casecard{.485\linewidth}{qualforest}{Sink fact: saved terminal record}{30mm}{%
\cardmeta{Fact \recordid{ade31094}}\par\smallskip
\cardlabel{entity}\enspace\casecode{sinkbasin}\par
\cardlabel{fact}\enspace``can be used to clean objects''\par
\cardlabel{support}\enspace18\quad
\cardlabel{used}\enspace24\quad
\cardlabel{helped}\enspace18\par\smallskip
\cardlabel{Derived outcome score}
\[
C=\frac{18+1}{24+2}\approx73.1\%.
\]}%
\caption{\textbf{Terminal use statistics for a skill and a fact.}
The fact score is computed from the saved \texttt{used} and \texttt{helped} counts.}
\label{fig:qualitative-clean}
\end{casefigure}

\begin{casefigure}
\casecard{.68\linewidth}{qualforest}{Vase fact: one successful exposure}{27mm}{%
\cardmeta{Original terminal fact \recordid{22d3a8ce}}\par\smallskip
\cardlabel{entity}\enspace\casecode{vase}\par
\cardlabel{fact}\enspace``can be picked up and examined''\par
\cardlabel{from\_episodes}\enspace\casecode{[81]}\par
\cardlabel{used}\enspace1\quad\cardlabel{helped}\enspace1\par
\cardlabel{supports\_skills}\enspace\casecode{[d94f43bc]}\par\smallskip
\cardlabel{Derived outcome score}\enspace$C=66.7\%$}%
\caption{\textbf{A fact record after one successful exposure.}
\texttt{from\_episodes} identifies the source task and \texttt{supports\_skills} lists co-served skills.}
\label{fig:qualitative-returns}
\end{casefigure}
\end{minipage}
}]

% Saved evidence: analysis/qualitative_frozen_cases.json, task 40, condition A.
\twocolumn[{%
\begin{minipage}{\textwidth}
\subsection{Served context on a successful held-out task}
\label{app:qualitative-contexts}

\begin{casefigure}
\qualcard{\linewidth}{casesage}{Goal: ``cool some bread and put it in countertop''}{%
\cardmeta{Held-out task 40 \enspace / \enspace success in 9 actions \enspace / \enspace 376 context tokens}\par\medskip
\begin{minipage}[t]{.48\linewidth}
\vspace{0pt}\cardlabel{Served skill excerpt}\par\smallskip
\casecode{Skill 1: cool-and-place}\par\smallskip
``Prep Cooling Appliance: Go to the fridge and open it so it is ready. Apply this as soon as the fridge is in view or right after acquiring \casecode{<target\_object>}.''
\end{minipage}\hfill
\begin{minipage}[t]{.48\linewidth}
\vspace{0pt}\cardlabel{Served fact excerpts}\par\smallskip
``bread: can be cooled using a fridge ($4\times$)''\par\smallskip
``bread: often located on a countertop ($9\times$); often located on a diningtable ($2\times$)''
\end{minipage}}
\caption{\textbf{Served context from the frozen-state evaluation in Appendix~\ref{app:frozen-context}.}
Excerpts show the cooling skill and bread facts from a 376-token context.
This record contains the input and outcome counters, but no action transcript.}
\label{fig:qualitative-frozen-selection}
\end{casefigure}

% Evidence: analysis/qualitative_dialogue_cases.json.
% Regenerate evidence with scripts/extract_qualitative_dialogue_cases.py.
\subsection{Conversation memory and answer outcomes}
\label{app:qualitative-dialogue}
\begin{casefigure}
\qualcard{\linewidth}{casesage}{Successful recall and completion-linked memory}{%
\cardmeta{Task 13 \enspace / \enspace name-substituted DialSim}\par\smallskip
\begin{minipage}[t]{.46\linewidth}
\vspace{0pt}\cardlabel{Original conversation}\par\smallskip
\cardmeta{September 23, 1994 \enspace / \enspace source task 10}\par\smallskip
Elena: How'd you pay for them?\par
Diana: Uh, credit card.\par
Elena: And who pays for that?\par
Diana: Um... my... father.\par\medskip
\cardlabel{Source episode}\enspace\recordid{76845064}\par\medskip
\cardlabel{Terminal fact}\enspace\recordid{7bb0e489}\par\smallskip
\cardlabel{entity}\enspace\casecode{dad's credit card}\par
\cardlabel{fact}\par
``Diana used Dad's credit card to buy the swanky new boots''\par\smallskip
\cardlabel{from\_episodes excerpt}\par
\casecode{[13, 14, 21, 35, 40, ...]}
\end{minipage}\hfill
\begin{minipage}[t]{.50\linewidth}
\vspace{0pt}\cardlabel{Original question}\par\smallskip
``So, you remember eavesdropping on that chat back on September 23, 1994, could you be any more secretive about how Diana snagged those swanky new boots?''\par\smallskip
\cardlabel{Options}\par\smallskip
She found money outside Central Perk / She got a job / Dad's credit card / Elena lent her some money because she felt bad for her / I don't know. (None of the above)\par\smallskip
``Answer with the exact option text.''\par\medskip
\cardlabel{Reference answer}\enspace Dad's credit card\par
\cardlabel{success}\enspace\casecode{true}\quad
\cardlabel{steps}\enspace1
\end{minipage}}

\caption{\textbf{Conversation evidence and a completion-linked fact.}
The answer shown is the dataset reference.
This task record preserves the outcome but not the actor's context or response.}
\label{fig:qualitative-dialogue}
\end{casefigure}

\end{minipage}
}]

% Complete actor-interface logs, generated by scripts/render_qualitative_demo4.py.
% Blocks can cross page boundaries. Only identical ALFWorld prompt blocks
% are shared, with an exact reconstruction assertion in the renderer.
\onecolumn
\raggedbottom
\subsection{Task execution trajectories across four benchmarks}
\label{app:qualitative-trajectories}

\newlength{\tracecharwidth}
\newcommand{\tracemeta}[1]{\par\smallskip\noindent{\small #1}\par}
\newcommand{\tracecall}[1]{\par\bigskip\noindent\textbf{#1}\par\nopagebreak}
\newcommand{\traceline}[1]{\noindent\strut #1\par}
\newcommand{\traceblank}{\par\vspace{1.5pt}}
\newenvironment{trajectoryblock}[1]{%
  \par\begingroup
  \nolinenumbers
  \fontsize{9}{10}\selectfont
  \setlength{\OuterFrameSep}{-1pt}%
  % Horizontal padding leaves the original line width unchanged.
  \def\FrameCommand##1{\fboxsep=0pt\colorbox{casesage!6!white}{%
    \hspace{4pt}##1\hspace{4pt}}}%
  \MakeFramed{\FrameRestore}%
  \noindent\makebox[\linewidth][c]{\colorbox{casesage!18!white}{%
    \parbox{\dimexpr\linewidth+8pt-2\fboxsep\relax}{%
      \small\bfseries\textcolor{casesage!90!black}{#1}}}}%
  \par\nobreak\vspace{2pt}
  \fontsize{9}{10}\selectfont\ttfamily\raggedright
  \settowidth{\tracecharwidth}{0}%
  \setlength{\parindent}{0pt}\setlength{\parskip}{0pt}%
  \hyphenpenalty=10000\exhyphenpenalty=10000
  \clubpenalty=10000\widowpenalty=10000
}{%
  \par\kern2pt{\color{casesage!35!white}\hrule height .3pt}%
  \endMakeFramed
  \endgroup\par
}

% Generated from saved model calls. Do not edit quoted content.
\subsubsection{ALFWorld}

\tracemeta{6 actor calls \quad / \quad 1 skill \quad / \quad 4 facts}

\begin{trajectoryblock}{Retrieved record inventory}
\traceline{skills: 1}
\traceline{\hspace*{2\tracecharwidth}b6188de6: examine-with-desklamp}
\traceline{facts: 4}
\traceline{\hspace*{2\tracecharwidth}9c5b3e4e: desklamp | often located on a sidetable}
\traceline{\hspace*{2\tracecharwidth}c2432ef9: desklamp | can be used to examine objects}
\traceline{\hspace*{2\tracecharwidth}810edf1e: bowl | can be found on a desk or sidetable}
\traceline{\hspace*{2\tracecharwidth}e6e9e249: desklamp | can be turned on to provide light}
\traceline{reference episodes: 0}
\end{trajectoryblock}

\begin{trajectoryblock}{System message for all six calls}
\traceline{You are an autonomous agent solving an ALFWorld household task.}
\traceblank
\traceline{\# Action policy}
\traceblank
\traceline{When a skill block is provided, follow its Procedure step by step. Identify which}
\traceline{step of the procedure you are currently executing and pick the action that advances}
\traceline{that step.}
\traceblank
\traceline{\# Action priority (apply in order)}
\traceblank
\traceline{1. **If the target object is visible in the current observation AND you are not}
\traceline{\hspace*{3\tracecharwidth}holding it yet, your next action MUST be `take <target\_object> from <receptacle>`.**}
\traceline{\hspace*{3\tracecharwidth}Do NOT navigate to another location once you have seen the target. Acting beats}
\traceline{\hspace*{3\tracecharwidth}exploring.}
\traceline{2. If you are already holding the target object and the next procedure step is a}
\traceline{\hspace*{3\tracecharwidth}manipulation primitive (`cool <obj> with fridge`, `heat <obj> with microwave`,}
\traceline{\hspace*{3\tracecharwidth}`clean <obj> with sink`, `use <lamp>`), execute it directly.}
\traceline{3. **Process exactly once.** After you have run `cool` / `heat` / `clean` / `use`}
\traceline{\hspace*{3\tracecharwidth}on the target ONE time, do NOT run another manipulation primitive on it. The}
\traceline{\hspace*{3\tracecharwidth}next action is either `go to <destination>` or `put <object> in/on <destination>`.}
\traceline{4. **Destination vs appliance.** In a task like `put a cool tomato in microwave`,}
\traceline{\hspace*{3\tracecharwidth}`microwave` is the DESTINATION, not the appliance. The cooling appliance is}
\traceline{\hspace*{3\tracecharwidth}`fridge`. After cooling, your action at the microwave MUST be}
\traceline{\hspace*{3\tracecharwidth}`put <target> in/on microwave`, NOT `heat <target> with microwave`. The same}
\traceline{\hspace*{3\tracecharwidth}rule applies for sink, fridge, and lamp when they appear as destinations.}
\traceline{5. If you are holding the target object and at the destination, execute}
\traceline{\hspace*{3\tracecharwidth}`put <object> in/on <destination>`.}
\traceline{6. Only emit `go to <X>` when you are not yet at the location required by the}
\traceline{\hspace*{3\tracecharwidth}next procedure step. Pick `take` or `put` over `go to` whenever both are valid.}
\traceline{7. The target object in the task is the noun that follows \char34{}a/some/the\char34{}. For}
\traceline{\hspace*{3\tracecharwidth}example in \char34{}put a cool tomato in microwave\char34{} the target is `tomato`. The}
\traceline{\hspace*{3\tracecharwidth}appliance words (microwave, fridge, sink, lamp) are NOT the target.}
\traceblank
\traceline{\# Output}
\traceblank
\traceline{Output exactly one action from the valid actions list. Be efficient.}
\end{trajectoryblock}

\begin{trajectoryblock}{Shared user prefix A: task and complete retrieved context}
\traceline{\# Task}
\traceline{look at bowl under the desklamp}
\traceblank
\traceline{\# Retrieved context (skill+afford+loc)}
\traceline{\#\# Procedure to follow (HOW \textemdash{} the step order; fill in WHERE from below + the room)}
\traceblank
\traceline{Skill 1: examine-with-desklamp}
\traceline{\hspace*{2\tracecharwidth}description: Use a desklamp to examine a target object.}
\traceline{\hspace*{2\tracecharwidth}capability: verb=examine affordance=desklamp output=object\_examined}
\traceline{\hspace*{2\tracecharwidth}when\_to\_use: - Examine an object with a desklamp}
\traceline{- Locate and use a desklamp to inspect a target item}
\traceline{\hspace*{2\tracecharwidth}procedure:}
\traceline{\hspace*{4\tracecharwidth}- Systematic Object Hunt: Sweep unvisited surfaces and open closed containers until <target\_object> is found; never revisit empty spots. Apply this right after parsing the goal, before holding anything.}
\traceline{\hspace*{4\tracecharwidth}- Confirm Object Match: After picking something up, verify it is <target\_object>; if not, put it back and resume searching. Apply this whenever you acquire an object but the goal object is still unmet.}
\traceline{\hspace*{4\tracecharwidth}- Locate Light Source: Find and go to a desklamp; if it is not visible, systematically check unvisited receptacles. Apply this when the target object is identified but the desklamp is not in view}
\traceblank
\traceline{\#\# Constraints \textemdash{} rules about how things work (RESPECT THESE)}
\traceline{These are reliable facts about how appliances/containers behave. Violating them causes failure.}
\traceblank
\traceline{- desklamp: can be used to examine objects (20\ensuremath{\times}); can be turned on to provide light (12\ensuremath{\times})}
\traceblank
\traceline{\#\# Where to look \textemdash{} location priors (HINTS, verify against the room)}
\traceline{Where these object types are usually found. Use to decide where to search first, then trust what THIS room actually shows over the prior.}
\traceblank
\traceline{- desklamp: often located on a sidetable (7\ensuremath{\times})}
\traceline{- bowl: can be found on a desk or sidetable (2\ensuremath{\times})}
\traceblank
\traceblank
\end{trajectoryblock}

\begin{trajectoryblock}{Shared user suffix C: action rules and response format}
\traceline{\# Rule}
\traceline{You can only take / put / clean / heat / cool / use an object at a receptacle you are CURRENTLY at. If the action you want is not in the Valid actions list above, you are not there yet \textemdash{} `go to <receptacle>` first, then act.}
\traceblank
\traceline{\# Your response format}
\traceline{Thought: <one sentence about the next subgoal>}
\traceline{Action: <exactly one action copied verbatim from the valid actions list>}
\end{trajectoryblock}

\tracemeta{User message for call $i$: A + B$_i$ + C.}

\tracecall{Call 1}

\begin{trajectoryblock}{B1: complete call-specific user input}
\traceline{\# Locations you have NOT yet checked (try these for the target)}
\traceline{bed 1, desk 2, desk 1, drawer 6, drawer 5, drawer 4, drawer 3, drawer 2, drawer 1, garbagecan 1, laundryhamper 1, safe 1, shelf 6, shelf 5, shelf 4, shelf 3, shelf 2, shelf 1}
\traceblank
\traceline{\# Current observation}
\traceline{-= Welcome to TextWorld, ALFRED! =-}
\traceblank
\traceline{You are in the middle of a room. Looking quickly around you, you see a bed 1, a desk 2, a desk 1, a drawer 6, a drawer 5, a drawer 4, a drawer 3, a drawer 2, a drawer 1, a garbagecan 1, a laundryhamper 1, a safe 1, a shelf 6, a shelf 5, a shelf 4, a shelf 3, a shelf 2, and a shelf 1.}
\traceblank
\traceline{Your task is to: look at bowl under the desklamp.}
\traceblank
\traceline{\# Valid actions}
\traceline{- go to bed 1}
\traceline{- go to desk 1}
\traceline{- go to desk 2}
\traceline{- go to drawer 1}
\traceline{- go to drawer 2}
\traceline{- go to drawer 3}
\traceline{- go to drawer 4}
\traceline{- go to drawer 5}
\traceline{- go to drawer 6}
\traceline{- go to garbagecan 1}
\traceline{- go to laundryhamper 1}
\traceline{- go to safe 1}
\traceline{- go to shelf 1}
\traceline{- go to shelf 2}
\traceline{- go to shelf 3}
\traceline{- go to shelf 4}
\traceline{- go to shelf 5}
\traceline{- go to shelf 6}
\traceline{- help}
\traceline{- inventory}
\traceline{- look}
\traceblank
\traceblank
\end{trajectoryblock}

\begin{trajectoryblock}{Call 1: complete actor response}
\traceline{Thought: I need to locate the desklamp first to examine the bowl.}
\traceline{Action: go to desk 1}
\end{trajectoryblock}

\begin{trajectoryblock}{Call 1: environment return}
\traceline{observation:}
\traceline{You arrive at desk 1. On the desk 1, you see a desklamp 1, a keychain 3, a mug 2, a pen 2, a pen 1, and a pencil 2.}
\traceline{reward: 0.0}
\traceline{done: false}
\end{trajectoryblock}

\tracecall{Call 2}

\begin{trajectoryblock}{B2: complete call-specific user input}
\traceline{\# Locations you have already checked (do NOT revisit unless retrieving)}
\traceline{- desk 1: a desklamp 1, a keychain 3, a mug 2, a pen 2, a pen 1, and a pencil 2}
\traceblank
\traceline{\# Locations you have NOT yet checked (try these for the target)}
\traceline{bed 1, desk 2, drawer 6, drawer 5, drawer 4, drawer 3, drawer 2, drawer 1, garbagecan 1, laundryhamper 1, safe 1, shelf 6, shelf 5, shelf 4, shelf 3, shelf 2, shelf 1}
\traceblank
\traceline{\# Recent history}
\traceline{obs: -= Welcome to TextWorld, ALFRED! =-}
\traceblank
\traceline{You are in the middle of a room. Looking quickly around you, you see a bed 1, a desk 2, a desk 1, a drawer 6, a drawer 5, a drawer 4, a drawer 3, a drawer 2, a dra}
\traceline{action: go to desk 1}
\traceblank
\traceline{\# Current observation}
\traceline{You arrive at desk 1. On the desk 1, you see a desklamp 1, a keychain 3, a mug 2, a pen 2, a pen 1, and a pencil 2.}
\traceblank
\traceline{\# Valid actions}
\traceline{- take keychain 3 from desk 1  [PRIORITY: target visible, take now]}
\traceline{- take mug 2 from desk 1}
\traceline{- take pen 1 from desk 1}
\traceline{- take pen 2 from desk 1}
\traceline{- take pencil 2 from desk 1}
\traceline{- use desklamp 1}
\traceline{- go to bed 1}
\traceline{- go to desk 2}
\traceline{- go to drawer 1}
\traceline{- go to drawer 2}
\traceline{- go to drawer 3}
\traceline{- go to drawer 4}
\traceline{- go to drawer 5}
\traceline{- go to drawer 6}
\traceline{- go to garbagecan 1}
\traceline{- go to laundryhamper 1}
\traceline{- go to safe 1}
\traceline{- go to shelf 1}
\traceline{- go to shelf 2}
\traceline{- go to shelf 3}
\traceline{- go to shelf 4}
\traceline{- go to shelf 5}
\traceline{- go to shelf 6}
\traceline{- examine desk 1}
\traceline{- help}
\traceline{- inventory}
\traceline{- look}
\traceblank
\traceblank
\end{trajectoryblock}

\begin{trajectoryblock}{Call 2: complete actor response}
\traceline{Thought: I need to use the desklamp to examine the bowl, but first I should check if the bowl is already visible or needs to be located.}
\traceline{Action: use desklamp 1}
\end{trajectoryblock}

\begin{trajectoryblock}{Call 2: environment return}
\traceline{observation:}
\traceline{You turn on the desklamp 1.}
\traceline{reward: 0.0}
\traceline{done: false}
\end{trajectoryblock}

\tracecall{Call 3}

\begin{trajectoryblock}{B3: complete call-specific user input}
\traceline{\# Locations you have already checked (do NOT revisit unless retrieving)}
\traceline{- desk 1: a desklamp 1, a keychain 3, a mug 2, a pen 2, a pen 1, and a pencil 2}
\traceblank
\traceline{\# Locations you have NOT yet checked (try these for the target)}
\traceline{bed 1, desk 2, drawer 6, drawer 5, drawer 4, drawer 3, drawer 2, drawer 1, garbagecan 1, laundryhamper 1, safe 1, shelf 6, shelf 5, shelf 4, shelf 3, shelf 2, shelf 1}
\traceblank
\traceline{\# Recent history}
\traceline{obs: -= Welcome to TextWorld, ALFRED! =-}
\traceblank
\traceline{You are in the middle of a room. Looking quickly around you, you see a bed 1, a desk 2, a desk 1, a drawer 6, a drawer 5, a drawer 4, a drawer 3, a drawer 2, a dra}
\traceline{action: go to desk 1}
\traceline{obs: You arrive at desk 1. On the desk 1, you see a desklamp 1, a keychain 3, a mug 2, a pen 2, a pen 1, and a pencil 2.}
\traceline{action: use desklamp 1}
\traceblank
\traceline{\# Current observation}
\traceline{You turn on the desklamp 1.}
\traceblank
\traceline{\# Valid actions}
\traceline{- take keychain 3 from desk 1}
\traceline{- take mug 2 from desk 1}
\traceline{- take pen 1 from desk 1}
\traceline{- take pen 2 from desk 1}
\traceline{- take pencil 2 from desk 1}
\traceline{- use desklamp 1}
\traceline{- go to bed 1}
\traceline{- go to desk 2}
\traceline{- go to drawer 1}
\traceline{- go to drawer 2}
\traceline{- go to drawer 3}
\traceline{- go to drawer 4}
\traceline{- go to drawer 5}
\traceline{- go to drawer 6}
\traceline{- go to garbagecan 1}
\traceline{- go to laundryhamper 1}
\traceline{- go to safe 1}
\traceline{- go to shelf 1}
\traceline{- go to shelf 2}
\traceline{- go to shelf 3}
\traceline{- go to shelf 4}
\traceline{- go to shelf 5}
\traceline{- go to shelf 6}
\traceline{- examine desk 1}
\traceline{- help}
\traceline{- inventory}
\traceline{- look}
\traceblank
\traceblank
\end{trajectoryblock}

\begin{trajectoryblock}{Call 3: complete actor response}
\traceline{Thought: I need to examine the bowl using the desklamp, so I should look for the bowl in the current location or nearby locations.}
\traceline{Action: examine desk 1}
\end{trajectoryblock}

\begin{trajectoryblock}{Call 3: environment return}
\traceline{observation:}
\traceline{On the desk 1, you see a desklamp 1, a keychain 3, a mug 2, a pen 2, a pen 1, and a pencil 2.}
\traceline{reward: 0.0}
\traceline{done: false}
\end{trajectoryblock}

\tracecall{Call 4}

\begin{trajectoryblock}{B4: complete call-specific user input}
\traceline{\# Locations you have already checked (do NOT revisit unless retrieving)}
\traceline{- desk 1: a desklamp 1, a keychain 3, a mug 2, a pen 2, a pen 1, and a pencil 2}
\traceblank
\traceline{\# Locations you have NOT yet checked (try these for the target)}
\traceline{bed 1, desk 2, drawer 6, drawer 5, drawer 4, drawer 3, drawer 2, drawer 1, garbagecan 1, laundryhamper 1, safe 1, shelf 6, shelf 5, shelf 4, shelf 3, shelf 2, shelf 1}
\traceblank
\traceline{\# Recent history}
\traceline{obs: -= Welcome to TextWorld, ALFRED! =-}
\traceblank
\traceline{You are in the middle of a room. Looking quickly around you, you see a bed 1, a desk 2, a desk 1, a drawer 6, a drawer 5, a drawer 4, a drawer 3, a drawer 2, a dra}
\traceline{action: go to desk 1}
\traceline{obs: You arrive at desk 1. On the desk 1, you see a desklamp 1, a keychain 3, a mug 2, a pen 2, a pen 1, and a pencil 2.}
\traceline{action: use desklamp 1}
\traceline{obs: You turn on the desklamp 1.}
\traceline{action: examine desk 1}
\traceblank
\traceline{\# Current observation}
\traceline{On the desk 1, you see a desklamp 1, a keychain 3, a mug 2, a pen 2, a pen 1, and a pencil 2.}
\traceblank
\traceline{\# Valid actions}
\traceline{- take keychain 3 from desk 1  [PRIORITY: target visible, take now]}
\traceline{- take mug 2 from desk 1}
\traceline{- take pen 1 from desk 1}
\traceline{- take pen 2 from desk 1}
\traceline{- take pencil 2 from desk 1}
\traceline{- use desklamp 1}
\traceline{- go to bed 1}
\traceline{- go to desk 2}
\traceline{- go to drawer 1}
\traceline{- go to drawer 2}
\traceline{- go to drawer 3}
\traceline{- go to drawer 4}
\traceline{- go to drawer 5}
\traceline{- go to drawer 6}
\traceline{- go to garbagecan 1}
\traceline{- go to laundryhamper 1}
\traceline{- go to safe 1}
\traceline{- go to shelf 1}
\traceline{- go to shelf 2}
\traceline{- go to shelf 3}
\traceline{- go to shelf 4}
\traceline{- go to shelf 5}
\traceline{- go to shelf 6}
\traceline{- examine desk 1}
\traceline{- help}
\traceline{- inventory}
\traceline{- look}
\traceblank
\traceblank
\end{trajectoryblock}

\begin{trajectoryblock}{Call 4: complete actor response}
\traceline{Thought: I need to examine the bowl using the desklamp, but the bowl is not visible yet. I should continue searching for the bowl in other locations.}
\traceline{Action: go to desk 2}
\end{trajectoryblock}

\begin{trajectoryblock}{Call 4: environment return}
\traceline{observation:}
\traceline{You arrive at desk 2. On the desk 2, you see a alarmclock 1, a bowl 1, and a cd 3.}
\traceline{reward: 0.0}
\traceline{done: false}
\end{trajectoryblock}

\tracecall{Call 5}

\begin{trajectoryblock}{B5: complete call-specific user input}
\traceline{\# Locations you have already checked (do NOT revisit unless retrieving)}
\traceline{- desk 1: a desklamp 1, a keychain 3, a mug 2, a pen 2, a pen 1, and a pencil 2}
\traceline{- desk 2: a alarmclock 1, a bowl 1, and a cd 3}
\traceblank
\traceline{\# Locations you have NOT yet checked (try these for the target)}
\traceline{bed 1, drawer 6, drawer 5, drawer 4, drawer 3, drawer 2, drawer 1, garbagecan 1, laundryhamper 1, safe 1, shelf 6, shelf 5, shelf 4, shelf 3, shelf 2, shelf 1}
\traceblank
\traceline{\# Recent history}
\traceline{obs: -= Welcome to TextWorld, ALFRED! =-}
\traceblank
\traceline{You are in the middle of a room. Looking quickly around you, you see a bed 1, a desk 2, a desk 1, a drawer 6, a drawer 5, a drawer 4, a drawer 3, a drawer 2, a dra}
\traceline{action: go to desk 1}
\traceline{obs: You arrive at desk 1. On the desk 1, you see a desklamp 1, a keychain 3, a mug 2, a pen 2, a pen 1, and a pencil 2.}
\traceline{action: use desklamp 1}
\traceline{obs: You turn on the desklamp 1.}
\traceline{action: examine desk 1}
\traceline{obs: On the desk 1, you see a desklamp 1, a keychain 3, a mug 2, a pen 2, a pen 1, and a pencil 2.}
\traceline{action: go to desk 2}
\traceblank
\traceline{\# Current observation}
\traceline{You arrive at desk 2. On the desk 2, you see a alarmclock 1, a bowl 1, and a cd 3.}
\traceblank
\traceline{\# Valid actions}
\traceline{- take alarmclock 1 from desk 2  [PRIORITY: target visible, take now]}
\traceline{- take bowl 1 from desk 2}
\traceline{- take cd 3 from desk 2}
\traceline{- go to bed 1}
\traceline{- go to desk 1}
\traceline{- go to drawer 1}
\traceline{- go to drawer 2}
\traceline{- go to drawer 3}
\traceline{- go to drawer 4}
\traceline{- go to drawer 5}
\traceline{- go to drawer 6}
\traceline{- go to garbagecan 1}
\traceline{- go to laundryhamper 1}
\traceline{- go to safe 1}
\traceline{- go to shelf 1}
\traceline{- go to shelf 2}
\traceline{- go to shelf 3}
\traceline{- go to shelf 4}
\traceline{- go to shelf 5}
\traceline{- go to shelf 6}
\traceline{- examine desk 2}
\traceline{- help}
\traceline{- inventory}
\traceline{- look}
\traceblank
\traceblank
\end{trajectoryblock}

\begin{trajectoryblock}{Call 5: complete actor response}
\traceline{Thought: The target object \char34{}bowl\char34{} is visible on desk 2, so I should take it.}
\traceline{Action: take bowl 1 from desk 2}
\end{trajectoryblock}

\begin{trajectoryblock}{Call 5: environment return}
\traceline{observation:}
\traceline{You pick up the bowl 1 from the desk 2.}
\traceline{reward: 0.0}
\traceline{done: false}
\end{trajectoryblock}

\tracecall{Call 6}

\begin{trajectoryblock}{B6: complete call-specific user input}
\traceline{\# Locations you have already checked (do NOT revisit unless retrieving)}
\traceline{- desk 1: a desklamp 1, a keychain 3, a mug 2, a pen 2, a pen 1, and a pencil 2}
\traceline{- desk 2: a alarmclock 1, a bowl 1, and a cd 3}
\traceblank
\traceline{\# Locations you have NOT yet checked (try these for the target)}
\traceline{bed 1, drawer 6, drawer 5, drawer 4, drawer 3, drawer 2, drawer 1, garbagecan 1, laundryhamper 1, safe 1, shelf 6, shelf 5, shelf 4, shelf 3, shelf 2, shelf 1}
\traceblank
\traceline{\# Recent history}
\traceline{obs: -= Welcome to TextWorld, ALFRED! =-}
\traceblank
\traceline{You are in the middle of a room. Looking quickly around you, you see a bed 1, a desk 2, a desk 1, a drawer 6, a drawer 5, a drawer 4, a drawer 3, a drawer 2, a dra}
\traceline{action: go to desk 1}
\traceline{obs: You arrive at desk 1. On the desk 1, you see a desklamp 1, a keychain 3, a mug 2, a pen 2, a pen 1, and a pencil 2.}
\traceline{action: use desklamp 1}
\traceline{obs: You turn on the desklamp 1.}
\traceline{action: examine desk 1}
\traceline{obs: On the desk 1, you see a desklamp 1, a keychain 3, a mug 2, a pen 2, a pen 1, and a pencil 2.}
\traceline{action: go to desk 2}
\traceline{obs: You arrive at desk 2. On the desk 2, you see a alarmclock 1, a bowl 1, and a cd 3.}
\traceline{action: take bowl 1 from desk 2}
\traceblank
\traceline{\# Current observation}
\traceline{You pick up the bowl 1 from the desk 2.}
\traceblank
\traceline{\# Valid actions}
\traceline{- go to bed 1}
\traceline{- go to desk 1}
\traceline{- go to drawer 1}
\traceline{- go to drawer 2}
\traceline{- go to drawer 3}
\traceline{- go to drawer 4}
\traceline{- go to drawer 5}
\traceline{- go to drawer 6}
\traceline{- go to garbagecan 1}
\traceline{- go to laundryhamper 1}
\traceline{- go to safe 1}
\traceline{- go to shelf 1}
\traceline{- go to shelf 2}
\traceline{- go to shelf 3}
\traceline{- go to shelf 4}
\traceline{- go to shelf 5}
\traceline{- go to shelf 6}
\traceline{- examine bowl 1}
\traceline{- examine desk 2}
\traceline{- help}
\traceline{- inventory}
\traceline{- look}
\traceline{- move bowl 1 to desk 2}
\traceblank
\traceblank
\end{trajectoryblock}

\begin{trajectoryblock}{Call 6: complete actor response}
\traceline{Thought: I need to examine the bowl using the desklamp, so I should go to the desklamp\char39{}s location.}
\traceline{Action: go to desk 1}
\end{trajectoryblock}

\begin{trajectoryblock}{Call 6: environment return}
\traceline{observation:}
\traceline{You arrive at desk 1. On the desk 1, you see a desklamp 1, a keychain 3, a mug 2, a pen 2, a pen 1, and a pencil 2.}
\traceline{reward: 1.0}
\traceline{done: true}
\end{trajectoryblock}

\par\nopagebreak\noindent\begin{minipage}{\linewidth}
\captionsetup{aboveskip=3pt}
\captionof{figure}{\textbf{ALFWorld task trajectory.} The six calls include every changing input, actor response, and environment return. The shared prompt blocks appear once.}
\label{fig:qualitative-demo-alfworld}
\end{minipage}

% Generated from saved model calls. Do not edit quoted content.
\clearpage

\subsubsection{DialSim-subst.}

\tracemeta{1 actor call \quad / \quad 0 skills / 12 facts / 10 reference episodes}

\begin{trajectoryblock}{Retrieved record inventory}
\traceline{skills: 0}
\traceline{facts: 12}
\traceline{\hspace*{2\tracecharwidth}0c58d947: diana | Diana snagged new boots using Dad\char39{}s credit card}
\traceline{\hspace*{2\tracecharwidth}7bb0e489: dad\char39{}s credit card | Diana used Dad\char39{}s credit card to buy the swanky new boots}
\traceline{\hspace*{2\tracecharwidth}c2c7c7ef: diana | Diana used Dad\char39{}s credit card to buy the swanky new boots on September 24, 1994.}
\traceline{\hspace*{2\tracecharwidth}1c388d07: diana | Diana is asking about the method used to get the new boots}
\traceline{\hspace*{2\tracecharwidth}2a94a7d5: diana | Diana found money outside Central Perk}
\traceline{\hspace*{2\tracecharwidth}1c533b0b: diana | Diana snagged swanky new boots}
\traceline{\hspace*{2\tracecharwidth}b2df521d: money | Elena lent Diana some money}
\traceline{\hspace*{2\tracecharwidth}9ee778e7: elena | Elena lent money to Diana because she felt bad for her}
\traceline{\hspace*{2\tracecharwidth}1d23b70b: diana | Diana got a job}
\traceline{\hspace*{2\tracecharwidth}66652c40: speaker | The options provided were Lincoln High, Central Perk High, New York High School, Liberty High, and \char34{}I don\char39{}t know\char34{}}
\traceline{\hspace*{2\tracecharwidth}00ac36c6: diana | Answered the question about the high school with \char34{}Lincoln High\char34{}}
\traceline{\hspace*{2\tracecharwidth}739c2e59: diana | discusses her job search and purchase of boots on September 23, 1994}
\traceline{reference episodes: 10}
\traceline{\hspace*{2\tracecharwidth}prompt indices: 1, 2, 3, 4, 5, 6, 7, 8, 9, 10}
\end{trajectoryblock}

\begin{trajectoryblock}{Complete user message: task, provided context, and answer instructions}
\traceline{You answer questions about people from your accumulated memory of their conversations. Today is September 23, 1994.}
\traceblank
\traceline{\#\# Answer style (follow these formats exactly)}
\traceline{- \char34{}When did X attend the workshop?\char34{} -> 7 May 2023}
\traceline{- If a session dated 27 June 2023 says something happened \char34{}last week\char34{} -> The week before 27 June 2023}
\traceline{- \char34{}What did X cook for the party?\char34{} -> spaghetti with meatballs}
\traceline{- \char34{}How long did X\char39{}s trip last?\char34{} -> two weeks}
\traceblank
\traceline{\#\# Where to look \textemdash{} location priors (HINTS, verify against the room)}
\traceline{Where these object types are usually found. Use to decide where to search first, then trust what THIS room actually shows over the prior.}
\traceblank
\traceline{- diana: Diana used Dad\char39{}s credit card to buy the swanky new boots on September 24, 1994. (234\ensuremath{\times}); Diana snagged new boots using Dad\char39{}s credit card (193\ensuremath{\times}); Answered the question about the high school with \char34{}Lincoln High\char34{} (31\ensuremath{\times}); Diana snagged swanky new boots (28\ensuremath{\times}); Diana is asking about the method used to get the new boots (16\ensuremath{\times}); Diana found money outside Central Perk (1\ensuremath{\times}); Diana got a job (1\ensuremath{\times}); discusses her job search and purchase of boots on September 23, 1994 (1\ensuremath{\times})}
\traceline{- dad\char39{}s credit card: Diana used Dad\char39{}s credit card to buy the swanky new boots (243\ensuremath{\times})}
\traceline{- money: Elena lent Diana some money (1\ensuremath{\times})}
\traceline{- elena: Elena lent money to Diana because she felt bad for her (1\ensuremath{\times})}
\traceline{- speaker: The options provided were Lincoln High, Central Perk High, New York High School, Liberty High, and \char34{}I don\char39{}t know\char34{} (41\ensuremath{\times})}
\traceblank
\traceline{\#\# Reference episodes (from PAST situations \textemdash{} the procedure may transfer; specific places/objects WILL differ here. Trust the current observations over these.)}
\traceblank
\traceline{Experience 1 (weight=1.00):}
\traceline{\hspace*{2\tracecharwidth}question: Review the conversation session dated October 13, 1995 and remember the important facts about the speakers. (part 2)}
\traceline{\hspace*{2\tracecharwidth}improvement: VERBATIM EXCERPT:}
\traceline{\hspace*{1\tracecharwidth}then maybe um brush the hair away from my face, and look far into my eyes in a way that let\char39{}s me know that something amazing is about to happen.}
\traceline{Victor:  Uh-huh.}
\traceline{Diana: And then, I don\char39{}t know, I mean you\char39{}d pull me really close to you so that, so that I\char39{}d be pressed up, you know, right against you. And, um, it would get kind of sweaty and uh, and blurry, and then it\char39{}s just happen}
\traceblank
\traceline{Experience 2 (weight=1.00):}
\traceline{\hspace*{2\tracecharwidth}question: Review the conversation session dated October 12, 1995 and remember the important facts about the speakers. (part 2)}
\traceline{\hspace*{2\tracecharwidth}improvement: VERBATIM EXCERPT:}
\traceline{s, what did you think a nubbin was?}
\traceline{Tobias: I don\char39{}t know, you see somethin\char39{}, you hear a word, I thought that\char39{}s what it was. Let me see it again.}
\traceline{All: Yeah, show it. Show it. The nubbin, the nubbin, the nubbin.}
\traceline{Marcus: Tobias was in a porno movie.}
\traceline{All: Huuh.}
\traceline{Marcus: If I\char39{}m goin\char39{} down, I\char39{}m takin\char39{} everybody with me.}
\traceline{Victor: You were in a porno?}
\traceline{Tobias: Ahh, alright, alright, alright}
\traceblank
\traceline{Experience 3 (weight=1.00):}
\traceline{\hspace*{2\tracecharwidth}question: Review the conversation session dated October 6, 1995 and remember the important facts about the speakers. (part 2)}
\traceline{\hspace*{2\tracecharwidth}improvement: VERBATIM EXCERPT:}
\traceline{, and then, all of a sudden, it\char39{}s just magically broken?}
\traceline{Elena: Iris, tell her!}
\traceline{Iris: Ok, I didn\char39{}t see it, because I was putting on my jacket, but I uh want to believe you.}
\traceline{Diana: Hey Marcus. Elena just broke my seashell lamp.}
\traceline{Marcus: Neat. I\char39{}m gonna die alone.}
\traceline{Diana: Ok, you win.}
\traceline{Elena: Marcus, you\char39{}re not gonna die alone.}
\traceline{Marcus: Paula was my safety net, ok? And now I have to ge}
\traceblank
\traceline{Experience 4 (weight=1.00):}
\traceline{\hspace*{2\tracecharwidth}question: Review the conversation session dated September 28, 1995 and remember the important facts about the speakers. (part 2)}
\traceline{\hspace*{2\tracecharwidth}improvement: VERBATIM EXCERPT:}
\traceline{e little toys move? Opposable thumbs. Without evolution, how do you explain opposable thumbs?}
\traceline{Iris: Maybe the overlords needed them to steer their spacecrafts.}
\traceline{Victor: Please tell me you\char39{}re joking.}
\traceline{Iris: Look, can\char39{}t we just say that you believe in something, and I don\char39{}t.}
\traceline{Victor: No, no, Pheebs, we can\char39{}t, ok, because--}
\traceline{Iris: What is this obsessive need you have to make everyone ag}
\traceblank
\traceline{Experience 5 (weight=1.30):}
\traceline{\hspace*{2\tracecharwidth}question: review a conversation session and remember important facts about the speakers}
\traceline{\hspace*{2\tracecharwidth}improvement: analyze the conversation to identify key speakers, extract their statements, and note recurring themes and relationships}
\traceblank
\traceline{VERBATIM EXCERPT:}
\traceline{Read this past conversation session and note the important facts (dated September 23, 1994):}
\traceblank
\traceblank
\traceline{Diana: Oh God, come on you guys, is this really necessary?  I mean, I can stop charging anytime I want.}
\traceline{Elena: C\char39{}mon, you can\char39{}t live off your parents your whole life}
\traceblank
\traceline{Experience 6 (weight=1.30):}
\traceline{\hspace*{2\tracecharwidth}question: Review a conversation session and remember important facts about the speakers}
\traceline{\hspace*{2\tracecharwidth}improvement: Identify all speakers in the conversation, note their key statements and interactions, and summarize their roles and relationships within the dialogue.}
\traceblank
\traceline{VERBATIM EXCERPT:}
\traceline{Read this past conversation session and note the important facts (dated September 23, 1994):}
\traceblank
\traceblank
\traceline{Tobias: Of course it was a line!}
\traceline{Elena: Why?! Why? Why, why would anybody do something like that?}
\traceline{Victor: I assume we\char39{}re looking for a}
\traceblank
\traceline{Experience 7 (weight=1.30):}
\traceline{\hspace*{2\tracecharwidth}question: Review a conversation session and remember important facts about the speakers}
\traceline{\hspace*{2\tracecharwidth}improvement: Identify the date and speakers involved, note key themes and relationships, summarize each speaker\char39{}s contribution, and extract main facts about the conversation\char39{}s context and outcome}
\traceblank
\traceline{VERBATIM EXCERPT:}
\traceline{Read this past conversation session and note the important facts (dated November 24, 1994):}
\traceblank
\traceblank
\traceline{Tobias: Nope, not that one.}
\traceline{Elena: Can you go any faster with that?}
\traceline{Tobias: Hey, I got one keyhole and a}
\traceblank
\traceline{Experience 8 (weight=1.30):}
\traceline{\hspace*{2\tracecharwidth}question: review a conversation session and remember important facts about the speakers}
\traceline{\hspace*{2\tracecharwidth}improvement: listen to the conversation, identify all speakers, note key topics discussed, and record relationships and themes between participants}
\traceblank
\traceline{VERBATIM EXCERPT:}
\traceline{Read this past conversation session and note the important facts (dated April 27, 1995):}
\traceblank
\traceblank
\traceline{Tobias: How could someone get a hold of your credit card number?}
\traceline{Elena: I have no idea. But look how much they spent!}
\traceline{Diana: Elena, would you calm down? Th}
\traceblank
\traceline{Experience 9 (weight=1.30):}
\traceline{\hspace*{2\tracecharwidth}question: review a conversation and remember important facts about the speakers}
\traceline{\hspace*{2\tracecharwidth}improvement: analyze the conversation to identify key speakers and their statements, then extract and remember the important facts about each speaker\char39{}s role and statements}
\traceblank
\traceline{VERBATIM EXCERPT:}
\traceline{Read this past conversation session and note the important facts (dated September 22, 1994):}
\traceblank
\traceblank
\traceline{Elena: Now I\char39{}m guessing that he bought her the big pipe organ, and she\char39{}s really not happy about it.}
\traceline{Marcus:  Tuna or egg salad?}
\traceblank
\traceline{Experience 10 (weight=1.30):}
\traceline{\hspace*{2\tracecharwidth}question: review a conversation session and remember important facts about the speakers}
\traceline{\hspace*{2\tracecharwidth}improvement: listen to the conversation session, identify the date and speakers, note key themes and speaker contributions, summarize the main points and character dynamics}
\traceblank
\traceline{VERBATIM EXCERPT:}
\traceline{Read this past conversation session and note the important facts (dated March 6, 1995):}
\traceblank
\traceblank
\traceline{Victor: So, you gals wanna hand over your money now? That way, we don\char39{}t have to go through the formality of actually playing.}
\traceline{Diana}
\traceblank
\traceline{Question: So, could we BE any more curious about how Diana snagged those new boots?}
\traceline{Options: She found money outside Central Perk / She got a job / Dad\char39{}s credit card / Elena lent her some money because she felt bad for her / I don\char39{}t know. (None of the above)}
\traceline{Answer with the exact option text.}
\traceline{Answer with ONLY a short phrase (a few words, a date, or a name). Give your best short answer whenever ANY fact or excerpt relates to the question \textemdash{} only if absolutely nothing relates, answer: No information available.}
\traceline{Answer:}
\end{trajectoryblock}

\begin{trajectoryblock}{Complete actor response}
\traceline{Dad\char39{}s credit card}
\end{trajectoryblock}

\begin{trajectoryblock}{Adapter submission}
\traceline{Answer: Dad\char39{}s credit card}
\end{trajectoryblock}

\begin{trajectoryblock}{Environment return}
\traceline{observation:}
\traceline{(answered)}
\traceline{reward: 1.0}
\traceline{done: true}
\end{trajectoryblock}

\begin{trajectoryblock}{Evaluation result}
\traceline{reference\_answer: Dad\char39{}s credit card}
\traceline{token F1: 100\%}
\end{trajectoryblock}

\par\nopagebreak\noindent\begin{minipage}{\linewidth}
\captionsetup{aboveskip=3pt}
\captionof{figure}{\textbf{DialSim task trajectory.} The complete prompt contains facts and conversation records without a skill. The answer matches the reference.}
\label{fig:qualitative-demo-dialsim}
\end{minipage}

% Full raw prompts and responses. No reference solution replaces model output.
% Generated from saved model calls. Do not edit quoted content.
\clearpage

\subsubsection{MATH-hard}

\tracemeta{1 actor call \quad / \quad 1 skill / 4 facts}

\begin{trajectoryblock}{Retrieved record inventory}
\traceline{skills: 1}
\traceline{\hspace*{2\tracecharwidth}12190a98: simplify-and-convert}
\traceline{facts: 4}
\traceline{\hspace*{2\tracecharwidth}68b4340e: circle | the equation can be transformed into standard form by rearranging and completing the square}
\traceline{\hspace*{2\tracecharwidth}72783530: circle | can intersect with another circle at most twice}
\traceline{\hspace*{2\tracecharwidth}1f0be7bd: circle | can overlap with another circle of different radius}
\traceline{\hspace*{2\tracecharwidth}3163899d: circular base | area is proportional to the square of the radius}
\traceline{reference episodes: 0}
\end{trajectoryblock}

\begin{trajectoryblock}{Complete system message}
\traceline{You are an expert problem solver working through a stream of tasks.}
\traceblank
\traceline{Skill blocks and memory notes, when provided, are advisory: use them if they help,}
\traceline{ignore them if they do not apply to the current task.}
\traceblank
\traceline{Work each task IN FULL \textemdash{} reason step by step as deeply as the problem requires.}
\traceline{Follow the answer-format instructions given in the observation (e.g. include ONE}
\traceline{```python block when allowed, and finish with the requested answer line).}
\end{trajectoryblock}

\begin{trajectoryblock}{Complete user message: task, provided context, and answer instructions}
\traceline{\# Task}
\traceline{Solve this competition math problem. Put your final answer in \textbackslash{}boxed\{\}. Find the center of the circle with equation \$x\textasciicircum{}2 - 6x + y\textasciicircum{}2 + 2y = 9\$.}
\traceblank
\traceline{\# Retrieved context (skill+afford)}
\traceline{\#\# Procedure to follow (HOW \textemdash{} the step order; fill in WHERE from below + the room)}
\traceblank
\traceline{Skill 1: simplify-and-convert}
\traceline{\hspace*{2\tracecharwidth}description: Use this skill to simplify mathematical expressions, convert units, or transform numbers between formats.}
\traceline{\hspace*{2\tracecharwidth}capability: verb=examine affordance=none output=object\_processed}
\traceline{\hspace*{2\tracecharwidth}when\_to\_use: - Simplify a mathematical expression to a specific form}
\traceline{- Convert a number between different units or number bases}
\traceline{- Transform a repeating decimal into a fraction}
\traceline{- Express a radical in its simplest form}
\traceline{\hspace*{2\tracecharwidth}procedure:}
\traceline{\hspace*{4\tracecharwidth}- Systematic Expression Analysis: Examine the structure of <target> to identify the type of transformation needed; look for radicals, repeating decimals, or unit conversions. Apply this when the task is first parsed.}
\traceline{\hspace*{4\tracecharwidth}- Confirm Target Type: If <target> is a repeating decimal, set up an equation to isolate the repeating part. Apply this when the target is identified as a repeating decimal.}
\traceline{\hspace*{4\tracecharwidth}- Apply Simplification Rules: If <target> contains square roots, rationalize denominators and combine like terms. Apply this when the target involves radicals.}
\traceline{\hspace*{4\tracecharwidth}- Convert Units or Bases: If <target> requires un}
\traceblank
\traceline{\#\# Constraints \textemdash{} rules about how things work (RESPECT THESE)}
\traceline{These are reliable facts about how appliances/containers behave. Violating them causes failure.}
\traceblank
\traceline{- circle: the equation can be transformed into standard form by rearranging and completing the square (1\ensuremath{\times}); can intersect with another circle at most twice (1\ensuremath{\times}); can overlap with another circle of different radius (1\ensuremath{\times})}
\traceline{- circular base: area is proportional to the square of the radius (1\ensuremath{\times})}
\traceblank
\traceline{\# Current observation}
\traceline{Solve this competition math problem. Put your final answer in \textbackslash{}boxed\{\}.}
\traceblank
\traceline{Find the center of the circle with equation \$x\textasciicircum{}2 - 6x + y\textasciicircum{}2 + 2y = 9\$.}
\traceblank
\traceline{You may reason, and you may include ONE ```python block \textemdash{} I will execute it and show you the output. Finish with: The answer is <answer>}
\traceblank
\traceline{\# Valid actions}
\traceline{- <your reasoning / python / answer>}
\traceblank
\traceline{\# Your response format}
\traceline{Free form. Reason step by step as deeply as needed, then follow the answer format asked for in the observation.}
\end{trajectoryblock}

\begin{trajectoryblock}{Complete actor response}
\traceline{To find the center of the circle given by the equation:}
\traceblank
\traceline{\$\$}
\traceline{x\textasciicircum{}2 - 6x + y\textasciicircum{}2 + 2y = 9}
\traceline{\$\$}
\traceblank
\traceline{we need to rewrite the equation in the **standard form of a circle**, which is:}
\traceblank
\traceline{\$\$}
\traceline{(x - h)\textasciicircum{}2 + (y - k)\textasciicircum{}2 = r\textasciicircum{}2}
\traceline{\$\$}
\traceblank
\traceline{where \$(h, k)\$ is the center of the circle and \$r\$ is the radius.}
\traceblank
\traceline{---}
\traceblank
\traceline{\#\#\# Step 1: Group and complete the square}
\traceblank
\traceline{We start by grouping the \$x\$-terms and \$y\$-terms:}
\traceblank
\traceline{\$\$}
\traceline{(x\textasciicircum{}2 - 6x) + (y\textasciicircum{}2 + 2y) = 9}
\traceline{\$\$}
\traceblank
\traceline{Now, complete the square for both \$x\$ and \$y\$.}
\traceblank
\traceline{\#\#\#\# For \$x\textasciicircum{}2 - 6x\$:}
\traceline{- Take half of \$-6\$, which is \$-3\$, and square it: \$(-3)\textasciicircum{}2 = 9\$}
\traceline{- Add and subtract 9:}
\traceblank
\traceline{\$\$}
\traceline{x\textasciicircum{}2 - 6x = (x - 3)\textasciicircum{}2 - 9}
\traceline{\$\$}
\traceblank
\traceline{\#\#\#\# For \$y\textasciicircum{}2 + 2y\$:}
\traceline{- Take half of \$2\$, which is \$1\$, and square it: \$1\textasciicircum{}2 = 1\$}
\traceline{- Add and subtract 1:}
\traceblank
\traceline{\$\$}
\traceline{y\textasciicircum{}2 + 2y = (y + 1)\textasciicircum{}2 - 1}
\traceline{\$\$}
\traceblank
\traceline{---}
\traceblank
\traceline{\#\#\# Step 2: Substitute back into the equation}
\traceblank
\traceline{\$\$}
\traceline{(x - 3)\textasciicircum{}2 - 9 + (y + 1)\textasciicircum{}2 - 1 = 9}
\traceline{\$\$}
\traceblank
\traceline{Simplify:}
\traceblank
\traceline{\$\$}
\traceline{(x - 3)\textasciicircum{}2 + (y + 1)\textasciicircum{}2 - 10 = 9}
\traceline{\$\$}
\traceblank
\traceline{\$\$}
\traceline{(x - 3)\textasciicircum{}2 + (y + 1)\textasciicircum{}2 = 19}
\traceline{\$\$}
\traceblank
\traceline{---}
\traceblank
\traceline{\#\#\# Step 3: Identify the center}
\traceblank
\traceline{From the standard form \$(x - h)\textasciicircum{}2 + (y - k)\textasciicircum{}2 = r\textasciicircum{}2\$, we can see that:}
\traceblank
\traceline{- \$h = 3\$}
\traceline{- \$k = -1\$}
\traceblank
\traceline{So, the center of the circle is:}
\traceblank
\traceline{\$\$}
\traceline{(3, -1)}
\traceline{\$\$}
\traceblank
\traceline{---}
\traceblank
\traceline{\#\#\# Final Answer}
\traceblank
\traceline{\$\$}
\traceline{\textbackslash{}boxed\{(3, -1)\}}
\traceline{\$\$}
\end{trajectoryblock}

\begin{trajectoryblock}{Environment return}
\traceline{observation:}
\traceline{(submitted)}
\traceline{reward: 1.0}
\traceline{done: true}
\end{trajectoryblock}

\begin{trajectoryblock}{Evaluation result}
\traceline{reference\_answer: (3, -1)}
\end{trajectoryblock}

\par\nopagebreak\noindent\begin{minipage}{\linewidth}
\captionsetup{aboveskip=3pt}
\captionof{figure}{\textbf{MATH-hard task trajectory.} The complete actor response ends with the reference answer.}
\label{fig:qualitative-demo-math}
\end{minipage}

% Generated from saved model calls. Do not edit quoted content.
\clearpage

\subsubsection{MBPP}

\tracemeta{1 actor call \quad / \quad 2 skills / 4 facts / 2 reference episodes}

\begin{trajectoryblock}{Retrieved record inventory}
\traceline{skills: 2}
\traceline{\hspace*{2\tracecharwidth}c1a741f2: filter-string-content}
\traceline{\hspace*{2\tracecharwidth}0a4e3b3e: string-processing}
\traceline{facts: 4}
\traceline{\hspace*{2\tracecharwidth}82ade9b6: python function | can be named to match expected test case names}
\traceline{\hspace*{2\tracecharwidth}28f550cb: string | can be modified by replacing specific characters}
\traceline{\hspace*{2\tracecharwidth}37dd4687: string | can be indexed to access individual characters}
\traceline{\hspace*{2\tracecharwidth}28c906da: function | can be defined to perform specific string manipulation tasks}
\traceline{reference episodes: 2}
\traceline{\hspace*{2\tracecharwidth}prompt indices: 1, 2}
\end{trajectoryblock}

\begin{trajectoryblock}{Complete system message}
\traceline{You are an expert problem solver working through a stream of tasks.}
\traceblank
\traceline{Skill blocks and memory notes, when provided, are advisory: use them if they help,}
\traceline{ignore them if they do not apply to the current task.}
\traceblank
\traceline{Work each task IN FULL \textemdash{} reason step by step as deeply as the problem requires.}
\traceline{Follow the answer-format instructions given in the observation (e.g. include ONE}
\traceline{```python block when allowed, and finish with the requested answer line).}
\end{trajectoryblock}

\begin{trajectoryblock}{Complete user message: task, provided context, and answer instructions}
\traceline{\# Task}
\traceline{Write a python function to find the first repeated character in a given string.}
\traceblank
\traceline{\# Retrieved context (skill+afford+raw)}
\traceline{\#\# Procedure to follow (HOW \textemdash{} the step order; fill in WHERE from below + the room)}
\traceblank
\traceline{Skill 1: filter-string-content}
\traceline{\hspace*{2\tracecharwidth}description: Use this skill to filter characters from a string based on specific criteria.}
\traceline{\hspace*{2\tracecharwidth}capability: verb=clean affordance=none output=object\_processed}
\traceline{\hspace*{2\tracecharwidth}when\_to\_use: - Filter characters from a string based on specific conditions.}
\traceline{- Remove characters that meet certain criteria from a given string.}
\traceline{- Process a string to retain only characters that do not match a defined set.}
\traceline{\hspace*{2\tracecharwidth}procedure:}
\traceline{\hspace*{4\tracecharwidth}- Systematic Character Inspection: Iterate through each character in <target>, checking if it meets the <criteria>; apply this when the target string is in inventory and the criteria is defined.}
\traceline{\hspace*{4\tracecharwidth}- Conditional Character Retention: Keep characters that do not meet the <criteria> and combine them into a new string; apply this when the character inspection is complete.}
\traceline{\hspace*{4\tracecharwidth}- Result Formation: Combine the filtered characters into a new string and return it as the result; apply this when all characters have been processed.}
\traceblank
\traceline{Skill 2: string-processing}
\traceline{\hspace*{2\tracecharwidth}description: Use this skill to process strings by filtering, transforming, or reformatting based on specific rules.}
\traceline{\hspace*{2\tracecharwidth}when\_to\_use: - When transforming a string by removing specific characters}
\traceline{- When applying a rule to filter or modify a string}
\traceline{- When reformatting a string according to a defined pattern}
\traceline{\hspace*{2\tracecharwidth}procedure:}
\traceline{\hspace*{4\tracecharwidth}- Define Processing Rule: Establish the rule that will be applied to <target> to determine which characters to keep or modify. Apply this before any string manipulation.}
\traceline{\hspace*{4\tracecharwidth}- Filter Characters: Iterate through each character in <target> and apply the defined rule to determine if it should be included in the result. Apply this when the rule is defined and the string is in inventory.}
\traceline{\hspace*{4\tracecharwidth}- Construct Result: Combine the filtered characters into a new string and return it as the processed result. Apply this when the filtering is complete and the result is ready.}
\traceline{\hspace*{4\tracecharwidth}- Verify Output: Confirm that the processe}
\traceblank
\traceline{\#\# Constraints \textemdash{} rules about how things work (RESPECT THESE)}
\traceline{These are reliable facts about how appliances/containers behave. Violating them causes failure.}
\traceblank
\traceline{- python function: can be named to match expected test case names (5\ensuremath{\times})}
\traceline{- string: can be modified by replacing specific characters (2\ensuremath{\times}); can be indexed to access individual characters (2\ensuremath{\times})}
\traceline{- function: can be defined to perform specific string manipulation tasks (3\ensuremath{\times})}
\traceblank
\traceline{\#\# Reference episodes (from PAST situations \textemdash{} the procedure may transfer; specific places/objects WILL differ here. Trust the current observations over these.)}
\traceblank
\traceline{Experience 1 (weight=1.30):}
\traceline{\hspace*{2\tracecharwidth}question: write a python function to find the first non-repeated character in a given string}
\traceline{\hspace*{2\tracecharwidth}improvement: count the frequency of each character in the string using a dictionary, then iterate through the string again to find the first character with a frequency of 1}
\traceblank
\traceline{Experience 2 (weight=1.30):}
\traceline{\hspace*{2\tracecharwidth}question: Write a function to remove first and last occurrence of a given character from a string}
\traceline{\hspace*{2\tracecharwidth}improvement: Use find and rfind to locate first and last occurrence of the character. If found, slice the string to exclude those positions. Return original string if character not found.}
\traceblank
\traceline{\# Current observation}
\traceline{Write a python function to find the first repeated character in a given string.}
\traceline{Name the function exactly `first\_repeated\_char`.}
\traceline{Your code should pass tests like: assert first\_repeated\_char(\char34{}abcabc\char34{}) == \char34{}a\char34{}}
\traceblank
\traceline{You may reason, and you may include ONE ```python block \textemdash{} I will execute it and show you the output. Finish with: The answer is <answer>}
\traceline{For coding tasks, provide the complete solution in the ```python block (the block itself is your answer).}
\traceblank
\traceline{\# Valid actions}
\traceline{- <your reasoning / python / answer>}
\traceblank
\traceline{\# Your response format}
\traceline{Free form. Reason step by step as deeply as needed, then follow the answer format asked for in the observation.}
\end{trajectoryblock}

\begin{trajectoryblock}{Complete actor response}
\traceline{To solve the problem of finding the first repeated character in a given string, we can follow a systematic approach:}
\traceblank
\traceline{1. **Track Character Frequencies**: We can use a dictionary to count the frequency of each character in the string. This allows us to identify which characters are repeated.}
\traceline{2. **Identify the First Repeated Character**: Once we have the frequency counts, we can iterate through the string again and check for the first character that has a frequency greater than 1.}
\traceblank
\traceline{This approach ensures that we efficiently determine the first repeated character in the string.}
\traceblank
\traceline{Here is the implementation of the function `first\_repeated\_char`:}
\traceblank
\begin{samepage}
\traceline{```python}
\traceline{def first\_repeated\_char(s):}
\traceline{\hspace*{4\tracecharwidth}\# Step 1: Count the frequency of each character}
\traceline{\hspace*{4\tracecharwidth}char\_count = \{\}}
\traceline{\hspace*{4\tracecharwidth}for char in s:}
\traceline{\hspace*{8\tracecharwidth}if char in char\_count:}
\traceline{\hspace*{12\tracecharwidth}char\_count[char] += 1}
\traceline{\hspace*{8\tracecharwidth}else:}
\traceline{\hspace*{12\tracecharwidth}char\_count[char] = 1}
\traceblank
\traceline{\hspace*{4\tracecharwidth}\# Step 2: Find the first character with a frequency greater than 1}
\traceline{\hspace*{4\tracecharwidth}for char in s:}
\traceline{\hspace*{8\tracecharwidth}if char\_count[char] > 1:}
\traceline{\hspace*{12\tracecharwidth}return char}
\traceblank
\traceline{\hspace*{4\tracecharwidth}\# If no repeated character is found, return None or an appropriate value}
\traceline{\hspace*{4\tracecharwidth}return None}
\traceline{```}
\end{samepage}
\traceblank
\traceline{The answer is `first\_repeated\_char`}
\end{trajectoryblock}

\begin{trajectoryblock}{Environment return}
\traceline{observation:}
\traceline{(submitted)}
\traceline{reward: 1.0}
\traceline{done: true}
\end{trajectoryblock}

\begin{trajectoryblock}{Evaluation result}
\traceline{provided tests passed: 3/3}
\end{trajectoryblock}

\par\nopagebreak\noindent\begin{minipage}{\linewidth}
\captionsetup{aboveskip=3pt}
\captionof{figure}{\textbf{MBPP task trajectory.} The original submission passes the three provided tests.}
\label{fig:qualitative-demo-mbpp}
\end{minipage}

% END INLINED FROM appendix.tex

\end{document}